\documentclass[11pt]{article}

\usepackage[final]{acl}

\usepackage{times}
\usepackage{latexsym}

\usepackage[T1]{fontenc}
\usepackage[utf8]{inputenc}
\usepackage{microtype}
\usepackage{inconsolata}

\usepackage{graphicx}
\usepackage{booktabs}
\usepackage{amsmath}
\usepackage{amssymb}
\usepackage{multirow}

\usepackage{fvextra}
\usepackage[most]{tcolorbox}
\usepackage{url}
\usepackage[export]{adjustbox}

\newtcolorbox{promptbox}[1]{%
  enhanced,
  colback=white, colframe=black,
  coltitle=white, colbacktitle=black,
  fonttitle=\bfseries\sffamily, title={#1},
  boxrule=0.5pt, arc=0pt,
  left=6pt, right=6pt, top=6pt, bottom=6pt,
}

\newcommand{\finding}[1]{%
  \par\vspace{2pt}\noindent\fbox{\parbox{0.97\columnwidth}{\small\textbf{Finding.} #1}}\par\vspace{3pt}}

\title{Token Distribution versus Data Volume: Domain Balancing in Multi-Domain Meeting Summarisation}

\author{Ashima Sood \quad Bryan Gardiner \quad Joan Condell \\
  School of Computing, Engineering and Intelligent Systems, Ulster University \\ Londonderry, Northern Ireland, United Kingdom \\
  \texttt{\{sood-a1, b.gardiner, j.condell\}@ulster.ac.uk}}

\begin{document}
\maketitle

\begin{abstract}
Jointly fine-tuning an LLM on meeting-summarisation corpora of widely varying size raises a question that prior work leaves confounded: when a domain-balanced training mixture helps, is the gain due to the distribution of tokens across domains, or merely to the volume of data seen? We disentangle these factors by constructing balanced and natural (native-proportional) token mixtures at matched token budgets (2-32M) over five English meeting corpora, fine-tuning Mistral-7B with QLoRA, and evaluating per domain. Balancing redistributes quality, improving the data-scarce minority domains at a low cost to the data-rich ones. The trade favours balancing whenever the minority domains matter: their share under proportional allocation is fixed at 1-2\% regardless of budget, so matching balanced quality on those domains requires far more total data. We further find that pruning low-value transcript lines removes $\sim$15\% of tokens from the conversational corpora at no measurable cost, and that balancing by tokens is not the same as balancing by examples. Fine-tuning one model per domain is competitive only on the data-rich domains and falls below the zero-shot model on the data-scarce ones. A two-annotator study of 741 judge-labelled facts validates our fact-level evaluation. Together these results give practitioners a basis for deciding when to balance an imbalanced multi-domain mixture, and on what unit.
\end{abstract}

\section{Introduction}
Meeting summarisation is increasingly handled by large language models \cite{fu-etal-2024-tiny,laskar-etal-2024-query,kirstein-etal-2025-whats}, but real deployments must support multiple meeting domains such as project meetings, academic discussions, parliamentary sessions, and others that differ by orders of magnitude in the amount of available training data. When such corpora are pooled to jointly fine-tune a single model, the larger domains dominate the training signal, and the smaller domains are drowned out. A natural instinct is to balance the training mixture so that each domain contributes comparably \cite{li2025datamixingsft}.

However, whether balancing actually helps is difficult to establish, as it entangles two effects. To give data-scarce domains a larger share within a fixed training budget, a balanced mixture reallocates tokens away from data-rich domains; granting scarce domains more data outright instead changes the total volume. Any change in quality could therefore stem from the distribution of tokens across domains or from the volume of data, and the two typically move together. Until the two are separated, the central question cannot be answered: is it the token distribution that helps, or the amount of data? The distinction matters wherever models are built from heterogeneous datasets,
where imbalance is the norm rather than the exception.

This paper separates them through a controlled analytical study. Holding data volume fixed and varying only the token distribution, we examine how domain balancing affects per-domain summarisation quality: when it helps, for which domains, and at what budget. We study five English meeting corpora spanning project, academic meetings, parliamentary, and municipal proceedings, whose training sizes are severely imbalanced. We keep volume constant by constructing the natural (native-proportional) and balanced distributions at the same token budgets, across a ladder of $2$, $4$, $8$, $16$, and $32$M tokens, so that a balanced-versus-natural comparison at a matched budget isolates the distribution effect. We fine-tune Mistral-7B-Instruct-v0.3 (Mistral-7B) and evaluate per domain. 
We release all fine-tuned adapters, including the per-domain baselines.\footnote{\url{https://huggingface.co/soodashima91/meeting-summarization-domain-balancing}}

\paragraph{Research Questions.} We organise the study around six research questions.
\begin{itemize}
\itemsep2pt
\item \textbf{RQ1 (Distribution).} Independent of data volume, to what extent does domain-balanced token allocation affect summarisation quality relative to the natural distribution?
\item \textbf{RQ2 (Token efficiency).} How does quality scale with the token budget ($2{\to}32$M), and is balancing more valuable at smaller budgets?
\item \textbf{RQ3 (Pruning).} Does removing low-value transcript lines from the training input preserve quality at a reduced token count?
\item \textbf{RQ4 (Unit of balancing).} Does balancing by token count differ from balancing by example count?
\item \textbf{RQ5 (Model scale).} Do the findings hold on a smaller model (Llama-3.2-3B-Instruct) from a different family?
\item \textbf{RQ6 (Joint versus per-domain).} Does jointly fine-tuning one model on a domain mixture outperform fine-tuning a separate model for each domain?
\end{itemize}

\paragraph{Contributions.}
\textbf{(i)} By comparing balanced and natural mixtures at matched token budgets, we separate token distribution from volume and show that balancing redistributes quality, raising minority domains while lowering majority domains, rather than adding it uniformly (RQ1). \textbf{(ii)} We characterise quality across the $2{\to}32$M budget ladder and find that the distribution gap persists at nearly every budget, closing only when natural allocation spends more budget~(RQ2). \textbf{(iii)} We show that pruning low-value transcript lines preserves quality while cutting $\sim$15\% of tokens from the conversational corpora (RQ3). \textbf{(iv)} We find that balancing by token rather than examples weights domains differently as equal meeting counts produce very unequal token amounts (RQ4). \textbf{(v)} We show these trends hold for a smaller model from a different family (Llama-3.2-3B), indicating they are not artifacts of a single backbone (RQ5). (vi) Under an identical protocol, per-domain fine-tuning matches joint training only on the data-rich domains; the data-scarce domains are learned only through joint training (RQ6).

\section{Related Work}

\paragraph{Meeting summarisation.} Abstractive meeting summarisation was first benchmarked on AMI \cite{kraaij2005ami} and ICSI
\citep{1198793}, with early encoder-decoder models such as HMNet \citep{zhu2020hierarchical} and
DialogLM \citep{zhong2022dialoglm} achieving strong single-domain results. Later resources
broadened the setting to longer, more heterogeneous transcripts, including MeetingBank
\citep{hu-etal-2023-meetingbank} and the ELITR Minuting Corpus
\citep{nedoluzhko-etal-2022-elitr}, and QMSum \citep{zhong-etal-2021-qmsum} which spans product, academic and committee meetings and reports both joint and single-domain training, and the AutoMin shared tasks \citep{ghosal2021overview,
ghosal-etal-2023-overview, shinde2025findings} established automatic minuting as a community
benchmark. Its most recent edition reports an unspecified GPT-4 baseline leading the minuting
task, though the organisers attribute this to self-preference in their GPT-based judge, and the
lead does not hold under the reference-based metrics they also report. Recent LLM-based work
targets summarisation quality \citep{fu-etal-2024-tiny} rather than data allocation across
domains; such systems typically train a separate model per dataset, and none addresses the
cross-domain imbalance that arises when one model is jointly fine-tuned on corpora of widely
varying size.

\paragraph{Data mixtures for fine-tuning.} How to weight sources of differing size is a
standard question in mixture construction: proportional allocation samples each source by its
native token mass, uniform allocation gives each an equal share. The trade-off is well studied
at pretraining scale where domain weights are learned \citep{xie2023doremi} or predicted from mixing laws \citep{ye2025data}, but data mixing for supervised fine-tuning remains comparatively
underexplored~\cite{li2025datamixingsft,renduchintala-etal-2024-smart}, and pretraining methods do not transfer directly. The
closest work to ours is budget-constrained instruction tuning: ADAPT~\cite{kadasi2025adapt}
compares uniform (equal-token) and size-proportional sampling under a token budget, the two
schemes we call balanced and natural, and reports diminishing returns beyond a
small fraction of the budget, as we do. But that line of work learns the mixture, whereas we
hold it fixed and use it as an analytical probe: we instantiate both schemes at matched
token budgets, so allocation is the only variable and its effect is not confounded with volume
(Section~\ref{sec:mixtures}), and we mix domains rather than task types, on real meeting corpora
ranging from tens of meetings to several thousand.

\paragraph{Positioning and PEFT.}
We fine-tune with QLoRA~\cite{3666122.3666563} under a fixed compute budget. The closest meeting-specific work, MeetMulti-X~\cite{SOOD2026130428}, benchmarks LLMs across meeting corpora zero-shot, without fine-tuning or a mixture strategy. Outside meetings, \citet{wang2026charting} fine-tune a 7B model with LoRA on five similarly imbalanced scientific disciplines and find that the joint model trails one adapter per discipline on average, with instability concentrated in the data-scarce disciplines. Their smallest discipline has tens of thousands of examples; ours have tens of meetings (RQ6). To our knowledge no prior work fine-tunes LLMs on token-level, domain-balanced mixtures of meeting corpora, or isolates token distribution from volume under matched budgets. No external system is jointly fine-tuned across these corpora, so our comparisons are internal: the difference between two allocation schemes at a fixed budget, which no external number can inform.

\section{Methodology}

We study how the distribution of training data across domains affects multi-domain meeting summarisation, keeping the total amount of data fixed. Because our five corpora vary widely in size, it is hard to tell whether domain balancing itself helps or whether the gains come from other factors it usually gets mixed up with (the amount of data, the input length, and the unit of balancing). To separate these factors, we build controlled training mixtures at a matched token budget and fine-tune a model on each.

\begin{figure*}[htbp]
\centering
\includegraphics[
  width=0.90\textwidth,
  height=0.29\textheight
]{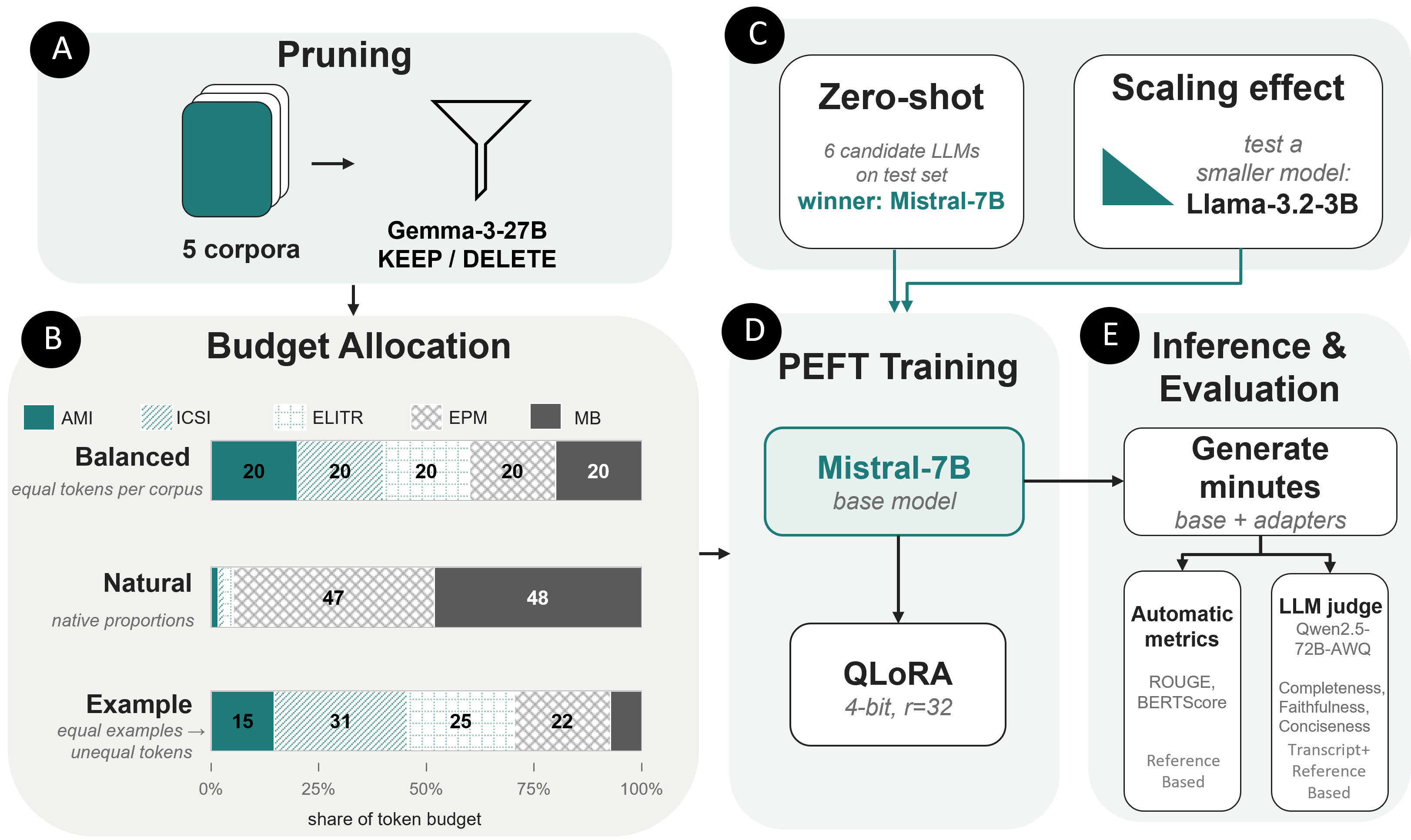}
\caption{Pipeline. \textbf{(A)} Transcripts are pruned of conversational filler with
Gemma-3-27B-it. \textbf{(B)} Training mixtures are built under three allocation
schemes (balanced, natural, example) at a fixed token budget. \textbf{(C)} Zero-shot
screening over six candidate LLMs selects Mistral-7B-Instruct-v0.3; Llama-3.2-3B is
retained as a scaling control. \textbf{(D)} Mistral-7B is fine-tuned with QLoRA.
\textbf{(E)} Generated minutes are scored with automatic metrics and an LLM judge.}
\label{fig}
\end{figure*}

Figure~\ref{fig} provides an overview of our pipeline, summarising the complete experimental workflow. This section describes each part in turn: transcript pruning (§\ref{pruning}), the three budget-allocation schemes (§\ref{sec:mixtures}), baselines and model selection (§\ref{sec:baselines}), fine-tuning (§\ref{sec:training}), inference (§\ref{sec:inference}) and evaluation (§\ref{sec:evaluation}).

\subsection{Pruning}
\label{pruning}
Meeting transcripts contain a large amount of conversational filler, such as greetings, backchannels, and false starts, that adds length without adding content. To reduce input length before training, we apply a pruning that decides, for each transcript line, whether to keep or delete it. This is a line-retention task, not summarisation: we remove noise but keep every line that carries any substantive content.

We prune with gemma-3-27B-it\footnote{\url{https://huggingface.co/google/gemma-3-27b-it}} under greedy decoding, using the same model and prompt for all five corpora (prompt in Appendix~\ref{app:prune-prompt}; detailed statistics in Appendix~\ref{app:pruning-stats}). This model was selected due to its strong instruction-following capability while remaining computationally practical. The prompt sets KEEP as the default and deletes only lines that are pure filler, keeping anything with a decision, action, number, date, name, question, answer, or opinion. To guard against over-deletion, a deterministic rule force-keeps any line containing a digit, a decision or action keyword, or at least twelve content words. Pruning is applied only to the training and development data; test transcripts are always used in full, so every model is evaluated on the same inputs. Pruning removes 6.8\% of training tokens overall, see Appendix Table \ref{tab:pruning-stats} for details. A two-annotator study of 500 lines (100 per corpus, 50 Keep and 50 Delete) validates these decisions against human judgement; inter-annotator agreement is substantial and the pruner tracks annotators on the conversational corpora, over-deleting only on EuroParlMin (Appendix~\ref{app:pruning-validation}).

\subsection{Budget Allocation}
\label{sec:mixtures}

We construct training mixtures by allocating a fixed token budget $B \in \{2, 4, 8, 16, 32\}\text{M tokens}$ across the five corpora and sampling within each corpus to fill its share. The budget fixes the size of the mixture, i.e. the tokens seen per epoch; the number of tokens processed in training additionally depends on the epoch at which early stopping selects the checkpoint (Table~\ref{tab:app-trajectories}). The three schemes differ only in how the budget is split across corpora; they share the same budgets, the same pruned and unpruned variants, and the same prompt, tokeniser, and random seed. This shared construction is what lets a comparison at a matched budget isolate one factor at a time.
Table~\ref{tab:budget-allocation} summarises how the schemes weight the smallest corpus (ICSI) across budgets; full per-corpus, per-budget, and per-seed statistics are in Appendix~\ref{app:mixtures}.

\paragraph{Balanced (equal-token) allocation.} Each corpus receives an equal share of the budget (B/5 or 20\% each). Meeting this quota means subsampling the large corpora, which always have more than enough data. For the small corpora the behaviour depends on the budget: at the smallest budget every meeting is used at most once, but as the budget grows they must be oversampled (their meetings repeated) to fill the equal share. At our largest budget (32M) this reaches roughly $11\times$ for ICSI.

\paragraph{Natural (proportional) allocation.} Each corpus receives a share proportional to its own token mass, so the two large corpora (EPM, MB) together take about 95\% of every budget and the three small corpora only 1-2\% each. These shares fit within each corpus, so natural allocation almost never oversamples. At a matched budget, natural is the volume-controlled counterpart to balanced: comparing the two measures the effect of distribution alone.

\paragraph{Example-level (equal-count) allocation.} Instead of balancing tokens, this scheme balances example count: each corpus contributes an equal number of meetings (43, capped by the smallest corpus), giving about 1.9M tokens in total. Because transcripts differ in length, equal example counts produce very unequal token counts (ICSI contributes 578K tokens, MB only 137K (seed 1)). Comparing this scheme with the balanced one at a similar budget therefore isolates the unit of balancing (tokens versus examples). We use it only for unit comparison (RQ4).

\begin{table}[t]
\centering
\small
\setlength{\tabcolsep}{4pt}
\begin{tabular}{llrrrrr}
\toprule
\textbf{Scheme} & & \textbf{2M} & \textbf{4M} & \textbf{8M} & \textbf{16M} & \textbf{32M} \\
\midrule
\multirow{2}{*}{Balanced}
 & tokens & 397K & 794K & 1.6M & 3.2M & 6.4M \\
 & os & $1.0\times$ & $1.4\times$ & $2.8\times$ & $5.6\times$ & $11.1\times$ \\
\midrule
\multirow{2}{*}{Natural}
 & tokens & 26K & 51K & 105K & 215K & 436K \\
 & os  & $1.0\times$ & $1.0\times$ & $1.0\times$ & $1.0\times$ & $1.0\times$ \\
\bottomrule
\end{tabular}
\caption{Token allocation and oversampling (os) for the smallest corpus (ICSI).
Balanced holds ICSI at a ${\sim}20$\% budget share throughout; natural gives it
${\sim}1.4$\%. Per-corpus breakdowns in Appendix~\ref{app:mixtures}.}
\label{tab:budget-allocation}
\end{table}

\subsection{Baselines and Model Selection}
\label{sec:baselines}
Before fine-tuning, we screen candidate base models zero-shot to choose a primary model. We prompt five instruction-tuned open models and one closed model to summarise the raw test transcripts, using the same prompt and decoding settings for all, and score their outputs with our automatic metrics. The six candidates span a range of sizes and families: Mistral-7B-Instruct-v0.3, Llama-3.1-8B-Instruct, Llama-3.2-3B-Instruct, Qwen3-8B\footnote{Default non-thinking mode suggested settings.}, gemma-4-12B-it from HuggingFace\footnote{\url{https://huggingface.co/models}} and one closed model, GPT-4o\footnote{We used the gpt-4o-2024-11-20 model.}. GPT-4o ranks last of the six on macro BERTScore-F1, its minutes padded with markdown structure our references do not contain (15-26\% of lines), which on MeetingBank yields the highest recall of any candidate alongside the lowest precision (Appendix~\ref{app:model-selection}). Mistral-7B leads on both micro and macro BERTScore-F1, so we select it as our primary base model for all subsequent experiments. We additionally keep the smaller Llama-3.2-3B as a scaling control, to test whether our findings hold at a lower model capacity (RQ5). The full zero-shot comparison across all six candidates is in Appendix~\ref{app:model-selection}.

\paragraph{Per-domain fine-tuned baselines.} Joint fine-tuning assumes that a single model must serve all domains. The alternative is to fine-tune one adapter per corpus. We therefore train five adapters, each on the full training set of one corpus, and evaluate each on the test meetings of its own domain. All five runs follow the protocol of §3.4 unchanged, with early stopping on the development loss of the corresponding domain, so the training data is the only variable relative to the mixture runs. Each adapter processes its corpus once per epoch, i.e. 0.66M (AMI), 0.58M (ICSI), 0.94M (ELITR), 19.7M (EPM) and 20.4M (MB) tokens, compared with 6.4M tokens per domain under balanced allocation at 32M. This baseline establishes whether joint fine-tuning, under either allocation scheme, improves on fine-tuning each domain in isolation (RQ6; Appendix~\ref{app:perdomain}).

\subsection{Fine-tuning}
\label{sec:training}
We fine-tune Mistral-7B-Instruct-v0.3 with QLoRA, which back-propagates through a frozen 4-bit base into low-rank adapters. Beyond cost, this suits our design: with the base model fixed and every run trained under identical hyperparameters and different training mixtures. Loss is cross-entropy over summary tokens, with prompt tokens masked so that optimisation targets generation rather than prompt memorisation. A single fixed prompt, instructing the model to ground minutes only in the transcript, is applied identically across all domains, budgets, and conditions (Appendix~\ref{app:summ-prompt}).

\subsection{Inference}
\label{sec:inference}
At test time we generate with greedy decoding and a fixed generation budget of $\texttt{max\_new\_tokens}=2048$, applied identically across all models and conditions. This budget covers every test reference in AMI, ICSI, ELITR, and MeetingBank in full, and $98.8\%$ of test references overall; the few longer references are undivided full-chapter documents in EuroParlMin rather than typical per-session minutes (Appendix~\ref{app:summary-lengths}). Test transcripts are never pruned, so zero-shot and fine-tuned conditions see identical inputs. For ELITR, which supplies multiple reference minutes per meeting, we score each generated minute as the maximum over its references, so ELITR contributes 38 meetings to evaluation (Appendix~\ref{app:corpus}).

\subsection{Evaluation}
\label{sec:evaluation}
We evaluate generated minutes with automatic metrics and a fact-level LLM judge.

\paragraph{Automatic metrics.} We report ROUGE-1/2/L/Lsum \cite{lin2004rouge} and BERTScore-F1 \cite{zhang2019bertscore}, computing each per domain and then aggregating two ways. The macro average weights the five domains equally (the mean of per-domain scores); the micro average weights every meeting equally (the mean over all test meetings). The distinction is central to our analysis because the test sets are as imbalanced as the training sets: MeetingBank (862 meetings) and EuroParlMin (242) together account for 94\% of test meetings, against 38 for ELITR, 20 for AMI and 6 for ICSI. The micro average is therefore dominated by the majority domains, while the macro average gives equal weight to the data-scarce domains where domain balancing has its largest effect, at the cost of inheriting the wide confidence intervals of their small test sets. We report both averages throughout, with 95\% bootstrap confidence intervals where relevant, and read per-domain results before either aggregate.

\paragraph{Fact-level LLM judge.} Automatic overlap metrics can miss factual correctness, so we additionally evaluate with an LLM-as-judge protocol adapted from \cite{zhou2026large} using an open judge model (Qwen2.5-72B-Instruct-AWQ\footnote{\url{https://huggingface.co/Qwen/Qwen2.5-72B-Instruct-AWQ}}), chosen from a different model family than our summarisers to avoid self-preference bias. The judge decomposes each summary into atomic facts and scores three dimensions: completeness, faithfulness, and conciseness. Full prompts and protocol are in Appendix~\ref{app:judge-prompts} and Appendix~\ref{judgeprotocol} respectively.

\paragraph{Human validation of the judge.} To check that the judge's per-fact labels track human judgement, two annotators independently re-labelled 741 of its decisions on the balanced-32M system (30 test meetings, six per domain; facts stratified by judge label). Inter-annotator agreement is substantial on all three dimensions (Cohen's $\kappa = 0.79$-$0.84$), and the judge agrees with the annotators at $\kappa = 0.71$--$0.76$ on completeness and $\kappa = 0.57$-$0.63$
on faithfulness and conciseness. The errors are one-sided where both annotators agree, the judge over-accepts facts in 87-94\% of disagreements, so the absolute scores it reports are best read as upper bounds. Protocol, annotation guidelines, and full agreement results are in Appendix~\ref{app:human-eval}.

\section{Experiments}

\subsection{Corpora}
\label{sec:corpora}

The work is done on five English meeting-summarisation corpora that vary in genre, scale,
transcription provenance, and reference convention. AMI \citep{kraaij2005ami} (scenario-driven design
meetings), ICSI \citep{1198793} (academic research meetings), MeetingBank (MB)
\citep{hu-etal-2023-meetingbank} (ASR-transcribed municipal proceedings), the ELITR Minuting
Corpus \citep{nedoluzhko-etal-2022-elitr} (technical project meetings), and EuroParlMin (EPM)
\citep{ghosal-etal-2023-overview} (parliamentary debate). They span more than two orders of
magnitude in scale (Table~\ref{tab:corpus-sizes}), reproducing the domain imbalance that
motivates our study. ELITR supplies multiple references per meeting; we consolidate each to a single reference for training and take the maximum over all references at test time. Per-corpus construction, splits, units, and the consolidation
procedure are given in Appendix~\ref{app:corpus}.

\subsection{Experimental settings}
\label{sec:expsettings}
The main experimental grid is $2\times2\times5$ (allocation scheme $\times$ pruning condition $\times$ token budget), giving 20 runs (\S\ref{sec:mixtures}). All grid runs use seed~$42$. We repeat both schemes under the pruned condition at 2M, 8M, and 32M with two further seeds ({2,15}), confirming the per-domain pattern is not seed-specific (12 runs; Appendix~\ref{app:bnseeds}). To this we add the example-level baseline (\S\ref{sec:mixtures}), run under both pruning conditions across three seeds ($\{1,2,15\}$; Appendix~\ref{app:seeds}) and reported as mean\,$\pm$\,std, for 6 runs. Unlike the token-budget schemes, it draws only 43 meetings per corpus from corpora that supply far more, so the particular draw affects both which meetings enter the mixture and its resulting token mass (Appendix Table~\ref{tab:app-example}); the seeds quantify that sensitivity rather than leaving it unmeasured. Finally, the Llama-3.2-3B scaling control (\S\ref{sec:baselines}) repeats the pruned condition of both schemes at the 2M, 8M, and 32M budgets, for 6 runs.

\subsection{Implementation Details}
\label{sec:implementation}
All runs use QLoRA with 4-bit NF4 quantisation and low-rank adapters of rank $r{=}32$ and scaling $\alpha{=}16$, applied to all seven attention and MLP projections (q, k, v, o, gate, up, down). This trains $\approx$83.9M parameters, $1.14\%$ of the 7B base model. Hyperparameters (Table~\ref{tab:hparams}) are held fixed across every scheme, budget, and condition, so that differences across runs reflect the training data rather than the training setup. We train for up to 5 epochs with early stopping on the development loss (patience 2) and retain the lowest-development-loss checkpoint; per-run loss curves and the selected epoch for each run are in Appendix~\ref{app:trajectories}. Each run uses a single NVIDIA A100-SXM4-80GB. The QLoRA formulation is given in Appendix~\ref{app:hparams}.

\section{Results and Discussion}
\label{sec:results}

We report the findings in RQ order. Our corpora are highly imbalanced, so aggregate
averages are majority-weighted and can hide the effect of allocation on the smaller
domains. We therefore read every result per domain and treat macro and micro
averages as summaries rather than as primary evidence.

\subsection{RQ1: Domain Balancing Redistributes Quality Across Domains}
\label{sec:rq1}
\finding{At matched token budgets, balancing does not lift every domain; it trades between them. It improves all three minority domains (AMI, ICSI, ELITR) on both metrics. The fact-level judge shows the same pattern.}

Table~\ref{tab:rq1-32m} contrasts the two schemes per domain at 32M. Balanced leads every minority domain on both metrics, and the gains are substantial: in ROUGE-Lsum, AMI $+0.102$, ICSI $+0.164$, ELITR $+0.163$. On the majority side the movement is smaller and one-sided: MB shifts to natural by $0.046$, while EPM is close. The same per-domain split holds across the full budget ladder (Table~\ref{tab:app-full-pruned}; Figure~\ref{fig:rq2}). The two aggregates therefore disagree by construction: the macro average, weighting domains equally, favours balanced (ROUGE-Lsum 0.500 vs.\ 0.421; BERTScore-F1 0.877 vs.\ 0.872), while the micro average, dominated by MB and EPM, favours natural (0.637 vs.\ 0.615). The effect is visible only once the domains are read separately.

\begin{table}[!tbp]
\centering
\small
\setlength{\tabcolsep}{4pt}
\resizebox{\columnwidth}{!}{%
\begin{tabular}{lccccc cc}
\toprule
Scheme & AMI & ICSI & ELITR & EPM & MB & Macro & Micro \\
 & \textit{20} & \textit{6} & \textit{38} & \textit{242} & \textit{862} & & \\
\midrule
\multicolumn{8}{l}{\textit{ROUGE-Lsum}} \\
Balanced & \textbf{0.495} & \textbf{0.445} & \textbf{0.367} & \textbf{0.547} & 0.648 & \textbf{0.500} & 0.615 \\
Natural  & 0.393 & 0.281 & 0.204 & 0.534 & \textbf{0.694} & 0.421 & \textbf{0.637} \\
\midrule
\multicolumn{8}{l}{\textit{BERTScore-F1}} \\
Balanced & \textbf{0.875} & \textbf{0.852} & \textbf{0.846} & 0.887 & 0.928 & \textbf{0.877} & 0.915 \\
Natural  & 0.865 & 0.833 & 0.837 & \textbf{0.889} & \textbf{0.938} & 0.872 & \textbf{0.923} \\
\bottomrule
\end{tabular}%
}
% \caption{RQ1: balanced vs.\ natural allocation per domain at the matched 32M budget (pruned). Bold marks the higher-scoring scheme, italics the test-set meeting counts. Macro weights domains equally, micro weights meetings. Full grid in
% Table~\ref{tab:app-full-pruned}.}
\caption{RQ1: balanced vs.\ natural allocation per domain at the matched 32M budget (pruned, seed~42). Bold marks the higher-scoring scheme, italics the test-set meeting counts. Macro weights domains equally, micro weights meetings. Full grid in Table~\ref{tab:app-full-pruned}; three-seed means in Table~\ref{tab:multiseed-all}.}
\label{tab:rq1-32m}
\end{table}

The fact-level judge (Table~\ref{tab:judge}) reproduces this at both budgets. On completeness, the recall of reference key facts, balanced leads on macro ($+0.017$ at 2M, $+0.041$ at 32M) and trails on micro ($-0.016$, $-0.015$), the same sign flip as in the automatic metrics, now measured over facts rather than surface overlap. Faithfulness is comparable between the schemes but falls as the budget grows  (balanced $0.765\!\rightarrow\!0.742$, natural $0.739\!\rightarrow\!0.728$), most sharply on EPM, where longer generations draw on transcript beyond our 16K truncation. Natural is more concise at 32M (0.506 vs.\ 0.472 macro): balanced recovers more reference content at a cost in per-fact
salience.

\begin{table}[!tbp]
\centering
\resizebox{\columnwidth}{!}{%
\begin{tabular}{lccc}
\toprule
\textbf{System} & \textbf{Completeness} & \textbf{Faithfulness} & \textbf{Conciseness} \\
\midrule
\multicolumn{4}{l}{\textit{Macro}} \\
balanced 2M   & $\mathbf{0.276_{[.25,.30]}}$ & $\mathbf{0.765_{[.74,.79]}}$ & $\mathbf{0.515_{[.48,.54]}}$ \\
natural 2M    & $0.259_{[.24,.28]}$ & $0.739_{[.70,.78]}$ & $0.467_{[.43,.51]}$ \\
balanced 32M  & $\mathbf{0.373_{[.35,.40]}}$ & $\mathbf{0.742_{[.73,.76]}}$ & $0.472_{[.45,.49]}$ \\
natural 32M   & $0.332_{[.31,.36]}$ & $0.728_{[.70,.75]}$ & $\mathbf{0.506_{[.46,.56]}}$ \\
\midrule
\multicolumn{4}{l}{\textit{Micro}} \\
balanced 2M   & $0.427_{[.41,.45]}$ & $\mathbf{0.707_{[.69,.73]}}$ & $0.573_{[.55,.59]}$ \\
natural 2M    & $\mathbf{0.443_{[.42,.46]}}$ & $0.685_{[.67,.70]}$ & $\mathbf{0.575_{[.55,.60]}}$ \\
balanced 32M  & $0.507_{[.49,.53]}$ & $\mathbf{0.642_{[.62,.66]}}$ & $0.589_{[.57,.61]}$ \\
natural 32M   & $\mathbf{0.522_{[.50,.54]}}$ & $0.636_{[.62,.66]}$ & $\mathbf{0.625_{[.61,.64]}}$ \\
\bottomrule
\end{tabular}%
}
\caption{Fact-level LLM-judge scores at the smallest and largest budgets, with
bootstrap 95\% CIs (10{,}000 resamples). Balanced leads on macro completeness and
trails on micro, the same sign flip the automatic metrics show. Per-domain breakdown
in Table~\ref{tab:judge-full}.}
\label{tab:judge}
\end{table}

\subsection{RQ2: Token efficiency and the budget dependence of balancing}
\label{sec:rq2}

\finding{Balancing's minority-domain advantage does not close with budget. Proportional allocation fixes their share at 1-2\%, so at 32M it gives ICSI barely more tokens than balancing gives it at 2M.}

\begin{figure}[t]
  \centering
  \includegraphics[width=0.99\linewidth]{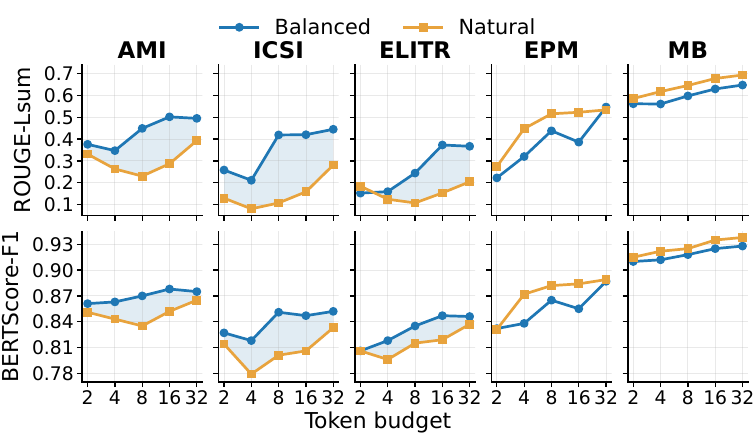}
  \caption{RQ2: Per-dataset quality across the token budget ladder
(2$\rightarrow$32M) for balanced (blue) and natural (orange) allocation, on
ROUGE-Lsum (top) and BERTScore-F1 (bottom). Shading marks balanced's advantage on
the minority panels. Quality rises with budget and flattens; the minority gap is
open at nearly every budget and does not close by 32M.}
  \label{fig:rq2}
\end{figure}

Figure~\ref{fig:rq2} plots per-dataset  quality across the budget ladder. Both metrics show the diminishing-returns shape of a token-efficiency curve, rising with budget and flattening at the top, most clearly in BERTScore-F1. In the minority domains balanced climbs with the budget and leads natural at nearly every budget, while natural stays low and flat, held down by the $\sim$1-2\% share proportional allocation assigns those domains at every budget.

As the budget grows, natural routes more absolute tokens to the scarce domains and closes part of the gap, but does not catch balanced: at 32M it still trails on every minority domain. The asymmetry is stark in absolute terms. Proportional allocation gives ICSI 436K tokens at 32M, barely more than the 397K balanced already assigns it at 2M (Table~\ref{tab:budget-allocation}), so natural must spend sixteen times the budget to buy the same exposure. On the majority side the schemes differ: on MB natural stays ahead throughout, while on EPM natural's early lead narrows with budget and at 32M the two are indistinguishable, the seed-42 difference (0.547 vs. 0.534 on ROUGE-Lsum) lying within seed variation (Table~\ref{tab:multiseed-all}). Balanced allocation is therefore attractive whenever the minority domains matter, since natural can close the gap only by spending budget it may not have.

\subsection{RQ3: Pruning preserves quality at fewer tokens}
\label{rq3}

\finding{Pruning low-value transcript lines removes 6.8\% of training tokens, most of it from the conversational corpora, while preserving quality: at matched budgets, pruned and unpruned runs are near-identical on BERTScore-F1 and converge on ROUGE-Lsum as the budget grows.}

We compare pruned and unpruned training at matched budgets. Equal token counts give equal compute (FLOPs differ by 0.04\% on average), so any quality difference reflects what the tokens contain rather than how many there are. Per-corpus reductions are in Appendix~\ref{app:pruning-stats}.

\begin{figure}[t]
  \centering
  \includegraphics[width=0.99\linewidth]{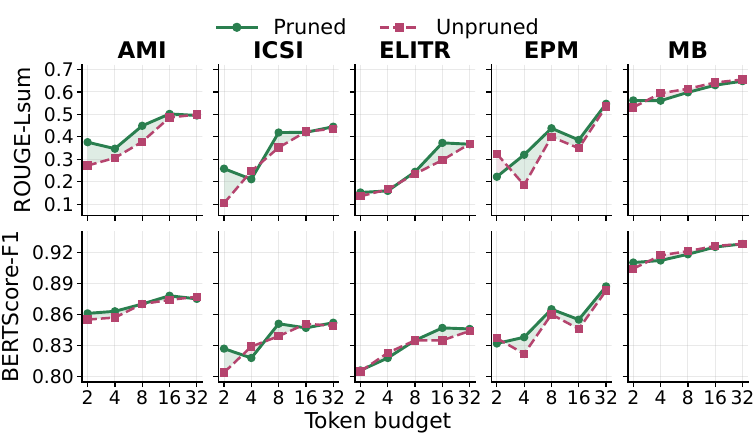}
  \caption{RQ3: Pruned vs.\ unpruned training quality per dataset across the budget
ladder (balanced allocation), on ROUGE-Lsum (top) and BERTScore-F1 (bottom). The two
conditions are near-identical on BERTScore-F1 and converge on ROUGE-Lsum as the
budget grows; the visible gaps are confined to the smallest budgets.}
  \label{fig:rq3}
\end{figure}

Figure~\ref{fig:rq3} plots the two conditions across the budget ladder. On BERTScore-F1 the curves are near-identical throughout, so removing the conversational fillers leaves semantic adequacy intact. In ROUGE-Lsum they track closely at the mid and large budgets and meet by 32M (mean absolute difference under 0.01); the gaps at 2M and 4M, running in both directions across ICSI, AMI, and EPM, are variance rather than a cost of pruning.

Pruning also interacts with balancing. Pruned transcripts are 7-16\% shorter, so a fixed budget spans 8-19\% more meetings. Where the balanced mixture oversamples the minority corpora heavily (up to $11\times$ for ICSI; Table~\ref{tab:app-balanced}), the shorter transcripts fill the same quota with less repetition and a higher share of distinct content, so pruning is most useful exactly where balancing is most aggressive.

Our matched-budget design cannot show this saving, since equalising tokens equalises compute by construction. The saving appears when the corpus rather than the budget is fixed: pruning then trains on 6.8\% fewer tokens for proportionally less compute, at no measurable cost.

\subsection{RQ4: The unit of balancing matters}
\label{sec:rq4}

\finding{Tokens, not examples, are the unit that matters. Equal meeting counts hand ICSI four times MB's token mass, and per-domain quality follows the tokens, moving the two domains in opposite directions. Balancing on examples therefore allocates by accident; balancing on tokens allocates by design.}

Example-level allocation gives every corpus the same number of meetings (43, capped by ICSI). But transcripts vary in length, so equal meeting counts turn into very unequal token mass: ICSI brings 578K tokens and MB only 137K (seed 1), a $\sim$4$\times$ gap (Table~\ref{tab:app-example}). Token-balancing does the reverse, holding every corpus to the same token share.

The difference shows up exactly where transcript lengths are most extreme (Table~\ref{tab:results-unit}). MB has short transcripts, is starved of tokens under equal example counts, and drops accordingly (0.505 vs.\ 0.562), a change well outside its seed variance ($\pm 0.008$). ICSI has long transcripts, so counting by meetings hands it far more tokens than token-balancing would, and it moves in the opposite direction (0.292 vs.\ 0.258); with only six test meetings its seed variance is large ($\pm 0.036$), so we read the direction rather than the magnitude. The two domains move opposite ways on the same data. Both moves track token mass, not meeting counts. On BERTScore-F1 the gap is small, so the unit mainly moves surface overlap rather than semantic adequacy.

ICSI's higher score under example-level balancing is therefore not a point in that scheme's favour, but rather transcript length leaking into the allocation. Token-balancing instead assigns each domain the same budget by construction, which is the control our matched-budget study
relies on: because example counts do not correspond to a fixed token budget, the
RQ1-RQ3 comparisons require balancing on tokens.

% \begin{table}[t]
% \centering
% \resizebox{\columnwidth}{!}{%
% \setlength{\tabcolsep}{4pt}
% \begin{tabular}{lcccccc}
% \toprule
% \textbf{Scheme ($\sim$2M)} & \textbf{AMI} & \textbf{ICSI} & \textbf{ELITR} & \textbf{EPM} & \textbf{MB} & \textbf{Macro} \\
%  & \textit{20} & \textit{6} & \textit{38} & \textit{242} & \textit{862} & \\
% \midrule
% \multicolumn{7}{l}{\textit{ROUGE-Lsum}} \\
% Balanced       & $0.376$ & $0.258$ & $0.152$ & $0.222$ & $\textbf{0.562}$ & $0.314$ \\
% Natural        & $0.331$ & $0.129$ & $0.184$ & $0.274$ & $0.586$ & $0.301$ \\
% Example-level  & $0.328_{\pm0.044}$ & $\mathbf{0.292_{\pm0.036}}$ & $0.229_{\pm0.014}$ & $0.271_{\pm0.044}$ & $0.505_{\pm0.008}$ & $0.325_{\pm0.017}$ \\
% \midrule
% \multicolumn{7}{l}{\textit{BERTScore-F1}} \\
% Balanced       & $0.861$ & $0.827$ & $0.806$ & $0.832$ & $\textbf{0.916}$ & $0.847$ \\
% Natural        & $0.851$ & $0.814$ & $0.806$ & $0.831$ & $0.915$ & $0.843$ \\
% Example-level  & $0.860_{\pm0.004}$ & $\mathbf{0.834_{\pm0.003}}$ & $0.826_{\pm0.006}$ & $0.833_{\pm0.013}$ & $0.898_{\pm0.000}$ & $0.850_{\pm0.004}$ \\
% \bottomrule
% \end{tabular}%
% }
% \caption{RQ4: Example-level balancing scores. Example-level (43 meetings per corpus) is mean$\pm$std over seeds \{1,2,15\}. Equal meeting counts give ICSI $\sim$578K tokens but MB only $\sim$137K. Full results in Appendix Table \ref{tab:app-ex-full}}
% \label{tab:results-unit}
% \end{table}

\begin{table}[t]
\centering\scriptsize\setlength{\tabcolsep}{0.8pt}
\begin{tabular}{@{}lcccccc@{}}
\toprule
\textbf{Scheme} & \textbf{AMI} & \textbf{ICSI} & \textbf{ELITR} & \textbf{EPM} & \textbf{MB} & \textbf{Macro} \\
 & \textit{20} & \textit{6} & \textit{38} & \textit{242} & \textit{862} & \\
\midrule
\multicolumn{7}{@{}l}{\textit{ROUGE-Lsum}} \\
Balanced   & .376 & .258 & .152 & .222 & \textbf{.562} & .314 \\
Natural    & .331 & .129 & .184 & .274 & .586 & .301 \\
Example    & .328$_{\pm.044}$ & \textbf{.292}$_{\pm.036}$ & .229$_{\pm.014}$ & .271$_{\pm.044}$ & .505$_{\pm.008}$ & .325$_{\pm.017}$ \\
\midrule
\multicolumn{7}{@{}l}{\textit{BERTScore-F1}} \\
Balanced   & .861 & .827 & .806 & .832 & \textbf{.916} & .847 \\
Natural    & .851 & .814 & .806 & .831 & .915 & .843 \\
Example    & .860$_{\pm.004}$ & \textbf{.834}$_{\pm.003}$ & .826$_{\pm.006}$ & .833$_{\pm.013}$ & .898$_{\pm.000}$ & .850$_{\pm.004}$ \\
\bottomrule
\end{tabular}
\caption{RQ4: Example-level balancing at $\sim$2M. Example-level (43 meetings per corpus) is mean$\pm$std over seeds \{1,2,15\}; Balanced and Natural are seed 42. Equal meeting counts give ICSI $\sim$578K tokens but MB only $\sim$137K. Full results in Appendix Table~\ref{tab:app-ex-full}.}
\label{tab:results-unit}
\end{table}

% \begin{table}[t]
% \centering\scriptsize\setlength{\tabcolsep}{2pt}
% \begin{tabular}{lcccccc}
% \toprule
% \textbf{Scheme ($\sim$2M)} & \textbf{AMI} & \textbf{ICSI} & \textbf{ELITR} & \textbf{EPM} & \textbf{MB} & \textbf{Macro} \\
%  & \textit{20} & \textit{6} & \textit{38} & \textit{242} & \textit{862} & \\
% \midrule
% \multicolumn{7}{l}{\textit{ROUGE-Lsum}} \\
% Balanced       & $0.376$ & $0.258$ & $0.152$ & $0.222$ & $\textbf{0.562}$ & $0.314$ \\
% Natural        & $0.331$ & $0.129$ & $0.184$ & $0.274$ & $0.586$ & $0.301$ \\
% Example-level  & $0.328_{\pm0.044}$ & $\mathbf{0.292_{\pm0.036}}$ & $0.229_{\pm0.014}$ & $0.271_{\pm0.044}$ & $0.505_{\pm0.008}$ & $0.325_{\pm0.017}$ \\
% \midrule
% \multicolumn{7}{l}{\textit{BERTScore-F1}} \\
% Balanced       & $0.861$ & $0.827$ & $0.806$ & $0.832$ & $\textbf{0.916}$ & $0.847$ \\
% Natural        & $0.851$ & $0.814$ & $0.806$ & $0.831$ & $0.915$ & $0.843$ \\
% Example-level  & $0.860_{\pm0.004}$ & $\mathbf{0.834_{\pm0.003}}$ & $0.826_{\pm0.006}$ & $0.833_{\pm0.013}$ & $0.898_{\pm0.000}$ & $0.850_{\pm0.004}$ \\
% \bottomrule
% \end{tabular}
% \caption{RQ4: Example-level balancing scores. Example-level (43 meetings per corpus) is mean$\pm$std over seeds \{1,2,15\}. Equal meeting counts give ICSI $\sim$578K tokens but MB only $\sim$137K. Full results in Appendix Table \ref{tab:app-ex-full}}
% \label{tab:results-unit}
% \end{table}

\subsection{RQ5: Generalisation to a smaller model}
\finding{The redistribution replicates on Llama-3.2-3B. Balancing helps the same three minority domains and costs the same two majority domains on both metrics. Absolute quality is lower, so what transfers is the direction of the effect, not its magnitude.}

An allocation effect is only useful if it survives a change of model and capacity, so we repeat the comparison on Llama-3.2-3B, a different family at half the scale. Across the three shared budgets (2M/8M/32M), Table~\ref{tab:results-rq5} reports the mean balanced$-$natural gap per dataset. The sign agrees on every dataset and both metrics: positive on the minority domains and negative on the majority domains. That the same pattern holds under a different, smaller model indicates the redistribution comes from the data allocation, not from Mistral-7B.

\begin{table}[t]
\centering
\small
\setlength{\tabcolsep}{6pt}
\begin{tabular}{lcccc}
\toprule
& \multicolumn{2}{c}{\textbf{ROUGE-Lsum}} & \multicolumn{2}{c}{\textbf{BERTScore-F1}} \\
\cmidrule(lr){2-3}\cmidrule(lr){4-5}
\textbf{Dataset} & M-7B & L-3B & M-7B & L-3B \\
\midrule
AMI    & $+0.122$ & $+0.155$ & $+0.018$ & $+0.034$ \\
ICSI   & $+0.202$ & $+0.141$ & $+0.027$ & $+0.040$ \\
ELITR  & $+0.089$ & $+0.072$ & $+0.010$ & $+0.032$ \\
EPM    & $-0.039$ & $-0.178$ & $-0.006$ & $-0.031$ \\
MB     & $-0.039$ & $-0.061$ & $-0.007$ & $-0.014$ \\
\bottomrule
\end{tabular}
\caption{RQ5: per-dataset mean balanced$-$natural gap for M-7B (Mistral-7B) and L-3B
(Llama-3.2-3B), averaged over the three shared budgets (2M/8M/32M; pruned, seed 42).
Positive favours balancing. Per-budget numbers in Table~\ref{tab:rq5-numeric}.}
\label{tab:results-rq5}
\end{table}

\subsection{RQ6: Joint versus per-domain fine-tuning}
\finding{Per-domain fine-tuning exceeds balanced allocation on the two data-rich domains and falls below it on the three data-scarce domains. Joint training is what makes the minority domains learnable.}
Table~\ref{tab:perdomain} compares the five per-domain adapters with balanced allocation at 32M. On EPM and MB the per-domain adapters lead on both metrics, and the per-domain system attains the highest micro average in this study (0.647 vs. 0.615). On AMI, ICSI and ELITR the ordering reverses: the per-domain adapters trail by 0.11–0.24 ROUGE-Lsum, with non-overlapping confidence intervals on ICSI and ELITR, and the macro average falls from 0.500 to 0.397. Trained on 43–98 meetings under the fixed protocol, these adapters acquire each corpus's reference format but not its content (Appendix~\ref{app:perdomain-train} for details). Per-domain fine-tuning therefore serves only majority-weighted deployments; balanced joint training remains the single configuration that serves all five domains.

\begin{table}[t]
\centering\small\setlength{\tabcolsep}{3.5pt}
\resizebox{\columnwidth}{!}{%
\begin{tabular}{lccccccc}
\toprule
\textbf{System} & \textbf{AMI} & \textbf{ICSI} & \textbf{ELITR} & \textbf{EPM} & \textbf{MB} & \textbf{Macro} & \textbf{Micro} \\
 & \textit{20} & \textit{6} & \textit{38} & \textit{242} & \textit{862} & & \\
\midrule
\multicolumn{8}{l}{\textit{ROUGE-Lsum}} \\
Per-domain   & 0.382 & 0.210 & 0.126 & \textbf{0.564} & \textbf{0.703} & 0.397 & \textbf{0.647} \\
Balanced 32M & \textbf{0.495} & \textbf{0.445} & \textbf{0.367} & 0.547 & 0.648 & \textbf{0.500} & 0.615 \\
\midrule
\multicolumn{8}{l}{\textit{BERTScore-F1}} \\
Per-domain   & 0.868 & 0.817 & 0.772 & \textbf{0.897} & \textbf{0.939} & 0.858 & \textbf{0.923} \\
Balanced 32M & \textbf{0.875} & \textbf{0.852} & \textbf{0.846} & 0.887 & 0.928 & \textbf{0.877} & 0.915 \\
\bottomrule
\end{tabular}%
}
\caption{RQ6: per-domain fine-tuned baselines vs.\ balanced allocation at 32M (pruned, seed 42). Each per-domain adapter is trained on the full training set of one corpus under the protocol of Table~\ref{tab:hparams} and evaluated on its own domain. Bold marks the higher system per column; italics are test-set meeting counts. Macro weights domains equally, micro weights meetings. Bootstrap CIs in Table~\ref{tab:app-perdomain}.}
\label{tab:perdomain}
\end{table}

\section{Conclusion}
\label{sec:conclusion}

EuroParlMin and MeetingBank account for roughly 95\% of the token mass across our five domains, so under proportional allocation a jointly fine-tuned model is trained almost exclusively on them, and its per-domain quality follows each domain's share of the training signal. Domain balancing removes this dependence: at a matched 32M budget it leads on all three data-scarce domains on both metrics. Balancing thus does not raise quality on every domain but reallocates it across them, and the reallocation is asymmetric. Proportional allocation holds the minority domains at a 1-2\% token share at every budget, and narrows the gap only by increasing their absolute volume; at 32M it has still not matched balancing on any of the three. Two further results bear on how the budget is spent: pruning conversational filler preserves quality at no measurable cost, and the unit of balancing is consequential, as equal example counts assign ICSI four times the token mass of MeetingBank. The effect replicates on a smaller model from a different family.
Fine-tuning each domain in isolation matches joint training only on the data-rich domains; on the data-scarce domains it falls below the untuned model, so joint training is what makes those domains learnable at all.

Imbalance of this severity is characteristic of real multi-domain deployments rather than an artefact of our design. Our findings offer practitioners a basis for allocation decisions: retain the native distribution when the deployment is majority-weighted, balance when all domains must be served, balance by tokens rather than examples, and prune prior to training in either case.

\section*{Limitations}
\label{sec:limitations}
Meeting transcripts are long, and a full transcript with its minute exceeds the
$16{,}384$-token window we can fit on a single device. Truncation falls unevenly
across the corpora: 54\% of ICSI and 33\% of ELITR training transcripts are cut,
against 24\% of EuroParlMin's, but severity runs the other way. ICSI loses 18\%
of a median 15.8K-token transcript, whereas EuroParlMin's truncated transcripts
retain only 65\% of content against a maximum length twelve times the window.
The consequence is a training target not fully supported by its input: the
reference minute covers the whole session while the input is cut, so on
EuroParlMin the model is trained to assert outcomes it cannot observe. Its low
faithfulness on that corpus is consistent with this, though we do not isolate it
as the cause; ICSI, truncated more often but less severely, records the highest
faithfulness of any corpus. At test time the constraint is far milder, affecting
$3.3\%$ of transcripts, so evaluation inputs are near-complete.

Our fact-level judge is validated against two annotators on one system, and its
errors are one-sided: it accepts facts the annotators reject in 87--94\% of
disagreements, so its absolute scores are too high. But every system is scored by
the same judge under the same rubric, so the differences between systems still
hold.

The findings are established up to 7B parameters under low-rank adaptation.
Whether the redistribution effect holds at larger scale or under full fine-tuning
is untested; our cross-family control, Llama-3.2-3B, reproduces the direction of
the effect on every domain.

Our automatic metrics reward agreement with each corpus's reference convention as
well as content coverage, as the zero-shot screening makes visible. This bounds
what the absolute scores mean, though not the balanced-versus-natural
comparisons, which hold the reference form fixed.

All experiments use English meeting transcripts. Whether domain balancing
interacts with language balancing, or transfers to other multi-domain tasks, is
left open.

\section*{Ethics}
\label{sec:ethics}

All corpora are publicly available and used under their released licences. We perform no re-identification and add no annotation beyond schema unification and the reference-selection and pruning steps described. The two annotators for the pruning and judge validation studies are colleagues of the authors with backgrounds in NLP, not involved in developing the model or producing the results. They were compensated at a locally appropriate rate and participated voluntarily. Our pruning and mixture procedures are designed to preserve substantive content, but automatic minuting can omit or distort information and should not be relied on as an authoritative record without human review. We report compute and seed budgets transparently and concentrate repeated runs on cheaper configurations to limit energy use. All fine-tuning and open-model inference run on a single A100, and the fact-level judge on a single H100; the only external cost is the GPT-4o zero-shot baseline, a single closed-model comparison run that incurred approximately USD $\approx$ 15.5 in API charges.

We used LLM-based tools for editorial assistance, such as rewriting for clarity, shortening text, improving grammar, and ensuring stylistic consistency. All experimental design, analyses, results, and claims are the authors' own, and the authors verified all AI-assisted text and take full responsibility for the content.

\section*{Acknowledgments}
The authors would like to thank the anonymous reviewers for the their valuable feedback which helped improve the quality of this paper. In addition, the authors are grateful for access to the Tier 2 High-Performance
Computing resources provided by the Northern Ireland High Performance Computing (NIHPC) facility funded by the UK Engineering and
Physical Sciences Research Council (EPSRC), Grant No. EP/T022175/1.

\bibliography{custom}

% \clearpage
\appendix
%\onecolumn

\section*{Appendix}

\noindent The appendix is organised as follows.

\vspace{6pt}
\noindent
% \begin{tabular}{@{}p{0.7cm}l@{}}
% \toprule
% \multicolumn{2}{@{}l}{\textbf{Appendix Contents}} \\
% \midrule
% \ref{app:pruning-stats}   & Pruning Statistics \\
% \ref{app:pruning-validation}   & Human Validation of Pruning \\
% \ref{app:mixtures}        & Data-Mixture Construction Details \\
% \ref{app:msrobust}           & Multi-Seed Robustness \\
% \ref{app:model-selection} & Model Selection \\
% \ref{app:summary-lengths} & Summary Lengths (Reference and Generated) \\
% \ref{app:corpus}          & Corpus Construction \\
% \ref{app:hparams}         & Fine-tuning Hyperparameters \\
% \ref{app:trajectories}    & Training Trajectories and Checkpoint Selection \\
% \ref{judgeprotocol}       & LLM Judge Protocol \\
% \ref{app:human-eval}      & Human Validation of the LLM Judge \\
% \ref{app:full-results}    & Full Per-Domain Results \\
% \ref{app:prompts}         & Prompts \\
% \bottomrule
% \end{tabular}

\begin{tabular}{@{}p{0.6cm}p{0.8cm}p{5.2cm}@{}}
\toprule
\multicolumn{3}{@{}l}{\textbf{Appendix Contents}} \\
\midrule
\ref{app:pruning-stats}      & & Pruning Statistics \\
\ref{app:pruning-validation} & & Human Validation of Pruning \\
\ref{app:mixtures}           & & Data-Mixture Construction Details \\
\ref{app:truncation}         & & Transcript Length and Truncation \\
\ref{app:msrobust}           & & Multi-Seed Robustness \\
 & \ref{app:bnseeds}         & Balanced and Natural Schemes \\
 & \ref{app:seeds}           & Example-Level Baseline \\
\ref{app:model-selection}    & & Model Selection \\
\ref{app:perdomain}    & & 
Per-Domain Fine-tuned Baselines \\
& \ref{app:perdomain-train}         & Setup and Training Trajectories \\
 & \ref{app:perdomain-results}           & Full Results \\
\ref{app:summary-lengths}    & & Summary Lengths (Reference and Generated) \\
\ref{app:corpus}             & & Corpus Construction \\
\ref{app:hparams}            & & Fine-tuning Hyperparameters \\
\ref{app:trajectories}       & & Training Trajectories and Checkpoint Selection \\
\ref{judgeprotocol}          & & LLM Judge Protocol \\
\ref{app:human-eval}         & & Human Validation of the LLM Judge \\
 & \ref{app:guidelines}      & Annotation Guidelines \\
\ref{app:full-results}       & & Full Per-Domain Results \\
\ref{app:prompts}            & & Prompts \\
 & \ref{app:prune-prompt}    & Pruning Prompt \\
 & \ref{app:summ-prompt}     & Summarisation Prompt \\
 & \ref{app:judge-prompts}   & LLM-as-a-Judge Prompts \\
\bottomrule
\end{tabular}

\section{Pruning Statistics}
\label{app:pruning-stats}
Table~\ref{tab:pruning-stats} reports the effect of transcript pruning on the training data, with token counts measured using the Gemma-3-27B-it tokenizer. Token reduction is strongly domain-dependent: pruning removes $14$--$16\%$ of tokens from the conversational corpora (AMI, ICSI, ELITR), which contain substantial backchannel and filler, but only $2.3\%$ from EuroParlMin, whose formal parliamentary prepared and edited speeches carries little removable noise. The aggregate reduction ($6.8\%$) is dominated by the two largest corpora and understates the savings on the smaller conversational domains. The final column reports the number of lines the deterministic content guard force-kept. The guard overrides the pruning model whenever a line contains a digit, a decision or action keyword, or at least twelve content words, ensuring such lines are never deleted regardless of the model's decision. Across all corpora it force-kept 225,191 lines, bounding the recall of substantive content independently of the pruning model. The full pruning prompt is given in Appendix subsection~\ref{app:prune-prompt}

\begin{table}[h]
\centering
\resizebox{\columnwidth}{!}{%
\small
\setlength{\tabcolsep}{4pt}
\begin{tabular}{lrrrr}
\toprule
\textbf{Corpus} & \textbf{Transcripts} & \textbf{Tokens (pre)} & \textbf{Reduction} & \textbf{Guard-saved} \\
\midrule
AMI    &   118 &    863{,}292 & 15.6\% &   3{,}030 \\
ICSI   &    52 &    771{,}303 & 14.7\% &   2{,}586 \\
ELITR  &    94 &  1{,}164{,}265 & 14.4\% &   3{,}770 \\
MB     & 6{,}030 & 28{,}616{,}375 &  9.8\% &  69{,}493 \\
EPM    & 2{,}252 & 23{,}878{,}795 &  2.3\% & 146{,}312 \\
\midrule
\textbf{Total} & \textbf{8{,}546} & \textbf{55{,}294{,}030} & \textbf{6.8\%} & \textbf{225{,}191} \\
\bottomrule
\end{tabular}%
}
\caption{Transcript pruning on the training data (train${+}$dev; test is unpruned),
Gemma-3-27B-it tokenizer. Reduction is the fraction of tokens removed;
Guard-saved is the number of lines force-retained by the content guard.}
\label{tab:pruning-stats}
\end{table}

\section{Human Validation of Pruning}
\label{app:pruning-validation}
\paragraph{Setup}  We validate the pruning decisions against two annotators, mirroring the judge-validation design. Pruning is applied to the training split, where KEEP dominates, so a uniform sample would leave few DELETE items on the majority corpora and let a constant-KEEP annotator agree by default. We therefore sampled 50 KEEP and 50 DELETE lines per corpus (500 items total), drawing each stratum across many transcripts so no single meeting dominates, and restricting KEEP items to non-guard lines—those the deterministic content guard did not force-keep—so that agreement reflects the pruning model's judgement rather than the guard. Each line was shown in ±4 lines of context with the model's decision hidden; the annotated line is the one judged, the context is reference only. Both annotators labelled every item independently, and items were shuffled so the KEEP/DELETE composition was not visible. The following annotation guidelines were given verbatim to both annotators.

\begin{quote}\small
\textbf{Your task.} Each row of the spreadsheet shows one meeting-transcript line. You decide whether that line should be kept or deleted from the transcript. The goal is to remove pure conversational noise while keeping anything with real content. In the \texttt{human\_decision} column, write \texttt{Keep} or \texttt{Delete} for each row.

\textbf{Which line to judge.} Judge only the target line -- the one marked with \texttt{>>>} in the \texttt{context} column (also shown in \texttt{target\_line}). The other lines in \texttt{context} are there only to help you understand the target line. Do not mark them.

\textbf{How to decide.} Delete a line only if it is pure conversational noise with no standalone content:
\begin{itemize}\itemsep0pt \small
  \item bare acknowledgements / fillers: ``okay'', ``right'', ``yeah'', ``mm-hmm'', ``uh'', ``cool'';
  \item greetings, goodbyes, thanks, social pleasantries;
  \item connection/logistics chatter: ``can you hear me?'', ``you're muted'';
  \item meaningless fragments or false starts that carry no information.
\end{itemize}
Keep a line if it contains any real content, even briefly:
\begin{itemize}\itemsep0pt \small
  \item a topic, fact, opinion, reason, problem, suggestion, or description;
  \item a decision, agreement, action item, deadline, number, price, name, or date;
  \item a question, or an answer to one --- even a short one (``Friday'', ``Yes'', ``25 euros'').
\end{itemize}

\textbf{Rules.}
\begin{itemize}\itemsep0pt \small
  \item Keep is the default. Only delete when you are confident the line is pure noise.
  \item Use the context to judge short lines. ``Friday.'' is Keep if it answers a question, Delete if it is just filler.
  \item When unsure, Keep.
  \item Do not confer with the other annotator; make every call yourself.
  \item Do not leave any row blank.
\end{itemize}
\end{quote}

\begin{figure}[t]
  \centering
  \includegraphics[width=0.99\linewidth]{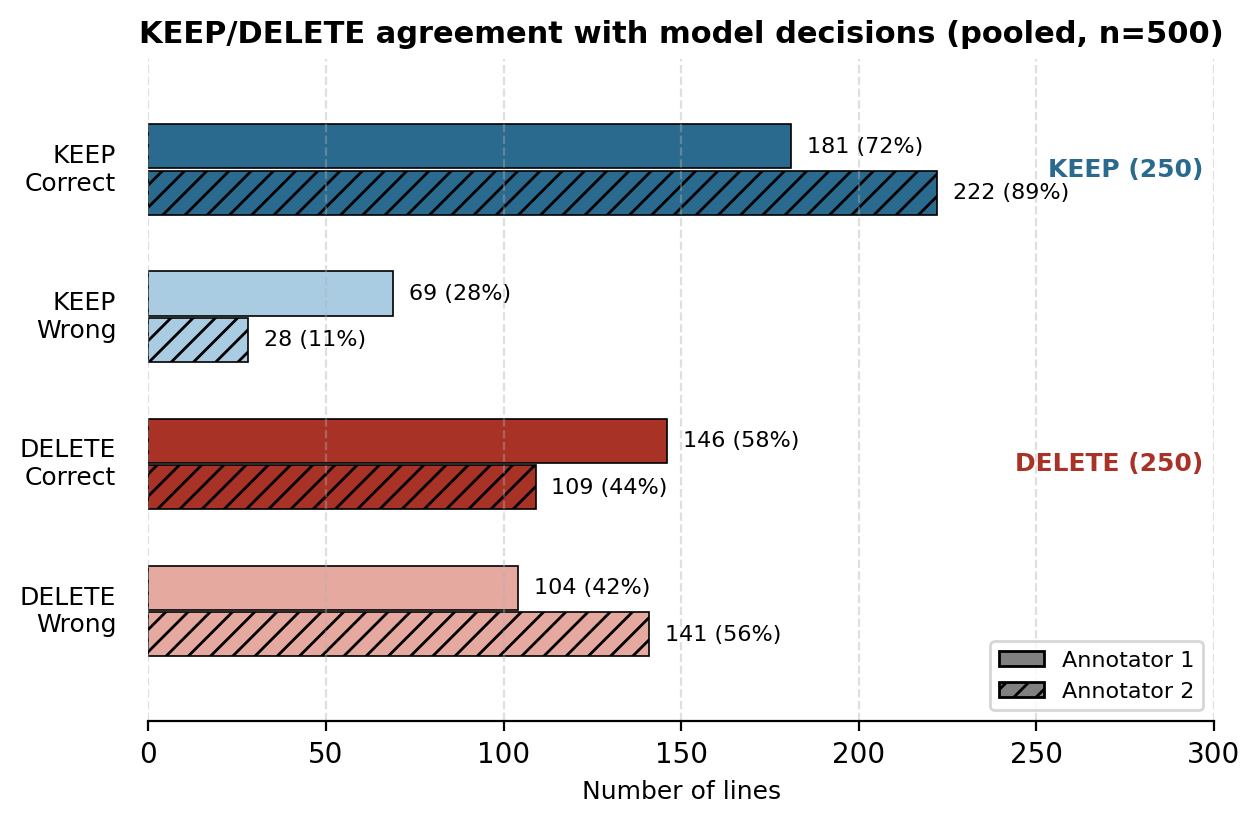}
  \caption{Annotator ratification of the pruner's KEEP/DELETE decisions, pooled over 500 sampled lines (250 KEEP, 250 DELETE; 100 per corpus). Bars show how many of the model's KEEP and DELETE calls each annotator judged correct vs. wrong.}
  \label{pruning_validation}
\end{figure}

\paragraph{Results}  Inter-annotator agreement is substantial (Cohen's $\kappa$=0.67 pooled; 0.62 on the four conversational corpora), establishing a reliable ground truth. Against it the pruner agrees with the annotators on KEEP decisions (ratified in 72–89\% of cases) and, on the conversational corpora, on DELETE decisions (52–84\% ratified; Figure \ref{pruning_validation}). EuroParlMin is the sole failure: annotators agreed almost perfectly there ($\kappa$=1.0) yet ratified only 4\% of the model's deletions, the pruner removing substantive parliamentary content that both annotators kept. This matches EuroParlMin's formal, low-filler style, a pruner tuned for conversational filler has little to remove and misfires and is consistent with the corpus's lower faithfulness under both allocation schemes. Because EuroParlMin loses only 5.6\% of lines (2.3\% of tokens) to pruning, this affects a small fraction of the corpus and does not bear on the matched-budget comparisons, which equalise tokens by construction.

\section{Data-Mixture Construction Details}
\label{app:mixtures}

This appendix section provides the complete per-corpus token and meeting counts for every mixture, across all budgets and (for example-level) all seeds. Table~\ref{tab:app-balanced} gives the balanced scheme, Table~\ref{tab:app-natural} the natural scheme, and Table~\ref{tab:app-example} the example-level seed variation.

% ---------------- BALANCED ----------------
\begin{table}[]
\centering
\resizebox{\columnwidth}{!}{%
\small
\setlength{\tabcolsep}{5pt}
\begin{tabular}{llrrrrr}
\toprule
\textbf{Corpus} & & \textbf{2M} & \textbf{4M} & \textbf{8M} & \textbf{16M} & \textbf{32M} \\
\midrule
\multirow{2}{*}{AMI}
 & tokens     & 399K & 800K & 1.6M & 3.2M & 6.4M \\
 & oversample & $1.0$ & $1.19$ & $2.45$ & $4.86$ & $9.73$ \\
\midrule
\multirow{2}{*}{ICSI}
 & tokens     & 397K & 794K & 1.6M & 3.2M & 6.4M \\
 & oversample & $1.0$ & $1.37$ & $2.79$ & $5.56$ & $11.09$ \\
\midrule
\multirow{2}{*}{ELITR}
 & tokens     & 399K & 800K & 1.6M & 3.2M & 6.4M \\
 & oversample & $1.0$ & $1.0$ & $1.70$ & $3.38$ & $6.82$ \\
\midrule
\multirow{2}{*}{EPM}
 & tokens     & 400K & 800K & 1.6M & 3.2M & 6.4M \\
 & oversample & $1.0$ & $1.0$ & $1.0$ & $1.0$ & $1.0$ \\
\midrule
\multirow{2}{*}{MB}
 & tokens     & 400K & 800K & 1.6M & 3.2M & 6.4M \\
 & oversample & $1.0$ & $1.0$ & $1.0$ & $1.0$ & $1.0$ \\
\bottomrule
\end{tabular}%
}
\caption{\textbf{Balanced scheme}: per-corpus token allocation and oversampling ratio
across budgets. Each corpus receives an equal ${\sim}B/5$ share, so the minority
corpora (AMI, ICSI, ELITR) are increasingly oversampled to fill it while the majority
corpora (EPM, MB) always draw distinct meetings ($1.0\times$).}
\label{tab:app-balanced}
\end{table}

% ---------------- NATURAL ----------------
\begin{table}[t]
\centering
\resizebox{\columnwidth}{!}{%
\small
\setlength{\tabcolsep}{5pt}
\begin{tabular}{llrrrrr}
\toprule
\textbf{Corpus} & & \textbf{2M} & \textbf{4M} & \textbf{8M} & \textbf{16M} & \textbf{32M} \\
\midrule
\multirow{2}{*}{AMI}
 & tokens   & 31K & 61K & 125K & 249K & 500K \\
 & meetings & 5   & 10  & 20   & 37   & 76 \\
\midrule
\multirow{2}{*}{ICSI}
 & tokens   & 26K & 51K & 105K & 215K & 436K \\
 & meetings & 3   & 4   & 8    & 15   & 33 \\
\midrule
\multirow{2}{*}{ELITR}
 & tokens   & 43K & 87K & 176K & 353K & 710K \\
 & meetings & 5   & 9   & 16   & 34   & 64 \\
\midrule
\multirow{2}{*}{EPM}
 & tokens   & 932K & 1.86M & 3.73M & 7.46M & 14.9M \\
 & meetings & 102  & 195   & 405   & 790   & 1554 \\
\midrule
\multirow{2}{*}{MB}
 & tokens   & 965K & 1.93M & 3.86M & 7.72M & 15.4M \\
 & meetings & 239  & 514   & 997   & 1968  & 3903 \\
\bottomrule
\end{tabular}%
}
\caption{\textbf{Natural scheme}: per-corpus token allocation and unique meeting count across budgets. Shares follow native corpus size (AMI~1.6\%, ICSI~1.4\%, ELITR~2.2\%, EPM~46.6\%, MB~48.3\%), so the majority corpora dominate every budget. The scheme never oversamples: even at 32M, ICSI draws only 33 of its 43 meetings.}
\label{tab:app-natural}
\end{table}

% ---------------- EXAMPLE-LEVEL ----------------
\begin{table}[]
\centering
\small
\setlength{\tabcolsep}{6pt}
\begin{tabular}{lrrrr}
\toprule
\textbf{Corpus} & \textbf{Meetings} & \textbf{s1} & \textbf{s2} & \textbf{s15} \\
\midrule
AMI   & 43 & 276K & 283K & 292K \\
ICSI  & 43 & 578K & 578K & 578K \\
ELITR & 43 & 464K & 473K & 457K \\
EPM   & 43 & 417K & 402K & 417K \\
MB    & 43 & 137K & 155K & 167K \\
\midrule
\textbf{Total} & 215 & 1.87M & 1.89M & 1.91M \\
\bottomrule
\end{tabular}
\caption{\textbf{Example-level scheme}: per-corpus token counts across three seeds (pruned). Each corpus contributes 43 meetings, but token counts differ because transcripts vary in length: ICSI ($\sim$4$\times$ MB). ICSI is identical across seeds as it has exactly 43 meetings; other corpora vary with the random draw.}
\label{tab:app-example}
\end{table}

\section{Transcript Length and Truncation}
\label{app:truncation}

Table~\ref{tab:truncation} reports transcript lengths and truncation counts per
corpus and split, measured with the Mistral-7B tokeniser. The budget available to
the transcript is not the full $16{,}384$-token window: the chat template consumes
$70$ tokens, and at training time the reference minute must also fit, so the
transcript budget is $16{,}384$ minus template, summary, and end-of-turn token.
Test-time inputs are subject only to the template overhead.

Truncation is a training-side constraint. Across train and development, $12.3\%$
of transcripts exceed the window; at test time only $3.3\%$ do, so evaluation
inputs are near-complete on every corpus. Rate and severity also come apart. ICSI
is truncated most often ($53.8\%$) but least severely, retaining $82\%$ of a
median $15.8$K-token transcript: its meetings sit just above the window, so
crossing it costs little. EuroParlMin is truncated less often ($23.9\%$) but far
more heavily, retaining $65\%$ on average, with a longest transcript of $209$K
tokens, twelve times the window. AMI is never truncated in any split.

This bears on the training target rather than on evaluation. For a truncated
example the reference minute still covers the whole session while the input does
not, so the model is trained to produce content its input does not contain. The
effect is concentrated where severity is highest, and EuroParlMin records the
lowest faithfulness of any corpus under both allocation schemes
(Table~\ref{tab:judge-full}); ICSI, truncated more often but far less severely,
records the highest. We note the association without isolating it as the cause.
Because test transcripts are almost never truncated, the annotation guideline for
faithfulness that addresses truncated tails (Appendix~\ref{app:guidelines},
Task~B) applies to very few items in practice.

\begin{table}[t]
\resizebox{\columnwidth}{!}{%
\centering
\small
\setlength{\tabcolsep}{4pt}
\begin{tabular}{llrrrrr}
\toprule
\textbf{Split} & \textbf{Corpus} & \textbf{$n$} & \textbf{\#Trunc.} & \textbf{\%} & \textbf{Median (tok)} & \textbf{Kept (\%)} \\
\midrule
\multirow{6}{*}{Train}
 & AMI   &   98 &   0 &  0.0 &  7{,}558 & --- \\
 & ICSI  &   43 &  22 & 51.2 & 15{,}667 & 81.9 \\
 & ELITR &   84 &  28 & 33.3 & 12{,}105 & 71.3 \\
 & EPM   & 2065 & 529 & 25.6 &  9{,}077 & 64.7 \\
 & MB    & 5169 & 397 &  7.7 &  1{,}890 & 62.3 \\
\cmidrule(lr){2-7}
 & \textbf{All} & \textbf{7459} & \textbf{976} & \textbf{13.1} & \textbf{3{,}261} & --- \\
\midrule
\multirow{6}{*}{Dev}
 & AMI   &   20 &  0 &  0.0 &  7{,}212 & --- \\
 & ICSI  &    9 &  6 & 66.7 & 16{,}596 & 83.4 \\
 & ELITR &   10 &  3 & 30.0 & 10{,}746 & 76.9 \\
 & EPM   &  187 &  9 &  4.8 &     517 & 60.8 \\
 & MB    &  861 & 58 &  6.7 &  1{,}589 & 63.4 \\
\cmidrule(lr){2-7}
 & \textbf{All} & \textbf{1087} & \textbf{76} & \textbf{7.0} & \textbf{1{,}402} & --- \\
\midrule
\multirow{6}{*}{Test}
 & AMI   &   20 &  0 &  0.0 &  8{,}632 & --- \\
 & ICSI  &    6 &  2 & 33.3 & 13{,}352 & 72.7 \\
 & ELITR &   38 &  5 & 13.2 & 11{,}200 & 91.3 \\
 & EPM   &  242 &  1 &  0.4 &     394 & 67.5 \\
 & MB    &  862 & 31 &  3.6 &  1{,}294 & 62.6 \\
\cmidrule(lr){2-7}
 & \textbf{All} & \textbf{1168} & \textbf{39} & \textbf{3.3} & \textbf{1{,}119} & --- \\
\bottomrule
\end{tabular}%
}
\caption{Transcript lengths and truncation per corpus and split (Mistral-7B
tokeniser, unpruned transcripts). \emph{\#Trunc.} counts transcripts exceeding the
available budget; \emph{Kept} is the mean percentage of a transcript retained,
computed over truncated transcripts only. Train and development use the training
budget (window minus template, summary, and end-of-turn token); test uses the
inference budget (window minus template). Truncation is far heavier in training
($12.3\%$ of train${+}$dev) than at test ($3.3\%$), and rate and severity diverge:
ICSI is truncated most often but retains the most, EuroParlMin least often of the
three affected corpora but retains the least.}
\label{tab:truncation}
\end{table}

\section{Multi-Seed Robustness}
\label{app:msrobust}
\subsection{Balanced and Natural Schemes}
\label{app:bnseeds}

The main grid (Section~\ref{sec:expsettings}) runs on seed~$42$. To check that the
per-domain crossover of RQ1 is a property of the allocation rather than of one
draw, we repeat both schemes under the pruned condition at the 2M, 8M, and 32M
budgets with two further seeds, $\{2, 15\}$, giving 12 additional runs. The seed
governs adapter initialisation, dropout, and training order, as well as which
meetings are sampled to fill each corpus's share of the budget; every other
setting is unchanged (Table~\ref{tab:hparams}). Table~\ref{tab:multiseed-all}
reports the resulting scores as mean\,$\pm$\,std over the three seeds, across the
full metric family.

The crossover reproduces. At 8M and 32M, on all five metrics without exception,
balanced leads each of the three data-scarce domains (AMI, ICSI, ELITR) and
natural leads the two data-rich ones (EPM, MB), so the macro average favours
balanced while the micro average favours natural, the same sign flip reported in
Section~\ref{sec:rq1}. The gaps are large relative to seed variation on the
minority domains: at 32M on ROUGE-Lsum, balanced leads AMI by $0.168$, ICSI by
$0.227$, and ELITR by $0.189$, against seed standard deviations below $0.07$. At
2M the picture is weaker, as the earlier budget rows also show: the minority
advantage is smaller, ELITR falls within seed variation, and the macro averages
of the two schemes are indistinguishable. On the majority side, only
MeetingBank is a consistent cost to balancing; the EuroParlMin gap narrows with
budget and at 32M is smaller than its own seed spread, matching the erosion of
natural's early EPM lead noted in Section~\ref{sec:rq2}.

\begin{figure}[t]
  \centering
  \includegraphics[width=0.99\linewidth]{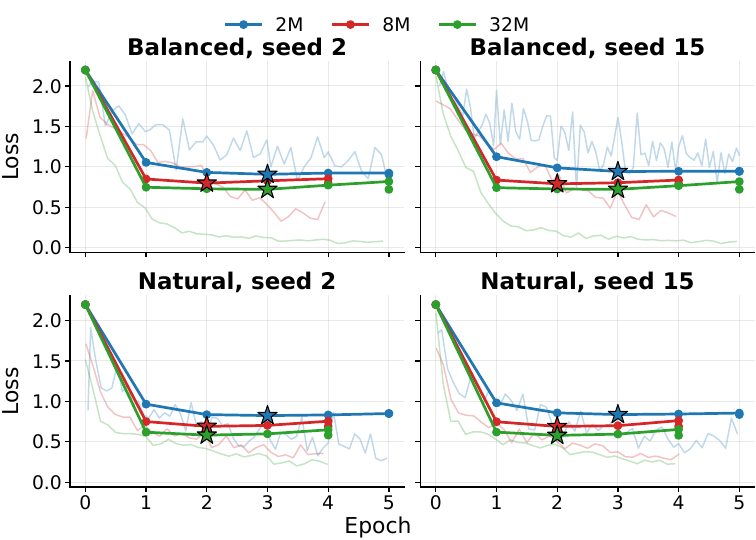}
  \caption{Multi-seed training trajectories (pruned, seeds 2 and 15): per-epoch
training loss (faint) and development loss (bold, with markers) for the balanced
(top) and natural (bottom) schemes at the 2M, 8M, and 32M budgets. The star marks
the selected checkpoint, at the lowest development loss. As in the seed-42 grid
(Figure~\ref{fig:checkpoints}), development loss bottoms out and then rises while
training loss keeps falling, and each run stops two epochs after its minimum under
patience-2 early stopping.}
  \label{fig:seed-checkpoints}
\end{figure}

\subsection{Example-Level Baseline}
\label{app:seeds}

The three example-level seeds \{1,2,15\} were selected to maximise the diversity of
the sampled data on the corpora where sampling variance is real. For each candidate
seed we simulated the builder's per-corpus selection of $43$ meetings and measured
the mean pairwise overlap on the large corpora (EPM, MB), where the pool greatly
exceeds $43$; the chosen trio minimises that overlap.
Table~\ref{tab:app-seeds} summarises the outcome: the large corpora are fully
distinct across seeds, while the small corpora overlap only because the pool is too
small to draw three disjoint $43$-subsets.

\begin{table}[h]
\centering
\small
\begin{tabular}{lrl}
\toprule
\textbf{Corpus} & \textbf{Pairwise overlap} & \textbf{Coverage across seeds} \\
\midrule
EPM   & 0\% (fully distinct) & --- \\
MB    & 0\% (fully distinct) & --- \\
AMI   & (arithmetic)         & 83/98 \\
ELITR & (arithmetic)         & 74/84 \\
ICSI  & 100\% (forced)       & 43/43 \\
\bottomrule
\end{tabular}
\caption{Example-level seed selection \{1,2,15\}. Large corpora are fully distinct
across seeds (genuine sampling variance); small-corpus overlap is arithmetically
unavoidable (pool $\leq 3\times43$) but offset by near-complete coverage of the
minority data. For ICSI ($|{\rm pool}|=43$) all meetings are used, so only training
order varies across seeds.}
\label{tab:app-seeds}
\end{table}

\section{Model Selection}
\label{app:model-selection}
Table~\ref{tab:zeroshot-full} reports zero-shot screening on ROUGE-Lsum and BERTScore-F1,
per corpus, with macro (equal weight per corpus) and micro (weighted by number of meetings)
averages; per-corpus and per-average bests are in \textbf{bold}. $n$ is the number of test
meetings, with ELITR collapsed to one prediction per meeting (scored as the maximum over its
references). We screen five open models (Mistral-7B-Instruct-v0.3\footnote{\url{https://huggingface.co/mistralai/Mistral-7B-Instruct-v0.3}}, Llama-3.1-8B-Instruct\footnote{\url{https://huggingface.co/meta-llama/Llama-3.1-8B-Instruct}}, Llama-3.2-3B-Instruct\footnote{\url{https://huggingface.co/meta-llama/Llama-3.2-3B-Instruct}}, Qwen3-8B\footnote{\url{https://huggingface.co/Qwen/Qwen3-8B}}, gemma-4-12B-it\footnote{\url{https://huggingface.co/google/gemma-4-12B-it}}) and one closed model (GPT-4o), the latter added
because the third AutoMin challenge reports closed frontier models
leading its shared task; we evaluate it under exactly the same prompt, decoding settings, and
metrics as the open candidates.

\paragraph{Choice of metric.} The two metrics disagree at the top: Gemma-4-12B leads
ROUGE-Lsum macro (0.347), while Mistral-7B leads BERTScore-F1 on both averages (macro 0.833,
micro 0.835); both rank Qwen3-8B weakest. We base selection on BERTScore-F1 rather than
ROUGE-Lsum, as embedding overlap tracks semantic adequacy more closely than surface $n$-gram
overlap.

\paragraph{Robust to the choice of average.} Mistral-7B's BERTScore-F1 lead holds under
both averaging schemes, which matters because the test sets are themselves imbalanced (ICSI
6 meetings, MB 862): the micro average is dominated by MB, while macro gives every corpus
equal weight. Leading on both means the choice is not an artifact of MB's size---the very
majority-domination this paper analyses.

\paragraph{Robust across domains.} Mistral-7B also ranks first on three of five corpora
(ELITR, EPM, MB), separates from the field on the two largest (EPM, MB) with non-overlapping
confidence intervals, and has the highest per-corpus floor of all six candidates: its
weakest-domain BERTScore-F1 is 0.824, above every other model's worst domain. In a
multi-domain setting this floor matters---it means Mistral does not sacrifice any single
domain. Llama-3.1-8B edges it on the two smallest corpora (AMI, ICSI), but within confidence
intervals (unsurprising at $n{=}20$ and $n{=}6$) and without matching Mistral's floor or its
separation on the large corpora.

\paragraph{GPT-4o does not lead under our protocol.} GPT-4o ranks last of the six on macro
BERTScore-F1 (0.815, against 0.833 for Mistral-7B and 0.826 for the 3B Llama) and on every
average we report. This does not so much contradict AutoMin's finding as expose a mismatch
between output format and reference form. The precision/recall split makes the mechanism
visible: on MB, GPT-4o attains the highest BERTScore recall of any candidate alongside the lowest precision. It locates the reference
content and then pads heavily around it, emitting section headers, attendee blocks, horizontal
rules, and closing formalities that our references-short, flat, procedural minutes do not
contain. Every unmatched token depresses precision, and F1 with it. We report all systems raw,
applying no output cleaning to GPT-4o that we do not apply to the open models, so the
comparison measures the models as prompted rather than as post-processed. We did not prompt GPT-4o to suppress markdown, since the screening applies one prompt to all candidates; a format-controlled prompt or post-processing would likely raise its precision, but GPT-4o was not a fine-tuning candidate and the choice does not affect the main results. GPT-4o is therefore
not a ceiling in our setting, and we do not treat it as one.

\paragraph{Reproducibility of the closed model.} We use \texttt{gpt-4o-2024-11-20} with
temperature 0, top-$p$ 1.0, and a fixed seed. OpenAI's \texttt{seed} parameter is best-effort:
the run returned eight distinct \texttt{system\_fingerprint} values, so outputs are not
bit-reproducible and we do not describe the setting as greedy decoding.

\paragraph{Llama-3.2-3B as a capacity control.} We carry Llama-3.2-3B forward as a
lower-capacity control for RQ5. Model size is not monotonic with quality in this screen
(7B Mistral $>$ 12B Gemma and $>$ GPT-4o on BERTScore-F1; 3B Llama is competitive with the 8B
models), so it is a genuine capacity probe rather than a uniformly weaker baseline.

\begin{table}[t]
\centering\small\setlength{\tabcolsep}{4pt}
\resizebox{\columnwidth}{!}{%
\begin{tabular}{lccccc|cc}
\toprule
\textbf{Model} & \textbf{AMI} & \textbf{ICSI} & \textbf{ELITR} & \textbf{EPM} & \textbf{MB} & \textbf{Macro} & \textbf{Micro} \\
 & \textit{20} & \textit{6} & \textit{38} & \textit{242} & \textit{862} & & \\
\midrule
\multicolumn{8}{l}{\textit{ROUGE-1}} \\
Llama-3.2-3B & 0.462 & 0.360 & 0.369 & 0.318 & 0.211 & 0.344 & 0.243 \\
Llama-3.1-8B & \textbf{0.468} & 0.370 & 0.370 & 0.329 & 0.217 & 0.351 & 0.250 \\
Qwen3-8B & 0.402 & 0.398 & 0.344 & 0.294 & 0.179 & 0.324 & 0.213 \\
Mistral-7B & 0.454 & 0.343 & 0.401 & \textbf{0.343} & \textbf{0.245} & 0.357 & \textbf{0.274} \\
Gemma-4-12B & 0.443 & \textbf{0.430} & \textbf{0.425} & 0.315 & 0.215 & \textbf{0.366} & 0.248 \\
GPT-4o & 0.395 & 0.403 & 0.392 & 0.307 & 0.194 & 0.338 & 0.228 \\
\midrule
\multicolumn{8}{l}{\textit{ROUGE-2}} \\
Llama-3.2-3B & 0.120 & 0.075 & 0.083 & 0.140 & 0.109 & 0.106 & 0.115 \\
Llama-3.1-8B & \textbf{0.132} & 0.079 & 0.088 & \textbf{0.147} & 0.115 & 0.112 & 0.121 \\
Qwen3-8B & 0.102 & 0.090 & 0.082 & 0.129 & 0.098 & 0.100 & 0.104 \\
Mistral-7B & 0.121 & 0.065 & \textbf{0.101} & 0.144 & \textbf{0.131} & \textbf{0.112} & \textbf{0.132} \\
Gemma-4-12B & 0.110 & \textbf{0.098} & 0.095 & 0.131 & 0.113 & 0.109 & 0.116 \\
GPT-4o & 0.095 & 0.087 & 0.087 & 0.129 & 0.101 & 0.100 & 0.106 \\
\midrule
\multicolumn{8}{l}{\textit{ROUGE-L}} \\
Llama-3.2-3B & 0.219 & \textbf{0.170} & 0.178 & 0.207 & 0.159 & 0.187 & 0.171 \\
Llama-3.1-8B & \textbf{0.220} & 0.166 & 0.170 & 0.213 & 0.164 & 0.187 & 0.175 \\
Qwen3-8B & 0.180 & 0.152 & 0.150 & 0.187 & 0.136 & 0.161 & 0.148 \\
Mistral-7B & 0.207 & 0.145 & 0.181 & \textbf{0.222} & \textbf{0.183} & \textbf{0.188} & \textbf{0.191} \\
Gemma-4-12B & 0.191 & 0.153 & \textbf{0.182} & 0.201 & 0.158 & 0.177 & 0.168 \\
GPT-4o & 0.172 & 0.146 & 0.169 & 0.195 & 0.142 & 0.165 & 0.155 \\
\midrule
\multicolumn{8}{l}{\textit{ROUGE-Lsum}} \\
Llama-3.2-3B & 0.439 & 0.345 & 0.359 & 0.307 & 0.188 & 0.328 & 0.223 \\
Llama-3.1-8B & \textbf{0.444} & 0.353 & 0.358 & 0.319 & 0.196 & 0.334 & 0.232 \\
Qwen3-8B & 0.385 & 0.383 & 0.334 & 0.285 & 0.164 & 0.310 & 0.199 \\
Mistral-7B & 0.435 & 0.325 & 0.389 & \textbf{0.330} & \textbf{0.220} & 0.340 & \textbf{0.252} \\
Gemma-4-12B & 0.420 & \textbf{0.410} & \textbf{0.409} & 0.303 & 0.194 & \textbf{0.347} & 0.229 \\
GPT-4o & 0.377 & 0.389 & 0.382 & 0.298 & 0.176 & 0.324 & 0.212 \\
\midrule
\multicolumn{8}{l}{\textit{BERTScore-F1}} \\
Llama-3.2-3B & 0.841 & 0.821 & 0.831 & 0.816 & 0.819 & 0.826 & 0.819 \\
Llama-3.1-8B & \textbf{0.844} & \textbf{0.828} & 0.830 & 0.822 & 0.831 & 0.831 & 0.830 \\
Qwen3-8B & 0.816 & 0.818 & 0.824 & 0.815 & 0.821 & 0.819 & 0.820 \\
Mistral-7B & 0.838 & 0.824 & \textbf{0.840} & \textbf{0.825} & \textbf{0.837} & \textbf{0.833} & \textbf{0.835} \\
Gemma-4-12B & 0.818 & 0.816 & 0.828 & 0.818 & 0.824 & 0.821 & 0.823 \\
GPT-4o & 0.808 & 0.809 & 0.832 & 0.814 & 0.813 & 0.815 & 0.814 \\
\bottomrule
\end{tabular}%
}
\caption{Zero-shot summarisation quality across five meeting corpora. Macro is the unweighted mean over corpora; Micro is weighted by test-set size (counts in italics). ELITR scores take the maximum over its multiple references. Best per column in \textbf{bold}.}
\label{tab:zeroshot-full}
\end{table}

\section{Per-Domain Fine-Tuned Baselines}
\label{app:perdomain}

This appendix reports the per-domain baselines for RQ6. One QLoRA adapter is trained per corpus on the full pruned training set of that corpus, with no token cap, mixing, or oversampling, so that every training meeting is seen exactly once per epoch. The development set is likewise restricted to the same corpus, so early stopping and checkpoint selection are driven by that domain alone. All other settings follow Table~\ref{tab:hparams} with seed 42. At test time each adapter generates minutes only for the test meetings of its own domain, under the decoding settings of Section \ref{sec:inference}. The five outputs are then concatenated into one 1{,}168-meeting system and scored with the procedure of Section \ref{sec:evaluation}, so that per-domain, macro, and micro results are directly comparable with Table~\ref{tab:rq1-32m}.

\begin{figure*}[t]
\centering
\includegraphics[width=\textwidth]{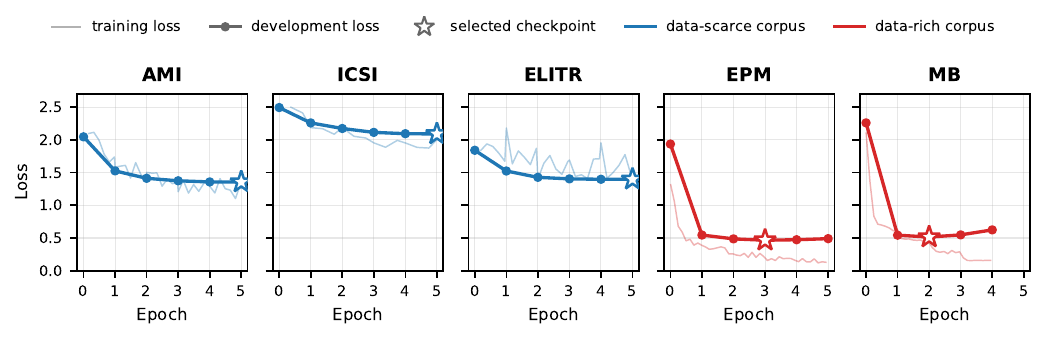}
\caption{Per-domain baselines: per-epoch training loss (faint) and development loss (bold, with markers) for each adapter, measured on the development set of the same corpus (pruned, seed 42). The star marks the selected checkpoint. The data-rich adapters (EPM, MB) reach a development-loss minimum and are terminated by patience-2 early stopping, as in Figure~\ref{fig:checkpoints}; the data-scarce adapters (AMI, ICSI, ELITR) are still improving at the five-epoch ceiling.}
\label{fig:perdomain-losses}
\end{figure*}

\subsection{Setup and Training Trajectories}
\label{app:perdomain-train}

The three minority corpora yield 0.58-0.94M tokens and 3-7 optimiser steps per epoch at the effective batch size of 16, against 130 and 324 steps for EPM and MB. Figure~\ref{fig:perdomain-losses} shows the resulting trajectories. Two regimes are visible. The EPM and MB adapters behave like the mixture runs of Figure~\ref{fig:checkpoints}: development loss reaches its minimum at epoch 3 and 2 respectively, then rises, and patience-2 early stopping terminates training. The AMI, ICSI, and ELITR adapters never reach this regime. Their development loss decreases monotonically through epoch 5 and flattens only as the cosine schedule drives the learning rate to zero, so the final checkpoint is selected in all three cases. Under the fixed protocol these corpora receive 15-35 updates in total, an order of magnitude fewer than balanced allocation at 32M provides them through oversampling.

The minority adapters also generate unreliably. Where an adapter terminates, its minutes are of reference length, but on 7 of 20 AMI and 2 of 6 ICSI meetings it repeats a single sentence until the 2{,}048-token cap, and 2 of 38 ELITR outputs are empty; balanced allocation reaches the cap on none of these two domains and produces no empty output. The EPM and MB adapters show no such behaviour: their median lengths (247 and 73 tokens) and cap rates (17/242 and 47/862) match those of balanced allocation (249 and 74; 29/242 and 50/862). The ICSI outputs reproduce the reference section structure with near-identical text under every heading. These adapters have learned the form of the minutes but not their content, which accounts for scores below the untuned model. Decoding was not modified, so all systems are scored as generated.

\subsection{Full Results}
\label{app:perdomain-results}

Table~\ref{tab:app-perdomain} reports the per-domain baselines across the complete metric family with bootstrap 95\% confidence intervals, in the layout of Table~\ref{tab:app-full-pruned}. Against balanced allocation at 32M, the baseline is below the interval on ICSI and ELITR on every metric, and its AMI interval excludes the balanced point estimate on all ROUGE variants. On MB the baseline interval excludes the balanced value in the other direction, and on EPM the two overlap. The macro interval (ROUGE-Lsum 0.37-0.42) lies wholly below the balanced macro (0.500), the micro interval (0.63-0.66) wholly above it (0.615).

\begin{table*}[t]
\centering\scriptsize\setlength{\tabcolsep}{3pt}
\resizebox{\textwidth}{!}{%
\begin{tabular}{lccccccc}
\toprule
\textbf{Metric} & \textbf{AMI} & \textbf{ICSI} & \textbf{ELITR} & \textbf{EPM} & \textbf{MB} & \textbf{Macro} & \textbf{Micro} \\
 & \textit{20} & \textit{6} & \textit{38} & \textit{242} & \textit{862} & & \\
\midrule
ROUGE-1    & 0.394$_{[0.30,0.48]}$ & 0.224$_{[0.14,0.30]}$ & 0.130$_{[0.09,0.17]}$ & 0.570$_{[0.54,0.60]}$ & 0.720$_{[0.70,0.74]}$ & 0.408$_{[0.38,0.43]}$ & 0.662$_{[0.65,0.68]}$ \\
ROUGE-2    & 0.168$_{[0.12,0.22]}$ & 0.049$_{[0.03,0.07]}$ & 0.041$_{[0.03,0.06]}$ & 0.432$_{[0.41,0.46]}$ & 0.632$_{[0.61,0.65]}$ & 0.264$_{[0.25,0.28]}$ & 0.561$_{[0.54,0.58]}$ \\
ROUGE-L    & 0.234$_{[0.18,0.29]}$ & 0.138$_{[0.10,0.18]}$ & 0.100$_{[0.07,0.13]}$ & 0.511$_{[0.48,0.54]}$ & 0.696$_{[0.68,0.71]}$ & 0.336$_{[0.32,0.35]}$ & 0.628$_{[0.61,0.64]}$ \\
ROUGE-Lsum & 0.382$_{[0.29,0.47]}$ & 0.210$_{[0.13,0.29]}$ & 0.126$_{[0.09,0.17]}$ & 0.564$_{[0.53,0.59]}$ & 0.703$_{[0.69,0.72]}$ & 0.397$_{[0.37,0.42]}$ & 0.647$_{[0.63,0.66]}$ \\
BERTScore  & 0.868$_{[0.86,0.88]}$ & 0.817$_{[0.79,0.84]}$ & 0.772$_{[0.70,0.82]}$ & 0.897$_{[0.89,0.90]}$ & 0.939$_{[0.94,0.94]}$ & 0.858$_{[0.84,0.87]}$ & 0.923$_{[0.92,0.93]}$ \\
\bottomrule
\end{tabular}}
\caption{Per-domain fine-tuned baselines (pruned, seed 42) across the complete metric family, with bootstrap 95\% CIs in subscripts (10{,}000 resamples over test meetings; the macro CI resamples within each domain and averages the domain means). Each adapter is trained on its full training set: 0.66M (AMI), 0.58M (ICSI), 0.94M (ELITR), 19.7M (EPM) and 20.4M (MB) tokens per epoch. Column italics are test-set meeting counts. The balanced and natural blocks for comparison are in Table~\ref{tab:app-full-pruned}.}
\label{tab:app-perdomain}
\end{table*}

\section{Summary Lengths (Reference and Generated)}
\label{app:summary-lengths}
Table~\ref{tab:summary-lengths} reports the token-length distribution of the test-set
reference summaries, measured with the Mistral-7B tokenizer (for ELITR, all annotator
references are counted). Reference summaries are short to moderate in length across
all domains, with medians between 87 and 664 tokens. The 2048-token generation budget
covers every test reference in AMI, ICSI, ELITR, and MeetingBank in full, and
$98.8\%$ of test references overall. The remaining references belong to EuroParlMin
and arise from a corpus formatting artefact: a small number of parliamentary minutes
are provided as undivided full-chapter documents rather than per-session summaries. Such documents are not representative meeting minutes, and
the most extreme of them occur only in the training and development splits. We
therefore set the generation budget generously above the typical reference length
rather than to these atypical outliers, which keeps decoding efficient without
affecting the vast majority of references.

\begin{table}[h]
\centering
\small
\begin{tabular}{lrrrrr}
\toprule
\textbf{Corpus} & \textbf{N} & \textbf{Mean} & \textbf{P95} & \textbf{P99} & \textbf{Max} \\
\midrule
AMI              &  20 & 376 &  517 &  582 &  599 \\
ICSI             &   6 & 628 &  918 &  938 &  944 \\
MB               & 862 & 101 &  230 &  313 &  959 \\
ELITR (test)     &  55 & 539 & 1122 & 1240 & 1272 \\
ELITR (test2)    &  10 & 907 & 1618 & 1808 & 1856 \\
ELITR (test2023) &  12 & 769 & 1330 & 1381 & 1394 \\
EPM              & 242 & 472 & 2400 & 4823 & 7253 \\
\bottomrule
\end{tabular}
\caption{Test-set reference-summary lengths in tokens (Mistral-7B tokenizer). $N$ is
the number of test references; ELITR is split by its three test sets, with all
annotator references counted. The 2048-token generation budget covers 98.8\% of
references in full; the longer ones are undivided full-chapter EuroParlMin documents
rather than per-session minutes.}
\label{tab:summary-lengths}
\end{table}

Table~\ref{tab:gen-tokens-ci} reports the length of summaries generated by the six screened candidates on the \textbf{test set} of each corpus, complementing the reference-length distribution above (Table~\ref{tab:summary-lengths}). Lengths are each model's own generated-token count, so they are measured with that model's tokeniser and are indicative rather than strictly commensurable across rows. ELITR is collapsed to one prediction per meeting ($n{=}38$).

The means sit well above the per-domain medians on EPM and MB, reflecting the same right-skewed, heavy-tailed length distributions seen in the references (Table~\ref{tab:summary-lengths}); the large standard deviations on these two corpora are therefore a property of the heterogeneous inputs rather than unstable generation, and a small number of outputs approach the $2048$-token cap. This is consistent with the cap being non-binding for over $98\%$ of generations (Section~\ref{sec:expsettings}).

The table also quantifies the length mismatch behind GPT-4o's ranking (Appendix~\ref{app:model-selection}). On MeetingBank, whose references average $101$ tokens, GPT-4o emits $575$ on average, the largest ratio of generation to reference of any candidate. The excess is structural rather than substantive: section headers, attendee blocks, and closing formalities that the reference minutes do not contain. Every such token is unmatched, which depresses precision and the F1 that follows from it, while leaving recall intact.
(Reference lengths in Table~\ref{tab:summary-lengths} count all annotator references; here we report one generation per meeting.)

\begin{table}[h]
\centering

\setlength{\tabcolsep}{5pt}
\footnotesize
\resizebox{\columnwidth}{!}{%
\begin{tabular}{llrcc}
\toprule
\textbf{Model} & \textbf{Dataset} & $n$ & \textbf{Mean $\pm$ SD} & \textbf{95\% CI} \\
\midrule
Llama-3.2-3B & AMI & 20 & 457.7 $\pm$ 65.1 & [427.2, 488.2] \\
 & ICSI & 6 & 420.5 $\pm$ 43.8 & [374.5, 466.5] \\
 & ELITR & 38 & 567.1 $\pm$ 261.0 & [481.3, 652.9] \\
 & EPM & 242 & 314.3 $\pm$ 157.4 & [294.4, 334.3] \\
 & MB & 862 & 413.9 $\pm$ 284.4 & [394.9, 433.0] \\
\midrule
Llama-3.1-8B & AMI & 20 & 443.2 $\pm$ 74.5 & [408.4, 478.1] \\
 & ICSI & 6 & 551.5 $\pm$ 140.1 & [404.4, 698.6] \\
 & ELITR & 38 & 733.7 $\pm$ 410.9 & [598.6, 868.7] \\
 & EPM & 242 & 304.2 $\pm$ 177.6 & [281.7, 326.7] \\
 & MB & 862 & 367.6 $\pm$ 159.1 & [357.0, 378.3] \\
\midrule
Mistral-7B & AMI & 20 & 520.6 $\pm$ 162.6 & [444.5, 596.7] \\
 & ICSI & 6 & 409.0 $\pm$ 126.0 & [276.8, 541.2] \\
 & ELITR & 38 & 739.3 $\pm$ 386.3 & [612.4, 866.3] \\
 & EPM & 242 & 313.9 $\pm$ 240.1 & [283.5, 344.3] \\
 & MB & 862 & 374.9 $\pm$ 194.2 & [361.9, 387.9] \\
\midrule
Qwen3-8B & AMI & 20 & 964.0 $\pm$ 183.9 & [877.9, 1050.1] \\
 & ICSI & 6 & 941.7 $\pm$ 201.6 & [730.1, 1153.2] \\
 & ELITR & 38 & 1317.6 $\pm$ 411.5 & [1182.4, 1452.9] \\
 & EPM & 242 & 493.3 $\pm$ 345.8 & [449.5, 537.1] \\
 & MB & 862 & 680.9 $\pm$ 379.3 & [655.6, 706.3] \\
\midrule
Gemma-4-12B & AMI & 20 & 781.1 $\pm$ 148.2 & [711.8, 850.4] \\
 & ICSI & 6 & 805.7 $\pm$ 79.2 & [722.6, 888.8] \\
 & ELITR & 38 & 798.4 $\pm$ 108.9 & [762.7, 834.2] \\
 & EPM & 242 & 361.7 $\pm$ 296.6 & [324.1, 399.2] \\
 & MB & 862 & 507.6 $\pm$ 253.7 & [490.6, 524.5] \\
 \midrule
 GPT-4o     & AMI   & 20  & 929.2 $\pm$ 152.4  & [857.9, 1000.5] \\
           & ICSI  & 6   & 965.0 $\pm$ 210.7  & [743.9, 1186.1] \\
           & ELITR & 38  & 1002.5 $\pm$ 150.6 & [953.0, 1052.0] \\
           & EPM   & 242 & 431.1 $\pm$ 268.3  & [397.1, 465.1] \\
           & MB    & 862 & 575.4 $\pm$ 224.2  & [560.4, 590.4] \\
\bottomrule
\end{tabular}%
}
\caption{Generated meeting-minute length in tokens (mean $\pm$ standard deviation, with 95\% confidence interval for the mean) per model and corpus. ELITR is collapsed to one prediction per meeting ($n{=}38$).}
\label{tab:gen-tokens-ci}
\end{table}

\section{Corpus Construction}
\label{app:corpus}

This appendix gives the per-corpus construction, splitting, and unit details
summarised in Section~\ref{sec:corpora}, and the verification of the ELITR
reference-selection criterion of Section~\ref{sec:inference}.

\paragraph{Schema and splits.}
All five corpora are cast into a uniform \texttt{(id, source, summary, split)}
schema, where \texttt{source} is a speaker-labelled transcript and
\texttt{summary} the reference minute. We adopt canonical splits where they
exist-the AMI scenario-only partition, the ICSI six-meeting test set, and the
official MB and EPM splits-and a seeded, released train/development division for
ICSI (seed~42), whose canonical train/development membership is not standardised.
Split sizes are given in Table~\ref{tab:corpus-sizes}.

\begin{table}[t]
\centering
\small
\begin{tabular}{llrrr}
\toprule
\textbf{Corpus} & \textbf{Domain} & \textbf{Train} & \textbf{Dev} & \textbf{Test} \\
\midrule
AMI   & product meetings      & 98   & 20  & 20  \\
ICSI  & academic meetings     & 43   & 9   & 6   \\
MB    & council meetings      & 5169 & 861 & 862 \\
ELITR & project meetings      & 84   & 10  & 38$^{*}$ \\
EPM   & parliamentary sessions & 2065 & 187 & 242 \\
\midrule
Total &                       & 7459 & 1087 & 1168 \\
\bottomrule
\end{tabular}
\caption{Final train/development/test partitions. Splits are canonical where available,
seeded-and-released for ICSI (seed~42). ELITR figures are for the English portion; its
Test column sums three test sets ($18{+}8{+}12$). $^{*}$ELITR test counts unique
meetings (38), each carrying multiple reference minutes over which evaluation takes the
maximum.}
\label{tab:corpus-sizes}
\end{table}

\paragraph{Transcript and reference form.}
AMI and ICSI transcripts are reconstructed as speaker-labelled turns from their
dialogue-act annotations with standard disfluency removal, and their structured,
section-headered summaries are retained. MB, ELITR, and EPM are used as
distributed, preserving their native speaker labelling and reference form. The
unit of summarisation differs by corpus, reflecting each corpus's native
structure: sub-meetings for AMI, whole meetings for ICSI and ELITR, and
agenda-item segments for MB and EPM.

\paragraph{ELITR test sets and multiple references.}
ELITR supplies three distinct test sets
(\textsc{test}/\textsc{test2}/\textsc{test2023}), which we keep separate following
the corpus authors' intent; their sizes are summed in
Table~\ref{tab:corpus-sizes} ($18{+}8{+}12 = 38$ meetings). ELITR also provides
multiple independent reference minutes per meeting. For training-mixture
construction we consolidate each meeting to a single reference; for zero-shot evaluation we instead retain all
references and score each prediction as the maximum over its references, so ELITR
contributes $38$ meetings to the evaluation.

\paragraph{Verification of ELITR reference selection.}
The single-reference consolidation selects, per meeting, the reference with the
highest mean ROUGE-Lsum recall against the other references. Because this rewards
content coverage it correlates with length, so we verified its behaviour rather
than assuming it. Across the $70$ multi-reference meetings (all splits), the selected reference
was the longest candidate in $68$ cases; in the two exceptions the candidates were
within three words of one another and selection was determined by content overlap.
The criterion therefore favours the most comprehensive reference while remaining
sensitive to content rather than length alone, and avoids the arbitrariness of a
fixed-annotator or original-minutes heuristic.

\section{Fine-tuning Hyperparameters}
\label{app:hparams}
For a frozen 4-bit weight matrix $W_0 \in \mathbb{R}^{d\times k}$, QLoRA learns a low-rank
update $\Delta W = BA$, with $B \in \mathbb{R}^{d\times r}$, $A \in \mathbb{R}^{r\times k}$,
and $r \ll \min(d,k)$, so the forward pass is $h = W_0 x + \tfrac{\alpha}{r} BA x$ with
scaling factor $\alpha$. The loss is cross-entropy over summary positions
$S = \{p{+}1,\dots,n\}$ only, with prompt tokens masked:
\begin{equation}
\mathcal{L} = -\frac{1}{|S|} \sum_{i \in S} \log P(y_i \mid y_{<i}, x; \theta),
\end{equation}
where $x$ is the transcript and $\theta$ the trainable adapter parameters. This trains
$\approx$83.9M parameters, $1.14\%$ of the model. Table~\ref{tab:hparams} lists the full
configuration, applied identically across all schemes, budgets, and conditions.

\begin{table}[h]
\resizebox{\columnwidth}{!}{%
\centering
\small
\begin{tabular}{ll}
\toprule
\textbf{Hyperparameter} & \textbf{Value} \\
\midrule
Quantisation          & 4-bit NF4, double-quant, bf16 compute \\
LoRA rank $r$         & 32 \\
LoRA $\alpha$         & 16 \\
LoRA dropout          & 0.05 \\
Adapted modules       & q, k, v, o, gate, up, down \\
Optimiser             & paged AdamW (8-bit) \\
Learning rate         & $1\times10^{-4}$ \\
LR schedule           & cosine, $3\%$ warmup \\
Weight decay          & 0.01 \\
Max gradient norm     & 0.3 \\
Effective batch size  & 16 (1 $\times$ 16 accum.) \\
Max sequence length   & 16{,}384 \\
Max epochs            & 5 (early stopping, patience 2) \\
Checkpoint selection  & lowest dev loss \\
Seed                  & 42 (main grid); 1, 2, 15$^{\dagger}$ (robustness) \\
\bottomrule
\end{tabular}%
}
\caption{QLoRA fine-tuning hyperparameters, applied identically across all schemes, budgets, and conditions. $^{\dagger}$Seeds 2 and 15 repeat both allocation schemes at 2M/8M/32M (Appendix~\ref{app:bnseeds}); seeds 1, 2, and 15 are used for the example-level baseline (Appendix~\ref{app:seeds}).}
\label{tab:hparams}
\end{table}

\section{Training Trajectories and Checkpoint Selection}
\label{app:trajectories}
\begin{figure*}[t]
  \centering
  \includegraphics[width=\textwidth]{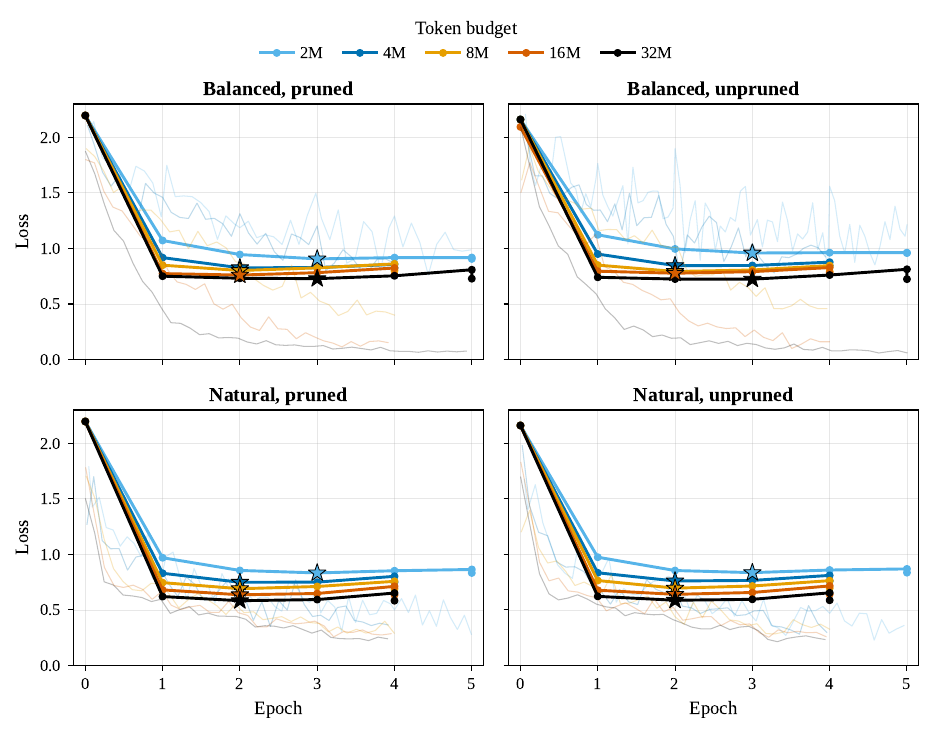}
  \caption{Per-epoch training loss (faint) and development loss (bold, with markers) for the
  balanced (top) and natural (bottom) schemes, pruned (left) and unpruned (right), on
  seed~42. The star
  marks the selected checkpoint, at the lowest development loss. Development loss bottoms out
  and then rises while training loss keeps falling; each run stops two epochs after its
  minimum under patience-2 early stopping.}
  \label{fig:checkpoints}
\end{figure*}

Figure~\ref{fig:checkpoints} shows per-epoch training and development loss for all twenty runs, and Table~\ref{tab:app-trajectories} reports the resulting checkpoint selection. Together they justify the early-stopping and best-checkpoint choices of Section~\ref{sec:training}. Development loss falls to a minimum and then rises as the model overfits, so the selected checkpoint (starred in the figure) is not the last. The minimum occurs at epoch~2 for the mid-range budgets and at epoch~3 at 2M; under balanced allocation the largest budget (32M) also selects epoch~3, the one mild departure from the trend of earlier convergence at larger budgets. Every run halts exactly two epochs after its minimum, confirming that patience-2 early stopping governed termination, while training loss continues to fall past the selected epoch. Pre-training development loss is $\approx 2.09$ (unpruned) and $\approx 2.20$ (pruned). These values determine only which checkpoint we evaluate; final quality is measured on the held-out test set (Section~\ref{sec:evaluation}), and development loss is never used as a quality signal.

\begin{table}[h]
\resizebox{\columnwidth}{!}{%
\centering\small\setlength{\tabcolsep}{4pt}
\begin{tabular}{llccccc}
\toprule
\textbf{Scheme} & \textbf{Cond.} & \textbf{2M} & \textbf{4M} & \textbf{8M} & \textbf{16M} & \textbf{32M} \\
\midrule
\multirow{2}{*}{Balanced} & pruned & 0.906 (3) & 0.826 (2) & 0.801 (2) & 0.760 (2) & 0.728 (3) \\
 & unpruned & 0.959 (3) & 0.845 (2) & 0.792 (2) & 0.777 (2) & 0.723 (3) \\
\midrule
\multirow{2}{*}{Natural} & pruned & 0.833 (3) & 0.749 (2) & 0.691 (2) & 0.636 (2) & 0.583 (2) \\
 & unpruned & 0.835 (3) & 0.760 (2) & 0.696 (2) & 0.640 (2) & 0.587 (2) \\
\bottomrule
\end{tabular}%
}
\caption{Checkpoint selection for all runs. Each cell gives the best development loss
and, in parentheses, the epoch at which it occurred (the selected checkpoint).
Development loss is used only for checkpoint selection, never as a quality measure
(Section~\ref{sec:evaluation}).}
\label{tab:app-trajectories}
\end{table}

\section{LLM Judge Protocol}
\label{judgeprotocol}
ROUGE-Lsum and BERTScore measure surface and embedding overlap, but can diverge from content coverage and faithfulness in structured-output domains (Section~\ref{sec:results}). To assess summary quality at the level of individual claims, we adopt the bidirectional atomic fact-checking framework of \cite{zhou2026large}, adapting it to our \{transcript, reference minute, generated minute\} setting. Following their protocol, we decompose minutes into atomic facts and use an LLM judge to check factual entailment, scoring three dimensions:  completeness (the fraction of reference-minute facts captured by the generated minute), faithfulness (the fraction of generated-minute facts supported by the transcript), and conciseness (the fraction of generated-minute facts that are salient with respect to the reference minute). Completeness and conciseness are the recall and precision of the generated minute against the reference; faithfulness is its precision against the source meeting. 

We depart from \cite{zhou2026large} in one respect where they extract a gold key-fact set from the source conversation via a dedicated human-LLM pipeline, we take the human-authored reference minute as the salience oracle, decomposing it directly into key facts. This suits our setting, where every meeting already carries an expert reference, and avoids a separate annotation stage; the trade-off is that any correct content absent from reference is neither credited (to completeness) nor considered salient (lowering conciseness). All judgments use a fixed rubric with explicit adjudication rules applied identically across mixtures,
budgets, and domains, and a judge model isolated from the fine-tuned systems to avoid self-preference bias. The judge (Qwen2.5-72B-Instruct-AWQ) runs under greedy decoding on a single NVIDIA H100; all other inference and fine-tuning use a single A100 (Section~\ref{sec:implementation})

\section{Human Validation of the LLM Judge}
\label{app:human-eval}

We validate the fact-level judge (Section~\ref{sec:evaluation}) against two human
annotators. The study measures how often the annotators agree with the judge's
per-fact labels and with each other; the human--human figure is the ceiling that
makes the human--judge figure interpretable, since a judge that agrees with humans
as often as they agree with each other is performing at human level.

\paragraph{Design.} Three choices follow. (i) Annotators adjudicate the
judge's own atomic-fact lists rather than generating their own, so labels align
fact-for-fact and a per-fact statistic is defined. (ii) We validate on
\texttt{balanced-32M} only, as the object of study is the judge, not the allocation
schemes. (iii) Facts are stratified by judge label: the judge's base rates
are skewed (faithfulness is $\sim$79\% \texttt{supported}), so under random sampling
a constant-majority annotator would appear to agree $\sim$79\% of the time, whereas
an even split makes such agreement worth only $\sim$50\%.

\paragraph{Sampling.} From the \texttt{balanced-32M} run we drew 30 test meetings
(six per domain, seed 42) and sampled facts within each (dimension $\times$ domain)
stratum: up to 30 judge-positive and 30 judge-negative facts per domain for
completeness and conciseness, and up to 25 each for faithfulness. Strata smaller than
the target were taken in full; six of thirty cells were capped by MeetingBank, whose
short minutes yield few facts, and further caps reflect genuine scarcity in the
judge's output (only 12 \texttt{not\_supported} facts exist across all six ICSI
meetings, the model being $\sim$92\% faithful there). This yields 741 items (257
completeness, 268 conciseness, 216 faithfulness). Because the caps bind hardest on
the majority corpora, the sample is deliberately not weighted to the test
distribution: the agreement figures are pooled across domains and stratified by
label, the right estimand for validating an instrument that must work on every
domain, not an estimate of agreement on the deployed test set.

\paragraph{Annotation.} Both annotators labelled every item, as $\kappa$ requires
paired decisions. Each saw the fact and only the material its task permits (the
generated minute for completeness, the transcript for faithfulness, the reference
for conciseness), with the judge's label stripped and items shuffled within meeting.
Faithfulness items were grouped by meeting so each transcript is read once. A
one-line reason was required on every item; annotators did not confer. The judge
labels compared here come from re-scoring each fact individually, matching the
annotators' condition; the aggregate scores in Table~\ref{tab:judge} use the batched
judge.

\begin{figure}[t]
  \centering
  \includegraphics[width=\columnwidth]{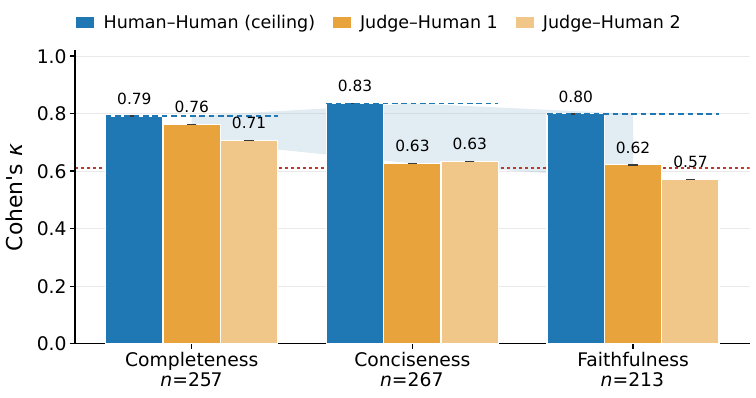}
  \caption{Inter-rater agreement on 741 judge-labelled facts from
    \texttt{balanced-32M} (Cohen's $\kappa$, bootstrap 95\% CIs, 4{,}000
    resamples). Human--human agreement is the ceiling (dashed rule); shading marks
    the judge's shortfall against it. The red dotted line is $\kappa = 0.61$, the
    threshold for substantial agreement. The judge reaches the human band on completeness but falls below it on conciseness and
    faithfulness.}
  \label{fig:agreement}
\end{figure}

\paragraph{Results.} Four items blank in one annotator are excluded pairwise
($n = 257/267/213$). Inter-annotator agreement is substantial on all three
dimensions ($\kappa = 0.79/0.84/0.80$; 91--92\% raw), establishing a reliable ground
truth. Against it the judge reaches $\kappa = 0.76/0.71$ on completeness, within the
human band, and $0.63/0.63$ and $0.62/0.57$ on conciseness and faithfulness,
substantial but below the ceiling (Figure~\ref{fig:agreement},
Table~\ref{tab:agreement}). The errors are directional: restricted to items where
both annotators agree, the judge over-accepts in 18/34/32 cases against 2/5/2 in the
other direction, so 87--94\% of its disagreements are over-acceptance. The absolute
scores in Table~\ref{tab:judge} are therefore upper bounds; because the offset is
applied by one judge under one rubric to every system and budget, the
balanced-versus-natural and 2M-versus-32M comparisons are unaffected.

\begin{table}[t]
  \centering
  \small
  \setlength{\tabcolsep}{4pt}
  \begin{tabular}{lccc}
    \toprule
    & \textbf{Compl.} & \textbf{Concis.} & \textbf{Faith.} \\
    $n$ & 257 & 267 & 213 \\
    \midrule
    \multicolumn{4}{l}{Cohen's $\kappa$, pairwise (bootstrap 95\% CI)} \\
    H1--H2   & 0.791\,\tiny[.71,.86] & 0.835\,\tiny[.77,.90] & 0.799\,\tiny[.71,.87] \\
    Judge--H1 & 0.762\,\tiny[.68,.84] & 0.627\,\tiny[.54,.71] & 0.622\,\tiny[.52,.72] \\
    Judge--H2 & 0.706\,\tiny[.61,.79] & 0.634\,\tiny[.54,.72] & 0.571\,\tiny[.47,.67] \\
    \midrule
    \multicolumn{4}{l}{Raw agreement, pairwise (\%)} \\
    H1--H2   & 91\% & 92\% & 91\% \\
    Judge--H1 & 89\% & 81\% & 81\% \\
    Judge--H2 & 86\% & 82\% & 78\% \\
    \midrule
    \multicolumn{4}{l}{Judge error direction (vs.\ H1$=$H2 consensus, counts)} \\
    over-accepts & 18 & 34 & 32 \\
    over-rejects & 2 & 5 & 2 \\
    \% lenient   & 90\% & 87\% & 94\% \\
    \bottomrule
  \end{tabular}
  \caption{Agreement between the two annotators (H1, H2) and the LLM judge, per
dimension. Top: chance-corrected agreement (Cohen's $\kappa$) for each rater
pair; H1--H2 is the human ceiling. Middle: the same pairs as raw percentage
agreement. Bottom: among items where the two annotators agree (a consensus
ground truth), the number of judge labels that are too generous (over-accept:
judge positive, humans negative) versus too strict (over-reject), showing the
judge errs by over-accepting in 87-94\% of cases.}
  \label{tab:agreement}
\end{table}

\subsection{Annotation guidelines}
\label{app:guidelines}

The guidelines below were given to both annotators verbatim. Every item is a
binary decision; annotators never write, summarise, or rate on a scale.

\paragraph{Materials.} The transcript is the raw record of what was said.
The reference minute is the summary a human wrote; it is treated as
authoritative and defines what was worth recording. The generated minute is
what the system produced. An atomic fact is a single indivisible claim
extracted from a minute: ``Sarah presented the Q3 budget and the team agreed to
postpone the launch'' becomes two facts. Annotators judge atomic facts, not whole
minutes.

\paragraph{Task A: Completeness.} Given one fact from the reference minute and
the full generated minute: does the generated minute contain this fact?
Label \texttt{captured} if the generated minute states the fact or states something
from which it follows directly and unambiguously; different wording is fine.
Label \texttt{not\_captured} if the fact is absent, contradicted, or only vaguely
gestured at. Losing the core value is \texttt{not\_captured}: mentioning that a
budget increase was agreed does not capture ``the increase was 12\%''. Extra detail
never hurts. Numbers, dates, and names must match. Use only the generated minute:
do not consult the transcript, and do not use outside knowledge to fill a gap the
generated minute left.

\paragraph{Task B: Faithfulness.} Given one fact from the generated minute and
the meeting transcript: does the transcript support this fact? Label
\texttt{supported} if the fact is explicitly stated or follows directly and
unambiguously from what was said; \texttt{not\_supported} if it is not said,
contradicted, or requires a leap the transcript does not license. Judge on the
transcript alone---do not mark a fact supported because it is plausible.
Over-specificity is \texttt{not\_supported} (transcript ``around 25 euros'', fact
``exactly 25 euros''), whereas hedging a precise figure is supported. Attribution
must itself be supported, not merely the content. Proposals are not decisions: a
fact asserting a decision the transcript only floated is \texttt{not\_supported}.
Long transcripts are truncated; if a fact refers to something that would have
occurred after the transcript ends, mark \texttt{not\_supported}, because only what
is visible can be judged.

\paragraph{Task C: Conciseness (salience).} Given one fact from the generated
minute and the reference minute: did the human who wrote the reference think
this was worth recording? Label \texttt{salient} if the reference states the fact
or states something from which it follows; \texttt{non\_salient} if it is absent
from the reference. This task is not a truth judgement. A fact may be perfectly
true of the meeting and still be \texttt{non\_salient}, because the minute-taker
chose not to record it: if the reference does not mention how long the meeting ran,
``the meeting lasted two hours'' is \texttt{non\_salient} even if it did. A fact
more specific than the reference on a point the reference does make is still
salient; a fact introducing content the reference does not touch at all is not. A
fact that merely restates another fact already matched to the same reference
content is \texttt{non\_salient}.

\paragraph{Procedure.} Each task is completed in full before the next is begun,
because the rules are deliberately different and switching between them causes
errors. For each item the annotator reads the fact, reads the permitted material,
chooses a label, and writes a one-line reason; the reasons are how disagreements
are understood afterwards. Annotators do not discuss items with each other, do not
look anything up, and do not leave items blank; where genuinely torn, they choose
the label they would defend and write ``borderline'' in the reason.

\begin{table*}[]
\centering
\small
\setlength{\tabcolsep}{5pt}
\begin{tabular}{llccccc|cc}
\toprule
\textbf{Condition} & \textbf{Metric} & \textbf{AMI} & \textbf{ICSI} & \textbf{ELITR} & \textbf{EPM} & \textbf{MB} & \textbf{Macro} & \textbf{Micro} \\
 & & \textit{20} & \textit{6} & \textit{38} & \textit{242} & \textit{862} & & \\
\midrule
\multirow{5}{*}{Pruned}
 & R-1    & $0.339_{\pm.044}$ & $0.310_{\pm.036}$ & $0.237_{\pm.016}$ & $0.277_{\pm.045}$ & $0.532_{\pm.008}$ & $0.339_{\pm.018}$ & $0.465_{\pm.009}$ \\
 & R-2    & $0.137_{\pm.016}$ & $0.075_{\pm.010}$ & $0.060_{\pm.007}$ & $0.165_{\pm.036}$ & $0.421_{\pm.007}$ & $0.172_{\pm.011}$ & $0.350_{\pm.007}$ \\
 & R-L    & $0.206_{\pm.028}$ & $0.170_{\pm.017}$ & $0.137_{\pm.003}$ & $0.223_{\pm.040}$ & $0.498_{\pm.008}$ & $0.247_{\pm.012}$ & $0.423_{\pm.008}$ \\
 & R-Lsum & $0.328_{\pm.044}$ & $0.292_{\pm.036}$ & $0.229_{\pm.014}$ & $0.271_{\pm.044}$ & $0.505_{\pm.008}$ & $0.325_{\pm.017}$ & $0.443_{\pm.009}$ \\
 & BERT   & $0.861_{\pm.004}$ & $0.834_{\pm.003}$ & $0.826_{\pm.006}$ & $0.833_{\pm.013}$ & $0.898_{\pm.000}$ & $0.850_{\pm.004}$ & $0.881_{\pm.003}$ \\
\midrule
\multirow{5}{*}{Unpruned}
 & R-1    & $0.370_{\pm.056}$ & $0.272_{\pm.031}$ & $0.220_{\pm.011}$ & $0.273_{\pm.041}$ & $0.534_{\pm.008}$ & $0.334_{\pm.014}$ & $0.466_{\pm.009}$ \\
 & R-2    & $0.155_{\pm.030}$ & $0.060_{\pm.007}$ & $0.058_{\pm.003}$ & $0.164_{\pm.030}$ & $0.426_{\pm.006}$ & $0.173_{\pm.010}$ & $0.353_{\pm.008}$ \\
 & R-L    & $0.223_{\pm.032}$ & $0.157_{\pm.016}$ & $0.129_{\pm.004}$ & $0.222_{\pm.034}$ & $0.502_{\pm.006}$ & $0.247_{\pm.012}$ & $0.425_{\pm.009}$ \\
 & R-Lsum & $0.357_{\pm.056}$ & $0.256_{\pm.028}$ & $0.213_{\pm.009}$ & $0.267_{\pm.041}$ & $0.508_{\pm.007}$ & $0.320_{\pm.016}$ & $0.445_{\pm.009}$ \\
 & BERTScore  & $0.863_{\pm.004}$ & $0.830_{\pm.005}$ & $0.824_{\pm.006}$ & $0.829_{\pm.010}$ & $0.899_{\pm.001}$ & $0.849_{\pm.004}$ & $0.881_{\pm.003}$ \\
\bottomrule
\end{tabular}
\caption{Example-level results across all five metrics: mean$_{\pm\text{std}}$ over seeds 1/2/15, per condition. Column italics are test-set meeting counts. Seed variation is small for aggregate measures (Macro/Micro std $\leq$0.018); minority domains with few test meetings (ICSI, AMI) show larger per-seed spread. R-1/2/L = ROUGE-1/2/L.}
\label{tab:app-ex-full}
\end{table*}

\section{Full Per-Domain Results}
\label{app:full-results}

This appendix reports the complete per-domain grid over the full metric family
(ROUGE-1/2/L/Lsum and BERTScore-F1), of which
Section~\ref{sec:results} presents the
ROUGE-Lsum${+}$BERTScore view that carries the finding.

%%%%zero-shot

%%%%%%%%%%%%%%%%%% PRUNED START %%%%%%%%%%%%%%%%%%%%%

\begin{table*}[t]
\centering\tiny\setlength{\tabcolsep}{4pt}
\begin{tabular}{llccccccc}
\toprule
\textbf{Metric} & \textbf{Budget} & \textbf{AMI} & \textbf{ICSI} & \textbf{ELITR} & \textbf{EPM} & \textbf{MB} & \textbf{Macro} & \textbf{Micro} \\
 & & \textit{20} & \textit{6} & \textit{38} & \textit{242} & \textit{862} & & \\
\midrule
\multicolumn{9}{c}{\textit{Balanced (equal-token) allocation}} \\
\midrule
\multirow{5}{*}{ROUGE-1} & 2M & $0.389_{[0.30,0.47]}$ & $0.272_{[0.24,0.30]}$ & $0.159_{[0.12,0.20]}$ & $0.227_{[0.20,0.25]}$ & $0.585_{[0.57,0.60]}$ & $0.326_{[0.31,0.35]}$ & $0.492_{[0.48,0.51]}$ \\
 & 4M & $0.359_{[0.27,0.45]}$ & $0.224_{[0.12,0.33]}$ & $0.166_{[0.13,0.21]}$ & $0.326_{[0.29,0.36]}$ & $0.587_{[0.57,0.61]}$ & $0.333_{[0.30,0.36]}$ & $0.514_{[0.50,0.53]}$ \\
 & 8M & $0.465_{[0.40,0.52]}$ & $0.443_{[0.40,0.49]}$ & $0.251_{[0.20,0.30]}$ & $0.443_{[0.41,0.47]}$ & $0.626_{[0.61,0.64]}$ & $0.446_{[0.43,0.46]}$ & $0.572_{[0.56,0.59]}$ \\
 & 16M & $0.522_{[0.50,0.55]}$ & $0.440_{[0.41,0.47]}$ & $0.383_{[0.34,0.42]}$ & $0.392_{[0.36,0.42]}$ & $0.649_{[0.63,0.67]}$ & $0.477_{[0.46,0.49]}$ & $0.584_{[0.57,0.60]}$ \\
 & 32M & $0.516_{[0.49,0.54]}$ & $0.465_{[0.42,0.50]}$ & $0.379_{[0.35,0.41]}$ & $0.554_{[0.52,0.58]}$ & $0.670_{[0.65,0.69]}$ & $0.517_{[0.50,0.53]}$ & $0.633_{[0.62,0.65]}$ \\
\cmidrule(lr){1-9}
\multirow{5}{*}{ROUGE-2} & 2M & $0.162_{[0.12,0.21]}$ & $0.054_{[0.04,0.06]}$ & $0.039_{[0.03,0.05]}$ & $0.145_{[0.13,0.16]}$ & $0.480_{[0.46,0.50]}$ & $0.176_{[0.17,0.19]}$ & $0.389_{[0.37,0.40]}$ \\
 & 4M & $0.150_{[0.10,0.20]}$ & $0.048_{[0.03,0.07]}$ & $0.039_{[0.03,0.05]}$ & $0.220_{[0.19,0.25]}$ & $0.480_{[0.46,0.50]}$ & $0.188_{[0.18,0.20]}$ & $0.404_{[0.39,0.42]}$ \\
 & 8M & $0.179_{[0.14,0.22]}$ & $0.121_{[0.10,0.14]}$ & $0.064_{[0.05,0.08]}$ & $0.315_{[0.29,0.34]}$ & $0.516_{[0.50,0.53]}$ & $0.239_{[0.23,0.25]}$ & $0.452_{[0.44,0.47]}$ \\
 & 16M & $0.193_{[0.17,0.22]}$ & $0.100_{[0.08,0.11]}$ & $0.112_{[0.09,0.13]}$ & $0.271_{[0.25,0.30]}$ & $0.553_{[0.53,0.57]}$ & $0.245_{[0.24,0.26]}$ & $0.471_{[0.46,0.49]}$ \\
 & 32M & $0.195_{[0.17,0.22]}$ & $0.115_{[0.10,0.14]}$ & $0.099_{[0.08,0.11]}$ & $0.407_{[0.38,0.43]}$ & $0.572_{[0.55,0.59]}$ & $0.278_{[0.27,0.29]}$ & $0.514_{[0.50,0.53]}$ \\
\cmidrule(lr){1-9}
\multirow{5}{*}{ROUGE-L} & 2M & $0.243_{[0.19,0.30]}$ & $0.158_{[0.15,0.17]}$ & $0.101_{[0.08,0.12]}$ & $0.194_{[0.17,0.22]}$ & $0.553_{[0.53,0.57]}$ & $0.250_{[0.24,0.26]}$ & $0.457_{[0.44,0.47]}$ \\
 & 4M & $0.211_{[0.16,0.26]}$ & $0.118_{[0.08,0.16]}$ & $0.106_{[0.09,0.13]}$ & $0.281_{[0.25,0.31]}$ & $0.553_{[0.53,0.57]}$ & $0.254_{[0.24,0.27]}$ & $0.474_{[0.46,0.49]}$ \\
 & 8M & $0.256_{[0.22,0.29]}$ & $0.198_{[0.18,0.22]}$ & $0.143_{[0.12,0.17]}$ & $0.395_{[0.37,0.42]}$ & $0.590_{[0.57,0.61]}$ & $0.316_{[0.30,0.33]}$ & $0.527_{[0.51,0.54]}$ \\
 & 16M & $0.289_{[0.27,0.31]}$ & $0.184_{[0.17,0.20]}$ & $0.194_{[0.17,0.21]}$ & $0.340_{[0.31,0.37]}$ & $0.623_{[0.60,0.64]}$ & $0.326_{[0.32,0.34]}$ & $0.543_{[0.53,0.56]}$ \\
 & 32M & $0.274_{[0.26,0.29]}$ & $0.192_{[0.18,0.21]}$ & $0.184_{[0.17,0.20]}$ & $0.486_{[0.46,0.51]}$ & $0.641_{[0.62,0.66]}$ & $0.355_{[0.35,0.36]}$ & $0.586_{[0.57,0.60]}$ \\
\cmidrule(lr){1-9}
\multirow{5}{*}{ROUGE-Lsum} & 2M & $0.376_{[0.29,0.45]}$ & $0.258_{[0.22,0.29]}$ & $0.152_{[0.12,0.19]}$ & $0.222_{[0.20,0.25]}$ & $0.562_{[0.54,0.58]}$ & $0.314_{[0.29,0.33]}$ & $0.473_{[0.46,0.49]}$ \\
 & 4M & $0.347_{[0.26,0.43]}$ & $0.211_{[0.12,0.32]}$ & $0.159_{[0.12,0.20]}$ & $0.320_{[0.29,0.35]}$ & $0.561_{[0.54,0.58]}$ & $0.320_{[0.29,0.35]}$ & $0.492_{[0.48,0.51]}$ \\
 & 8M & $0.449_{[0.38,0.51]}$ & $0.419_{[0.38,0.47]}$ & $0.244_{[0.20,0.29]}$ & $0.438_{[0.41,0.47]}$ & $0.598_{[0.58,0.62]}$ & $0.429_{[0.41,0.45]}$ & $0.550_{[0.54,0.56]}$ \\
 & 16M & $0.502_{[0.48,0.53]}$ & $0.420_{[0.40,0.45]}$ & $0.373_{[0.33,0.41]}$ & $0.386_{[0.36,0.42]}$ & $0.630_{[0.61,0.65]}$ & $0.462_{[0.45,0.48]}$ & $0.568_{[0.55,0.58]}$ \\
 & 32M & $0.495_{[0.47,0.52]}$ & $0.445_{[0.40,0.49]}$ & $0.367_{[0.33,0.40]}$ & $0.547_{[0.52,0.57]}$ & $0.648_{[0.63,0.67]}$ & $0.500_{[0.49,0.51]}$ & $0.615_{[0.60,0.63]}$ \\
\cmidrule(lr){1-9}
\multirow{5}{*}{BERTScore} & 2M & $0.861_{[0.85,0.87]}$ & $0.827_{[0.82,0.84]}$ & $0.806_{[0.80,0.82]}$ & $0.832_{[0.83,0.84]}$ & $0.910_{[0.91,0.91]}$ & $0.847_{[0.84,0.85]}$ & $0.890_{[0.89,0.89]}$ \\
 & 4M & $0.863_{[0.85,0.87]}$ & $0.818_{[0.80,0.84]}$ & $0.818_{[0.81,0.83]}$ & $0.838_{[0.83,0.85]}$ & $0.912_{[0.91,0.92]}$ & $0.850_{[0.84,0.85]}$ & $0.892_{[0.89,0.90]}$ \\
 & 8M & $0.870_{[0.86,0.88]}$ & $0.851_{[0.84,0.86]}$ & $0.835_{[0.83,0.84]}$ & $0.865_{[0.86,0.87]}$ & $0.918_{[0.91,0.92]}$ & $0.868_{[0.86,0.87]}$ & $0.903_{[0.90,0.91]}$ \\
 & 16M & $0.878_{[0.87,0.88]}$ & $0.847_{[0.84,0.86]}$ & $0.847_{[0.84,0.85]}$ & $0.855_{[0.85,0.86]}$ & $0.925_{[0.92,0.93]}$ & $0.870_{[0.87,0.87]}$ & $0.907_{[0.90,0.91]}$ \\
 & 32M & $0.875_{[0.87,0.88]}$ & $0.852_{[0.85,0.86]}$ & $0.846_{[0.84,0.85]}$ & $0.887_{[0.88,0.89]}$ & $0.928_{[0.92,0.93]}$ & $0.877_{[0.87,0.88]}$ & $0.915_{[0.91,0.92]}$ \\
\midrule
\multicolumn{9}{c}{\textit{Natural (proportional) allocation}} \\
\midrule
\multirow{5}{*}{ROUGE-1} & 2M & $0.344_{[0.28,0.40]}$ & $0.137_{[0.08,0.20]}$ & $0.189_{[0.16,0.22]}$ & $0.280_{[0.25,0.31]}$ & $0.613_{[0.59,0.63]}$ & $0.312_{[0.29,0.33]}$ & $0.523_{[0.51,0.54]}$ \\
 & 4M & $0.275_{[0.21,0.35]}$ & $0.087_{[0.06,0.11]}$ & $0.130_{[0.10,0.16]}$ & $0.454_{[0.42,0.48]}$ & $0.641_{[0.62,0.66]}$ & $0.317_{[0.30,0.34]}$ & $0.577_{[0.56,0.59]}$ \\
 & 8M & $0.242_{[0.17,0.32]}$ & $0.116_{[0.08,0.14]}$ & $0.116_{[0.09,0.15]}$ & $0.524_{[0.50,0.55]}$ & $0.670_{[0.65,0.69]}$ & $0.333_{[0.31,0.35]}$ & $0.611_{[0.60,0.62]}$ \\
 & 16M & $0.298_{[0.21,0.39]}$ & $0.171_{[0.08,0.27]}$ & $0.159_{[0.12,0.20]}$ & $0.530_{[0.50,0.56]}$ & $0.702_{[0.69,0.72]}$ & $0.372_{[0.34,0.40]}$ & $0.639_{[0.62,0.65]}$ \\
 & 32M & $0.410_{[0.33,0.48]}$ & $0.297_{[0.23,0.36]}$ & $0.212_{[0.17,0.26]}$ & $0.539_{[0.51,0.57]}$ & $0.713_{[0.70,0.73]}$ & $0.434_{[0.41,0.46]}$ & $0.653_{[0.64,0.67]}$ \\
\cmidrule(lr){1-9}
\multirow{5}{*}{ROUGE-2} & 2M & $0.110_{[0.08,0.14]}$ & $0.027_{[0.01,0.04]}$ & $0.040_{[0.03,0.05]}$ & $0.179_{[0.16,0.20]}$ & $0.507_{[0.49,0.53]}$ & $0.172_{[0.16,0.18]}$ & $0.414_{[0.40,0.43]}$ \\
 & 4M & $0.094_{[0.06,0.13]}$ & $0.017_{[0.01,0.02]}$ & $0.028_{[0.02,0.04]}$ & $0.319_{[0.29,0.34]}$ & $0.538_{[0.52,0.56]}$ & $0.199_{[0.19,0.21]}$ & $0.466_{[0.45,0.48]}$ \\
 & 8M & $0.085_{[0.05,0.12]}$ & $0.018_{[0.02,0.02]}$ & $0.023_{[0.02,0.03]}$ & $0.375_{[0.35,0.40]}$ & $0.569_{[0.55,0.59]}$ & $0.214_{[0.20,0.22]}$ & $0.500_{[0.48,0.51]}$ \\
 & 16M & $0.123_{[0.08,0.17]}$ & $0.033_{[0.01,0.06]}$ & $0.046_{[0.03,0.06]}$ & $0.393_{[0.36,0.42]}$ & $0.600_{[0.58,0.62]}$ & $0.239_{[0.23,0.25]}$ & $0.528_{[0.51,0.54]}$ \\
 & 32M & $0.162_{[0.12,0.20]}$ & $0.065_{[0.05,0.08]}$ & $0.058_{[0.04,0.08]}$ & $0.402_{[0.37,0.43]}$ & $0.621_{[0.60,0.64]}$ & $0.262_{[0.25,0.27]}$ & $0.547_{[0.53,0.56]}$ \\
\cmidrule(lr){1-9}
\multirow{5}{*}{ROUGE-L} & 2M & $0.202_{[0.17,0.23]}$ & $0.094_{[0.06,0.13]}$ & $0.115_{[0.10,0.13]}$ & $0.232_{[0.21,0.26]}$ & $0.579_{[0.56,0.60]}$ & $0.244_{[0.23,0.26]}$ & $0.483_{[0.47,0.50]}$ \\
 & 4M & $0.172_{[0.13,0.22]}$ & $0.075_{[0.05,0.10]}$ & $0.085_{[0.07,0.10]}$ & $0.398_{[0.37,0.43]}$ & $0.611_{[0.59,0.63]}$ & $0.268_{[0.26,0.28]}$ & $0.539_{[0.52,0.55]}$ \\
 & 8M & $0.160_{[0.11,0.21]}$ & $0.081_{[0.06,0.09]}$ & $0.075_{[0.06,0.10]}$ & $0.459_{[0.43,0.48]}$ & $0.639_{[0.62,0.66]}$ & $0.283_{[0.27,0.30]}$ & $0.572_{[0.56,0.59]}$ \\
 & 16M & $0.186_{[0.13,0.24]}$ & $0.114_{[0.07,0.16]}$ & $0.105_{[0.08,0.13]}$ & $0.471_{[0.44,0.50]}$ & $0.670_{[0.65,0.69]}$ & $0.309_{[0.29,0.33]}$ & $0.600_{[0.59,0.61]}$ \\
 & 32M & $0.247_{[0.20,0.29]}$ & $0.160_{[0.13,0.19]}$ & $0.134_{[0.11,0.16]}$ & $0.483_{[0.45,0.51]}$ & $0.688_{[0.67,0.71]}$ & $0.342_{[0.33,0.36]}$ & $0.617_{[0.60,0.63]}$ \\
\cmidrule(lr){1-9}
\multirow{5}{*}{ROUGE-Lsum} & 2M & $0.331_{[0.27,0.39]}$ & $0.129_{[0.08,0.18]}$ & $0.184_{[0.15,0.22]}$ & $0.274_{[0.25,0.30]}$ & $0.586_{[0.57,0.60]}$ & $0.301_{[0.28,0.32]}$ & $0.502_{[0.49,0.52]}$ \\
 & 4M & $0.263_{[0.20,0.33]}$ & $0.080_{[0.06,0.11]}$ & $0.124_{[0.09,0.16]}$ & $0.448_{[0.42,0.48]}$ & $0.618_{[0.60,0.64]}$ & $0.306_{[0.29,0.32]}$ & $0.558_{[0.54,0.57]}$ \\
 & 8M & $0.230_{[0.16,0.31]}$ & $0.107_{[0.08,0.13]}$ & $0.107_{[0.08,0.14]}$ & $0.516_{[0.49,0.54]}$ & $0.646_{[0.63,0.66]}$ & $0.321_{[0.30,0.34]}$ & $0.592_{[0.58,0.61]}$ \\
 & 16M & $0.287_{[0.20,0.37]}$ & $0.158_{[0.07,0.25]}$ & $0.153_{[0.11,0.19]}$ & $0.523_{[0.49,0.55]}$ & $0.678_{[0.66,0.69]}$ & $0.360_{[0.33,0.39]}$ & $0.620_{[0.61,0.63]}$ \\
 & 32M & $0.393_{[0.32,0.46]}$ & $0.281_{[0.21,0.35]}$ & $0.204_{[0.16,0.25]}$ & $0.534_{[0.50,0.56]}$ & $0.694_{[0.68,0.71]}$ & $0.421_{[0.40,0.44]}$ & $0.637_{[0.62,0.65]}$ \\
\cmidrule(lr){1-9}
\multirow{5}{*}{BERTScore} & 2M & $0.851_{[0.84,0.86]}$ & $0.814_{[0.79,0.83]}$ & $0.806_{[0.80,0.81]}$ & $0.831_{[0.82,0.84]}$ & $0.915_{[0.91,0.92]}$ & $0.843_{[0.84,0.85]}$ & $0.892_{[0.89,0.90]}$ \\
 & 4M & $0.843_{[0.83,0.85]}$ & $0.779_{[0.77,0.79]}$ & $0.796_{[0.79,0.81]}$ & $0.872_{[0.86,0.88]}$ & $0.922_{[0.92,0.93]}$ & $0.843_{[0.84,0.85]}$ & $0.905_{[0.90,0.91]}$ \\
 & 8M & $0.835_{[0.82,0.85]}$ & $0.801_{[0.79,0.81]}$ & $0.815_{[0.80,0.83]}$ & $0.882_{[0.88,0.89]}$ & $0.925_{[0.92,0.93]}$ & $0.852_{[0.85,0.86]}$ & $0.910_{[0.91,0.91]}$ \\
 & 16M & $0.852_{[0.84,0.87]}$ & $0.806_{[0.78,0.82]}$ & $0.819_{[0.81,0.83]}$ & $0.884_{[0.88,0.89]}$ & $0.935_{[0.93,0.94]}$ & $0.859_{[0.85,0.86]}$ & $0.919_{[0.92,0.92]}$ \\
 & 32M & $0.865_{[0.85,0.88]}$ & $0.833_{[0.83,0.84]}$ & $0.837_{[0.83,0.85]}$ & $0.889_{[0.88,0.90]}$ & $0.938_{[0.93,0.94]}$ & $0.872_{[0.87,0.88]}$ & $0.923_{[0.92,0.93]}$ \\
\bottomrule
\end{tabular}
\caption{Full per-domain results (\textbf{pruned}, seed~42) across the complete metric family (ROUGE-1/2/L/Lsum and BERTScore-F1), with bootstrap 95\% CIs in subscripts. Rows are grouped by allocation scheme (Balanced vs.\ Natural) and, within each scheme, by metric; each metric block reports the five-budget ladder (2-32M). Column italics are test-set meeting counts. The per-domain crossover of RQ1 is visible in every metric: Balanced leads on the minority domains (AMI/ICSI/ELITR) while Natural leads on the majority domains (EPM/MB), with the gap widest at small budgets.}
\label{tab:app-full-pruned}
\end{table*}

%%%%%%%%%%%%%%%%%% PRUNED END %%%%%%%%%%%%%%%%%%%%%

%%%%%%%%%%%%%%%%%%%%% UNPRUNED START %%%%%%%%%%%%%%%%%%%%%%%

\begin{table*}[t]
\centering\tiny\setlength{\tabcolsep}{4pt}
\begin{tabular}{llccccccc}
\toprule
\textbf{Metric} & \textbf{Budget} & \textbf{AMI} & \textbf{ICSI} & \textbf{ELITR} & \textbf{EPM} & \textbf{MB} & \textbf{Macro} & \textbf{Micro} \\
 & & \textit{20} & \textit{6} & \textit{38} & \textit{242} & \textit{862} & & \\
\midrule
\multicolumn{9}{c}{\textit{Balanced (equal-token) allocation}} \\
\midrule
\multirow{5}{*}{ROUGE-1} & 2M & $0.281_{[0.20,0.37]}$ & $0.113_{[0.08,0.14]}$ & $0.143_{[0.11,0.18]}$ & $0.332_{[0.31,0.35]}$ & $0.559_{[0.54,0.58]}$ & $0.285_{[0.27,0.31]}$ & $0.491_{[0.48,0.51]}$ \\
 & 4M & $0.318_{[0.23,0.41]}$ & $0.260_{[0.17,0.34]}$ & $0.174_{[0.13,0.22]}$ & $0.189_{[0.17,0.21]}$ & $0.615_{[0.60,0.63]}$ & $0.311_{[0.28,0.34]}$ & $0.505_{[0.49,0.52]}$ \\
 & 8M & $0.393_{[0.31,0.47]}$ & $0.367_{[0.27,0.44]}$ & $0.241_{[0.20,0.28]}$ & $0.406_{[0.37,0.44]}$ & $0.637_{[0.62,0.65]}$ & $0.409_{[0.38,0.43]}$ & $0.571_{[0.56,0.58]}$ \\
 & 16M & $0.508_{[0.47,0.54]}$ & $0.449_{[0.40,0.50]}$ & $0.306_{[0.26,0.35]}$ & $0.354_{[0.32,0.38]}$ & $0.664_{[0.65,0.68]}$ & $0.456_{[0.44,0.47]}$ & $0.585_{[0.57,0.60]}$ \\
 & 32M & $0.521_{[0.50,0.54]}$ & $0.459_{[0.41,0.50]}$ & $0.382_{[0.34,0.42]}$ & $0.541_{[0.51,0.57]}$ & $0.678_{[0.66,0.69]}$ & $0.516_{[0.50,0.53]}$ & $0.636_{[0.62,0.65]}$ \\
\cmidrule(lr){1-9}
\multirow{5}{*}{ROUGE-2} & 2M & $0.120_{[0.08,0.17]}$ & $0.020_{[0.01,0.03]}$ & $0.029_{[0.02,0.04]}$ & $0.196_{[0.18,0.21]}$ & $0.453_{[0.43,0.47]}$ & $0.164_{[0.15,0.17]}$ & $0.378_{[0.36,0.39]}$ \\
 & 4M & $0.144_{[0.09,0.20]}$ & $0.065_{[0.04,0.09]}$ & $0.045_{[0.03,0.06]}$ & $0.120_{[0.11,0.14]}$ & $0.513_{[0.49,0.53]}$ & $0.177_{[0.16,0.19]}$ & $0.408_{[0.39,0.42]}$ \\
 & 8M & $0.158_{[0.12,0.20]}$ & $0.084_{[0.05,0.11]}$ & $0.062_{[0.05,0.08]}$ & $0.286_{[0.26,0.31]}$ & $0.533_{[0.51,0.55]}$ & $0.225_{[0.21,0.24]}$ & $0.458_{[0.44,0.47]}$ \\
 & 16M & $0.189_{[0.16,0.21]}$ & $0.101_{[0.09,0.12]}$ & $0.081_{[0.07,0.10]}$ & $0.249_{[0.22,0.27]}$ & $0.563_{[0.54,0.58]}$ & $0.237_{[0.23,0.25]}$ & $0.473_{[0.46,0.49]}$ \\
 & 32M & $0.196_{[0.17,0.22]}$ & $0.109_{[0.10,0.12]}$ & $0.106_{[0.09,0.12]}$ & $0.389_{[0.36,0.41]}$ & $0.578_{[0.56,0.60]}$ & $0.276_{[0.27,0.28]}$ & $0.515_{[0.50,0.53]}$ \\
\cmidrule(lr){1-9}
\multirow{5}{*}{ROUGE-L} & 2M & $0.176_{[0.12,0.24]}$ & $0.081_{[0.06,0.10]}$ & $0.095_{[0.08,0.12]}$ & $0.263_{[0.25,0.28]}$ & $0.526_{[0.51,0.55]}$ & $0.228_{[0.21,0.24]}$ & $0.449_{[0.43,0.46]}$ \\
 & 4M & $0.203_{[0.14,0.26]}$ & $0.148_{[0.10,0.19]}$ & $0.112_{[0.09,0.14]}$ & $0.159_{[0.14,0.18]}$ & $0.587_{[0.57,0.61]}$ & $0.242_{[0.23,0.26]}$ & $0.474_{[0.46,0.49]}$ \\
 & 8M & $0.231_{[0.18,0.28]}$ & $0.183_{[0.14,0.21]}$ & $0.133_{[0.11,0.15]}$ & $0.359_{[0.33,0.39]}$ & $0.606_{[0.59,0.62]}$ & $0.302_{[0.29,0.32]}$ & $0.531_{[0.52,0.55]}$ \\
 & 16M & $0.274_{[0.25,0.30]}$ & $0.188_{[0.18,0.20]}$ & $0.161_{[0.14,0.18]}$ & $0.309_{[0.28,0.34]}$ & $0.635_{[0.62,0.65]}$ & $0.313_{[0.30,0.32]}$ & $0.544_{[0.53,0.56]}$ \\
 & 32M & $0.289_{[0.27,0.31]}$ & $0.200_{[0.19,0.21]}$ & $0.186_{[0.17,0.20]}$ & $0.473_{[0.45,0.50]}$ & $0.648_{[0.63,0.67]}$ & $0.359_{[0.35,0.37]}$ & $0.589_{[0.57,0.60]}$ \\
\cmidrule(lr){1-9}
\multirow{5}{*}{ROUGE-Lsum} & 2M & $0.272_{[0.19,0.36]}$ & $0.104_{[0.08,0.13]}$ & $0.135_{[0.11,0.17]}$ & $0.324_{[0.30,0.34]}$ & $0.531_{[0.51,0.55]}$ & $0.273_{[0.25,0.29]}$ & $0.469_{[0.45,0.48]}$ \\
 & 4M & $0.306_{[0.22,0.40]}$ & $0.247_{[0.16,0.32]}$ & $0.166_{[0.13,0.21]}$ & $0.185_{[0.16,0.21]}$ & $0.594_{[0.58,0.61]}$ & $0.300_{[0.27,0.33]}$ & $0.488_{[0.47,0.50]}$ \\
 & 8M & $0.378_{[0.30,0.46]}$ & $0.352_{[0.25,0.42]}$ & $0.235_{[0.20,0.28]}$ & $0.400_{[0.37,0.43]}$ & $0.614_{[0.60,0.63]}$ & $0.396_{[0.37,0.42]}$ & $0.551_{[0.54,0.57]}$ \\
 & 16M & $0.486_{[0.45,0.52]}$ & $0.423_{[0.38,0.47]}$ & $0.296_{[0.25,0.34]}$ & $0.349_{[0.32,0.38]}$ & $0.642_{[0.63,0.66]}$ & $0.439_{[0.42,0.46]}$ & $0.566_{[0.55,0.58]}$ \\
 & 32M & $0.499_{[0.48,0.52]}$ & $0.436_{[0.39,0.48]}$ & $0.369_{[0.33,0.40]}$ & $0.535_{[0.51,0.56]}$ & $0.655_{[0.64,0.67]}$ & $0.499_{[0.48,0.51]}$ & $0.617_{[0.60,0.63]}$ \\
\cmidrule(lr){1-9}
\multirow{5}{*}{BERTScore} & 2M & $0.855_{[0.84,0.87]}$ & $0.804_{[0.80,0.81]}$ & $0.805_{[0.80,0.81]}$ & $0.837_{[0.83,0.84]}$ & $0.904_{[0.90,0.91]}$ & $0.841_{[0.84,0.84]}$ & $0.886_{[0.88,0.89]}$ \\
 & 4M & $0.857_{[0.84,0.87]}$ & $0.829_{[0.81,0.85]}$ & $0.823_{[0.81,0.83]}$ & $0.822_{[0.82,0.83]}$ & $0.917_{[0.91,0.92]}$ & $0.850_{[0.84,0.85]}$ & $0.893_{[0.89,0.90]}$ \\
 & 8M & $0.870_{[0.86,0.88]}$ & $0.839_{[0.83,0.85]}$ & $0.835_{[0.83,0.84]}$ & $0.860_{[0.85,0.87]}$ & $0.921_{[0.92,0.92]}$ & $0.865_{[0.86,0.87]}$ & $0.904_{[0.90,0.91]}$ \\
 & 16M & $0.874_{[0.87,0.88]}$ & $0.851_{[0.85,0.86]}$ & $0.835_{[0.83,0.84]}$ & $0.846_{[0.84,0.85]}$ & $0.926_{[0.92,0.93]}$ & $0.866_{[0.86,0.87]}$ & $0.905_{[0.90,0.91]}$ \\
 & 32M & $0.877_{[0.87,0.88]}$ & $0.849_{[0.84,0.86]}$ & $0.844_{[0.84,0.85]}$ & $0.883_{[0.88,0.89]}$ & $0.928_{[0.92,0.93]}$ & $0.876_{[0.87,0.88]}$ & $0.915_{[0.91,0.92]}$ \\
\midrule
\multicolumn{9}{c}{\textit{Natural (proportional) allocation}} \\
\midrule
\multirow{5}{*}{ROUGE-1} & 2M & $0.177_{[0.13,0.22]}$ & $0.087_{[0.03,0.15]}$ & $0.150_{[0.12,0.19]}$ & $0.438_{[0.41,0.46]}$ & $0.620_{[0.60,0.64]}$ & $0.294_{[0.28,0.31]}$ & $0.557_{[0.54,0.57]}$ \\
 & 4M & $0.243_{[0.16,0.33]}$ & $0.186_{[0.09,0.28]}$ & $0.129_{[0.10,0.16]}$ & $0.370_{[0.34,0.40]}$ & $0.625_{[0.61,0.64]}$ & $0.311_{[0.28,0.34]}$ & $0.547_{[0.53,0.56]}$ \\
 & 8M & $0.266_{[0.18,0.35]}$ & $0.223_{[0.13,0.32]}$ & $0.139_{[0.10,0.18]}$ & $0.373_{[0.34,0.41]}$ & $0.667_{[0.65,0.68]}$ & $0.334_{[0.31,0.36]}$ & $0.580_{[0.56,0.59]}$ \\
 & 16M & $0.280_{[0.20,0.36]}$ & $0.135_{[0.08,0.21]}$ & $0.158_{[0.12,0.19]}$ & $0.502_{[0.47,0.54]}$ & $0.697_{[0.68,0.71]}$ & $0.354_{[0.33,0.38]}$ & $0.629_{[0.61,0.64]}$ \\
 & 32M & $0.341_{[0.25,0.43]}$ & $0.235_{[0.14,0.33]}$ & $0.183_{[0.15,0.22]}$ & $0.500_{[0.47,0.53]}$ & $0.719_{[0.70,0.74]}$ & $0.396_{[0.37,0.42]}$ & $0.647_{[0.63,0.66]}$ \\
\cmidrule(lr){1-9}
\multirow{5}{*}{ROUGE-2} & 2M & $0.050_{[0.03,0.07]}$ & $0.016_{[0.00,0.03]}$ & $0.034_{[0.02,0.05]}$ & $0.287_{[0.27,0.31]}$ & $0.502_{[0.48,0.52]}$ & $0.178_{[0.17,0.19]}$ & $0.432_{[0.42,0.45]}$ \\
 & 4M & $0.088_{[0.05,0.13]}$ & $0.028_{[0.01,0.04]}$ & $0.027_{[0.02,0.04]}$ & $0.253_{[0.23,0.28]}$ & $0.521_{[0.50,0.54]}$ & $0.183_{[0.17,0.19]}$ & $0.439_{[0.42,0.45]}$ \\
 & 8M & $0.104_{[0.06,0.15]}$ & $0.046_{[0.02,0.07]}$ & $0.026_{[0.02,0.04]}$ & $0.272_{[0.24,0.30]}$ & $0.564_{[0.54,0.58]}$ & $0.202_{[0.19,0.21]}$ & $0.475_{[0.46,0.49]}$ \\
 & 16M & $0.110_{[0.07,0.15]}$ & $0.028_{[0.01,0.05]}$ & $0.046_{[0.04,0.06]}$ & $0.378_{[0.35,0.41]}$ & $0.606_{[0.59,0.63]}$ & $0.234_{[0.22,0.25]}$ & $0.529_{[0.51,0.54]}$ \\
 & 32M & $0.147_{[0.10,0.20]}$ & $0.053_{[0.03,0.08]}$ & $0.042_{[0.03,0.05]}$ & $0.380_{[0.35,0.41]}$ & $0.625_{[0.61,0.64]}$ & $0.249_{[0.24,0.26]}$ & $0.544_{[0.53,0.56]}$ \\
\cmidrule(lr){1-9}
\multirow{5}{*}{ROUGE-L} & 2M & $0.111_{[0.08,0.14]}$ & $0.059_{[0.02,0.10]}$ & $0.097_{[0.08,0.12]}$ & $0.371_{[0.35,0.40]}$ & $0.581_{[0.56,0.60]}$ & $0.244_{[0.23,0.26]}$ & $0.511_{[0.50,0.53]}$ \\
 & 4M & $0.161_{[0.12,0.21]}$ & $0.119_{[0.07,0.16]}$ & $0.094_{[0.08,0.11]}$ & $0.322_{[0.29,0.35]}$ & $0.590_{[0.57,0.61]}$ & $0.257_{[0.24,0.27]}$ & $0.508_{[0.49,0.52]}$ \\
 & 8M & $0.169_{[0.12,0.22]}$ & $0.137_{[0.09,0.18]}$ & $0.084_{[0.07,0.10]}$ & $0.334_{[0.30,0.37]}$ & $0.637_{[0.62,0.65]}$ & $0.272_{[0.26,0.29]}$ & $0.545_{[0.53,0.56]}$ \\
 & 16M & $0.171_{[0.12,0.22]}$ & $0.094_{[0.06,0.13]}$ & $0.113_{[0.09,0.14]}$ & $0.457_{[0.43,0.49]}$ & $0.675_{[0.66,0.69]}$ & $0.302_{[0.29,0.32]}$ & $0.600_{[0.58,0.61]}$ \\
 & 32M & $0.215_{[0.16,0.28]}$ & $0.143_{[0.10,0.18]}$ & $0.114_{[0.09,0.13]}$ & $0.459_{[0.43,0.49]}$ & $0.692_{[0.67,0.71]}$ & $0.325_{[0.31,0.34]}$ & $0.614_{[0.60,0.63]}$ \\
\cmidrule(lr){1-9}
\multirow{5}{*}{ROUGE-Lsum} & 2M & $0.171_{[0.13,0.21]}$ & $0.083_{[0.02,0.15]}$ & $0.141_{[0.11,0.18]}$ & $0.430_{[0.40,0.46]}$ & $0.591_{[0.57,0.61]}$ & $0.283_{[0.26,0.30]}$ & $0.533_{[0.52,0.55]}$ \\
 & 4M & $0.233_{[0.16,0.31]}$ & $0.178_{[0.08,0.27]}$ & $0.122_{[0.10,0.15]}$ & $0.364_{[0.34,0.39]}$ & $0.597_{[0.58,0.62]}$ & $0.299_{[0.27,0.32]}$ & $0.525_{[0.51,0.54]}$ \\
 & 8M & $0.254_{[0.17,0.34]}$ & $0.204_{[0.11,0.29]}$ & $0.134_{[0.10,0.17]}$ & $0.368_{[0.33,0.40]}$ & $0.644_{[0.63,0.66]}$ & $0.321_{[0.29,0.35]}$ & $0.562_{[0.55,0.58]}$ \\
 & 16M & $0.267_{[0.19,0.35]}$ & $0.123_{[0.07,0.19]}$ & $0.147_{[0.12,0.18]}$ & $0.497_{[0.46,0.53]}$ & $0.680_{[0.66,0.70]}$ & $0.343_{[0.32,0.37]}$ & $0.615_{[0.60,0.63]}$ \\
 & 32M & $0.328_{[0.24,0.42]}$ & $0.219_{[0.13,0.31]}$ & $0.177_{[0.14,0.21]}$ & $0.495_{[0.46,0.53]}$ & $0.698_{[0.68,0.72]}$ & $0.383_{[0.36,0.41]}$ & $0.630_{[0.62,0.64]}$ \\
\cmidrule(lr){1-9}
\multirow{5}{*}{BERTScore} & 2M & $0.817_{[0.81,0.82]}$ & $0.769_{[0.75,0.79]}$ & $0.823_{[0.81,0.83]}$ & $0.866_{[0.86,0.87]}$ & $0.916_{[0.91,0.92]}$ & $0.838_{[0.83,0.84]}$ & $0.900_{[0.90,0.90]}$ \\
 & 4M & $0.832_{[0.82,0.85]}$ & $0.800_{[0.78,0.82]}$ & $0.788_{[0.78,0.80]}$ & $0.853_{[0.85,0.86]}$ & $0.918_{[0.91,0.92]}$ & $0.838_{[0.83,0.84]}$ & $0.898_{[0.89,0.90]}$ \\
 & 8M & $0.839_{[0.82,0.85]}$ & $0.817_{[0.79,0.84]}$ & $0.790_{[0.78,0.80]}$ & $0.850_{[0.84,0.86]}$ & $0.928_{[0.92,0.93]}$ & $0.845_{[0.84,0.85]}$ & $0.905_{[0.90,0.91]}$ \\
 & 16M & $0.851_{[0.84,0.86]}$ & $0.805_{[0.79,0.82]}$ & $0.834_{[0.82,0.85]}$ & $0.878_{[0.87,0.89]}$ & $0.935_{[0.93,0.94]}$ & $0.861_{[0.86,0.87]}$ & $0.918_{[0.92,0.92]}$ \\
 & 32M & $0.862_{[0.85,0.87]}$ & $0.813_{[0.79,0.84]}$ & $0.832_{[0.82,0.84]}$ & $0.879_{[0.87,0.89]}$ & $0.939_{[0.94,0.94]}$ & $0.865_{[0.86,0.87]}$ & $0.921_{[0.92,0.92]}$ \\
\bottomrule
\end{tabular}
\caption{Full per-domain results (\textbf{unpruned}, seed~42) across the complete metric family (ROUGE-1/2/L/Lsum and BERTScore-F1), with bootstrap 95\% CIs in subscripts. Rows are grouped by allocation scheme (Balanced vs.\ Natural) and, within each scheme, by metric; each metric block reports the five-budget ladder (2--32M). Column italics are test-set meeting counts. This is the unpruned companion to the pruned grid; comparing the two at matched (scheme, budget, metric) cells isolates the effect of transcript pruning (RQ3).}
\label{tab:app-full-unpruned}
\end{table*}

%%%%%%%%%%%%%%%%%%%%% UNPRUNED END %%%%%%%%%%%%%%%%%%%%%%%

\begin{table*}[t]\centering\small\setlength{\tabcolsep}{4pt}
\begin{tabular}{llccccccc}
\toprule
\textbf{Metric} & \textbf{Budget} & \textbf{AMI} & \textbf{ICSI} & \textbf{ELITR} & \textbf{EPM} & \textbf{MB} & \textbf{Macro} & \textbf{Micro} \\
 & & \textit{20} & \textit{6} & \textit{38} & \textit{242} & \textit{862} & & \\
\midrule
\multicolumn{9}{l}{\textit{Balanced (equal-token) allocation}} \\
\midrule
\multirow{3}{*}{ROUGE-1}
 & 2M  & 0.105 & 0.090 & 0.096 & 0.219 & 0.530 & 0.208 & 0.442 \\
 & 8M  & 0.324 & 0.185 & 0.128 & 0.256 & 0.587 & 0.296 & 0.497 \\
 & 32M & 0.497 & 0.424 & 0.238 & 0.224 & 0.613 & 0.399 & 0.518 \\
\cmidrule(l){1-9}
\multirow{3}{*}{ROUGE-2}
 & 2M  & 0.033 & 0.015 & 0.018 & 0.123 & 0.416 & 0.121 & 0.334 \\
 & 8M  & 0.137 & 0.040 & 0.028 & 0.170 & 0.474 & 0.170 & 0.388 \\
 & 32M & 0.174 & 0.102 & 0.060 & 0.155 & 0.505 & 0.199 & 0.410 \\
\cmidrule(l){1-9}
\multirow{3}{*}{ROUGE-L}
 & 2M  & 0.077 & 0.077 & 0.072 & 0.172 & 0.496 & 0.178 & 0.405 \\
 & 8M  & 0.198 & 0.111 & 0.083 & 0.213 & 0.550 & 0.231 & 0.457 \\
 & 32M & 0.265 & 0.182 & 0.127 & 0.195 & 0.578 & 0.269 & 0.477 \\
\cmidrule(l){1-9}
\multirow{3}{*}{ROUGE-Lsum}
 & 2M  & 0.098 & 0.080 & 0.091 & 0.213 & 0.502 & 0.197 & 0.420 \\
 & 8M  & 0.312 & 0.174 & 0.122 & 0.252 & 0.560 & 0.284 & 0.476 \\
 & 32M & 0.476 & 0.402 & 0.229 & 0.220 & 0.587 & 0.383 & 0.496 \\
\cmidrule(l){1-9}
\multirow{3}{*}{BERTScore-F1}
 & 2M  & 0.821 & 0.788 & 0.786 & 0.813 & 0.897 & 0.821 & 0.874 \\
 & 8M  & 0.861 & 0.823 & 0.817 & 0.838 & 0.910 & 0.850 & 0.891 \\
 & 32M & 0.872 & 0.844 & 0.836 & 0.842 & 0.915 & 0.862 & 0.896 \\
\midrule
\multicolumn{9}{l}{\textit{Natural (proportional) allocation}} \\
\midrule
\multirow{3}{*}{ROUGE-1}
 & 2M  & 0.161 & 0.031 & 0.042 & 0.268 & 0.563 & 0.213 & 0.475 \\
 & 8M  & 0.121 & 0.086 & 0.087 & 0.432 & 0.642 & 0.274 & 0.569 \\
 & 32M & 0.163 & 0.131 & 0.111 & 0.538 & 0.699 & 0.328 & 0.634 \\
\cmidrule(l){1-9}
\multirow{3}{*}{ROUGE-2}
 & 2M  & 0.039 & 0.004 & 0.007 & 0.169 & 0.458 & 0.135 & 0.374 \\
 & 8M  & 0.035 & 0.014 & 0.016 & 0.296 & 0.536 & 0.179 & 0.458 \\
 & 32M & 0.064 & 0.026 & 0.019 & 0.392 & 0.599 & 0.220 & 0.525 \\
\cmidrule(l){1-9}
\multirow{3}{*}{ROUGE-L}
 & 2M  & 0.101 & 0.027 & 0.030 & 0.218 & 0.533 & 0.182 & 0.442 \\
 & 8M  & 0.087 & 0.076 & 0.067 & 0.374 & 0.609 & 0.243 & 0.531 \\
 & 32M & 0.110 & 0.101 & 0.071 & 0.474 & 0.672 & 0.286 & 0.599 \\
\cmidrule(l){1-9}
\multirow{3}{*}{ROUGE-Lsum}
 & 2M  & 0.152 & 0.029 & 0.040 & 0.263 & 0.539 & 0.204 & 0.456 \\
 & 8M  & 0.113 & 0.079 & 0.082 & 0.426 & 0.615 & 0.263 & 0.547 \\
 & 32M & 0.156 & 0.124 & 0.105 & 0.531 & 0.679 & 0.319 & 0.617 \\
\cmidrule(l){1-9}
\multirow{3}{*}{BERTScore-F1}
 & 2M  & 0.800 & 0.750 & 0.759 & 0.834 & 0.906 & 0.810 & 0.884 \\
 & 8M  & 0.821 & 0.784 & 0.779 & 0.867 & 0.922 & 0.835 & 0.904 \\
 & 32M & 0.831 & 0.800 & 0.805 & 0.885 & 0.935 & 0.851 & 0.918 \\
\bottomrule
\end{tabular}
\caption{Llama-3.2-3B (RQ5): full per-domain results across the complete metric family, for both allocation schemes at the three shared budgets (2M/8M/32M; pruned, seed 42). Column italics are test-set meeting counts. The per-domain crossover of RQ1 reproduces here: balanced leads the minority domains (AMI/ICSI/ELITR) while natural leads the majority domains (EPM/MB).}
\label{tab:llama-full}
\end{table*}

\begin{table*}[t]

\centering\tiny\setlength{\tabcolsep}{4pt}
\begin{tabular}{llccccc|cc}
\toprule
\textbf{System} & \textbf{Dim.} & \textbf{AMI} & \textbf{ICSI} & \textbf{ELITR} & \textbf{EPM} & \textbf{MB} & \textbf{Macro} & \textbf{Micro} \\
\midrule
 \multirow{3}{*}{balanced 2M} & Comple. & $0.370_{[0.29,0.46]}$ & $0.147_{[0.07,0.22]}$ & $0.077_{[0.05,0.11]}$ & $0.310_{[0.28,0.34]}$ & $0.478_{[0.45,0.50]}$ & $0.276_{[0.25,0.30]}$ & $0.427_{[0.41,0.45]}$ \\
  & Faithf. & $0.884_{[0.83,0.93]}$ & $0.906_{[0.87,0.94]}$ & $0.784_{[0.68,0.88]}$ & $0.502_{[0.46,0.55]}$ & $0.749_{[0.73,0.77]}$ & $0.765_{[0.74,0.79]}$ & $0.707_{[0.69,0.73]}$ \\
  & Concis. & $0.541_{[0.48,0.61]}$ & $0.549_{[0.44,0.65]}$ & $0.443_{[0.36,0.52]}$ & $0.427_{[0.39,0.46]}$ & $0.615_{[0.59,0.64]}$ & $0.515_{[0.48,0.54]}$ & $0.573_{[0.55,0.59]}$ \\
\midrule
 \multirow{3}{*}{balanced 32M} & Comple. & $0.496_{[0.42,0.57]}$ & $0.209_{[0.15,0.28]}$ & $0.201_{[0.15,0.25]}$ & $0.411_{[0.38,0.44]}$ & $0.550_{[0.53,0.57]}$ & $0.373_{[0.35,0.40]}$ & $0.507_{[0.49,0.53]}$ \\
  & Faithf. & $0.892_{[0.86,0.92]}$ & $0.916_{[0.87,0.96]}$ & $0.791_{[0.74,0.83]}$ & $0.418_{[0.38,0.45]}$ & $0.691_{[0.67,0.71]}$ & $0.742_{[0.73,0.76]}$ & $0.642_{[0.62,0.66]}$ \\
  & Concis. & $0.481_{[0.42,0.54]}$ & $0.457_{[0.39,0.51]}$ & $0.313_{[0.26,0.36]}$ & $0.469_{[0.44,0.50]}$ & $0.639_{[0.62,0.66]}$ & $0.472_{[0.45,0.49]}$ & $0.589_{[0.57,0.61]}$ \\
\midrule
 \multirow{3}{*}{natural 2M} & Comple. & $0.308_{[0.26,0.36]}$ & $0.107_{[0.05,0.17]}$ & $0.077_{[0.04,0.11]}$ & $0.298_{[0.27,0.33]}$ & $0.505_{[0.48,0.53]}$ & $0.259_{[0.24,0.28]}$ & $0.443_{[0.42,0.46]}$ \\
  & Faithf. & $0.886_{[0.84,0.93]}$ & $0.854_{[0.69,0.99]}$ & $0.774_{[0.67,0.86]}$ & $0.438_{[0.39,0.49]}$ & $0.744_{[0.72,0.76]}$ & $0.739_{[0.70,0.78]}$ & $0.685_{[0.67,0.70]}$ \\
  & Concis. & $0.498_{[0.43,0.57]}$ & $0.361_{[0.22,0.53]}$ & $0.478_{[0.38,0.58]}$ & $0.351_{[0.31,0.39]}$ & $0.645_{[0.62,0.67]}$ & $0.467_{[0.43,0.51]}$ & $0.575_{[0.55,0.60]}$ \\
\midrule
 \multirow{3}{*}{natural 32M} & Comple. & $0.428_{[0.34,0.52]}$ & $0.136_{[0.07,0.22]}$ & $0.125_{[0.09,0.17]}$ & $0.392_{[0.36,0.42]}$ & $0.580_{[0.56,0.60]}$ & $0.332_{[0.31,0.36]}$ & $0.522_{[0.50,0.54]}$ \\
  & Faithf. & $0.875_{[0.85,0.91]}$ & $0.916_{[0.82,0.99]}$ & $0.786_{[0.70,0.86]}$ & $0.359_{[0.32,0.39]}$ & $0.701_{[0.68,0.72]}$ & $0.728_{[0.70,0.75]}$ & $0.636_{[0.62,0.66]}$ \\
  & Concis. & $0.559_{[0.50,0.62]}$ & $0.499_{[0.26,0.72]}$ & $0.347_{[0.28,0.42]}$ & $0.434_{[0.40,0.47]}$ & $0.694_{[0.67,0.72]}$ & $0.506_{[0.46,0.56]}$ & $0.625_{[0.61,0.64]}$ \\
\bottomrule
\end{tabular}
\caption{Per-domain fact-level LLM-judge scores (Qwen2.5-72B-Instruct-AWQ) for the
balanced and natural schemes at the 2M and 32M budgets, with bootstrap 95\% CIs in
subscripts. Comple. (completeness) is the fraction of reference-minute facts
recovered by the generated minute; Faithf. (faithfulness) is the fraction of
generated-minute facts supported by the transcript; Concis. (conciseness) is
the fraction of generated-minute facts that are salient with respect to the reference.
Macro weights the five domains equally, micro weights meetings. This is the
per-domain breakdown of the aggregate scores in Table~\ref{tab:judge}; the RQ1
crossover on completeness (balanced leads the minority domains, natural the majority)
is visible per domain, while faithfulness falls from 2M to 32M under both schemes,
most sharply on EPM.}
\label{tab:judge-full}
\end{table*}

\begin{table*}[t]
\centering
\small
\setlength{\tabcolsep}{5pt}
\begin{tabular}{ll ccc ccc}
\toprule
& & \multicolumn{3}{c}{\textbf{Mistral-7B}} & \multicolumn{3}{c}{\textbf{Llama-3.2-3B}} \\
\cmidrule(lr){3-5}\cmidrule(lr){6-8}
\textbf{Dataset} & \textbf{Budget} & Bal. & Nat. & $\Delta$ & Bal. & Nat. & $\Delta$ \\
\midrule
\multicolumn{8}{l}{\textit{ROUGE-Lsum}} \\
\multirow{3}{*}{AMI}
 & 2M  & 0.376 & 0.331 & $+0.045$ & 0.098 & 0.152 & $-0.054$ \\
 & 8M  & 0.449 & 0.230 & $+0.219$ & 0.312 & 0.113 & $+0.199$ \\
 & 32M & 0.495 & 0.393 & $+0.102$ & 0.476 & 0.156 & $+0.320$ \\
\cmidrule(lr){1-8}
\multirow{3}{*}{ICSI}
 & 2M  & 0.258 & 0.129 & $+0.129$ & 0.080 & 0.029 & $+0.051$ \\
 & 8M  & 0.419 & 0.107 & $+0.312$ & 0.174 & 0.079 & $+0.095$ \\
 & 32M & 0.445 & 0.281 & $+0.164$ & 0.402 & 0.124 & $+0.278$ \\
\cmidrule(lr){1-8}
\multirow{3}{*}{ELITR}
 & 2M  & 0.152 & 0.184 & $-0.032$ & 0.091 & 0.040 & $+0.051$ \\
 & 8M  & 0.244 & 0.107 & $+0.137$ & 0.122 & 0.082 & $+0.040$ \\
 & 32M & 0.367 & 0.204 & $+0.163$ & 0.229 & 0.105 & $+0.124$ \\
\cmidrule(lr){1-8}
\multirow{3}{*}{EPM}
 & 2M  & 0.222 & 0.274 & $-0.052$ & 0.213 & 0.263 & $-0.050$ \\
 & 8M  & 0.438 & 0.516 & $-0.078$ & 0.252 & 0.426 & $-0.174$ \\
 & 32M & 0.547 & 0.534 & $+0.013$ & 0.220 & 0.531 & $-0.311$ \\
\cmidrule(lr){1-8}
\multirow{3}{*}{MB}
 & 2M  & 0.562 & 0.586 & $-0.024$ & 0.502 & 0.539 & $-0.037$ \\
 & 8M  & 0.598 & 0.646 & $-0.048$ & 0.560 & 0.615 & $-0.055$ \\
 & 32M & 0.648 & 0.694 & $-0.046$ & 0.587 & 0.679 & $-0.092$ \\
\midrule
\multicolumn{8}{l}{\textit{BERTScore-F1}} \\
\multirow{3}{*}{AMI}
 & 2M  & 0.861 & 0.851 & $+0.010$ & 0.821 & 0.800 & $+0.021$ \\
 & 8M  & 0.870 & 0.835 & $+0.035$ & 0.861 & 0.821 & $+0.040$ \\
 & 32M & 0.875 & 0.865 & $+0.010$ & 0.872 & 0.831 & $+0.041$ \\
\cmidrule(lr){1-8}
\multirow{3}{*}{ICSI}
 & 2M  & 0.827 & 0.814 & $+0.013$ & 0.788 & 0.750 & $+0.038$ \\
 & 8M  & 0.851 & 0.801 & $+0.050$ & 0.823 & 0.784 & $+0.039$ \\
 & 32M & 0.852 & 0.833 & $+0.019$ & 0.844 & 0.800 & $+0.044$ \\
\cmidrule(lr){1-8}
\multirow{3}{*}{ELITR}
 & 2M  & 0.806 & 0.806 & $+0.000$ & 0.786 & 0.759 & $+0.027$ \\
 & 8M  & 0.835 & 0.815 & $+0.020$ & 0.817 & 0.779 & $+0.038$ \\
 & 32M & 0.846 & 0.837 & $+0.009$ & 0.836 & 0.805 & $+0.031$ \\
\cmidrule(lr){1-8}
\multirow{3}{*}{EPM}
 & 2M  & 0.832 & 0.831 & $+0.001$ & 0.813 & 0.834 & $-0.021$ \\
 & 8M  & 0.865 & 0.882 & $-0.017$ & 0.838 & 0.867 & $-0.029$ \\
 & 32M & 0.887 & 0.889 & $-0.002$ & 0.842 & 0.885 & $-0.043$ \\
\cmidrule(lr){1-8}
\multirow{3}{*}{MB}
 & 2M  & 0.910 & 0.915 & $-0.005$ & 0.897 & 0.906 & $-0.009$ \\
 & 8M  & 0.918 & 0.925 & $-0.007$ & 0.910 & 0.922 & $-0.012$ \\
 & 32M & 0.928 & 0.938 & $-0.010$ & 0.915 & 0.935 & $-0.020$ \\
\bottomrule
\end{tabular}
\caption{RQ5 full numeric detail backing Table~\ref{tab:results-rq5}. For each dataset,
budget, and model, Bal. and Nat. are the raw balanced and natural scores
and $\Delta=$ balanced $-$ natural (from the displayed rounded values); all runs are
pruned, single seed~42. The sign of $\Delta$ agrees across the two models for every
dataset on both metrics---positive on AMI/ICSI/ELITR, negative on EPM/MB---the only
per-cell exceptions being at 2M (AMI and ELITR on ROUGE-Lsum, EPM on BERTScore-F1).}
\label{tab:rq5-numeric}
\end{table*}

% Multi-seed robustness, all metrics: mean $\pm$ std over seeds {42, 2, 15}
\begin{table*}[t]
\centering
\small
\setlength{\tabcolsep}{4pt}
\resizebox{\textwidth}{!}{%
\begin{tabular}{lllccccccc}
\toprule
\textbf{Metric} & \textbf{Scheme} & \textbf{Budget} & \textbf{AMI} & \textbf{ICSI} & \textbf{ELITR} & \textbf{EPM} & \textbf{MB} & \textbf{Macro} & \textbf{Micro} \\
 & & & \textit{20} & \textit{6} & \textit{38} & \textit{242} & \textit{862} & & \\
\midrule
\multirow{6}{*}{ROUGE-1} & Balanced & 2M & $0.380_{\pm0.057}$ & $0.271_{\pm0.022}$ & $0.161_{\pm0.005}$ & $0.253_{\pm0.048}$ & $0.576_{\pm0.029}$ & $0.328_{\pm0.003}$ & $0.491_{\pm0.028}$ \\
 & Balanced & 8M & $0.455_{\pm0.022}$ & $0.391_{\pm0.046}$ & $0.235_{\pm0.024}$ & $0.365_{\pm0.087}$ & $0.630_{\pm0.005}$ & $0.415_{\pm0.031}$ & $0.558_{\pm0.016}$ \\
 & Balanced & 32M & $0.527_{\pm0.011}$ & $0.452_{\pm0.016}$ & $0.370_{\pm0.008}$ & $0.524_{\pm0.027}$ & $0.675_{\pm0.007}$ & $0.510_{\pm0.006}$ & $0.630_{\pm0.006}$ \\
 & Natural & 2M & $0.316_{\pm0.031}$ & $0.176_{\pm0.079}$ & $0.176_{\pm0.012}$ & $0.355_{\pm0.065}$ & $0.611_{\pm0.017}$ & $0.327_{\pm0.026}$ & $0.536_{\pm0.018}$ \\
 & Natural & 8M & $0.241_{\pm0.004}$ & $0.131_{\pm0.042}$ & $0.105_{\pm0.010}$ & $0.481_{\pm0.046}$ & $0.669_{\pm0.007}$ & $0.325_{\pm0.007}$ & $0.602_{\pm0.013}$ \\
 & Natural & 32M & $0.351_{\pm0.061}$ & $0.217_{\pm0.072}$ & $0.178_{\pm0.030}$ & $0.548_{\pm0.025}$ & $0.715_{\pm0.002}$ & $0.402_{\pm0.028}$ & $0.654_{\pm0.005}$ \\
\cmidrule(lr){1-10}
\multirow{6}{*}{ROUGE-2} & Balanced & 2M & $0.155_{\pm0.026}$ & $0.059_{\pm0.007}$ & $0.037_{\pm0.003}$ & $0.163_{\pm0.035}$ & $0.473_{\pm0.033}$ & $0.177_{\pm0.006}$ & $0.387_{\pm0.030}$ \\
 & Balanced & 8M & $0.179_{\pm0.003}$ & $0.098_{\pm0.020}$ & $0.061_{\pm0.006}$ & $0.251_{\pm0.070}$ & $0.522_{\pm0.005}$ & $0.222_{\pm0.018}$ & $0.443_{\pm0.011}$ \\
 & Balanced & 32M & $0.201_{\pm0.007}$ & $0.110_{\pm0.005}$ & $0.092_{\pm0.005}$ & $0.386_{\pm0.020}$ & $0.577_{\pm0.007}$ & $0.273_{\pm0.004}$ & $0.513_{\pm0.006}$ \\
 & Natural & 2M & $0.108_{\pm0.007}$ & $0.034_{\pm0.015}$ & $0.037_{\pm0.003}$ & $0.234_{\pm0.048}$ & $0.499_{\pm0.018}$ & $0.183_{\pm0.013}$ & $0.420_{\pm0.015}$ \\
 & Natural & 8M & $0.087_{\pm0.006}$ & $0.020_{\pm0.005}$ & $0.021_{\pm0.002}$ & $0.348_{\pm0.032}$ & $0.569_{\pm0.004}$ & $0.209_{\pm0.007}$ & $0.495_{\pm0.009}$ \\
 & Natural & 32M & $0.139_{\pm0.023}$ & $0.044_{\pm0.018}$ & $0.043_{\pm0.013}$ & $0.416_{\pm0.021}$ & $0.624_{\pm0.003}$ & $0.253_{\pm0.007}$ & $0.551_{\pm0.005}$ \\
\cmidrule(lr){1-10}
\multirow{6}{*}{ROUGE-L} & Balanced & 2M & $0.233_{\pm0.036}$ & $0.158_{\pm0.018}$ & $0.103_{\pm0.005}$ & $0.214_{\pm0.042}$ & $0.547_{\pm0.032}$ & $0.251_{\pm0.003}$ & $0.456_{\pm0.030}$ \\
 & Balanced & 8M & $0.260_{\pm0.007}$ & $0.189_{\pm0.008}$ & $0.132_{\pm0.014}$ & $0.318_{\pm0.084}$ & $0.596_{\pm0.005}$ & $0.299_{\pm0.020}$ & $0.515_{\pm0.014}$ \\
 & Balanced & 32M & $0.284_{\pm0.009}$ & $0.191_{\pm0.006}$ & $0.174_{\pm0.010}$ & $0.463_{\pm0.022}$ & $0.647_{\pm0.007}$ & $0.352_{\pm0.003}$ & $0.585_{\pm0.006}$ \\
 & Natural & 2M & $0.190_{\pm0.016}$ & $0.112_{\pm0.025}$ & $0.108_{\pm0.007}$ & $0.297_{\pm0.056}$ & $0.573_{\pm0.018}$ & $0.256_{\pm0.016}$ & $0.492_{\pm0.017}$ \\
 & Natural & 8M & $0.156_{\pm0.006}$ & $0.091_{\pm0.021}$ & $0.072_{\pm0.003}$ & $0.426_{\pm0.037}$ & $0.639_{\pm0.006}$ & $0.277_{\pm0.006}$ & $0.566_{\pm0.011}$ \\
 & Natural & 32M & $0.214_{\pm0.035}$ & $0.125_{\pm0.031}$ & $0.114_{\pm0.018}$ & $0.495_{\pm0.021}$ & $0.690_{\pm0.002}$ & $0.328_{\pm0.013}$ & $0.620_{\pm0.005}$ \\
\cmidrule(lr){1-10}
\multirow{6}{*}{ROUGE-Lsum} & Balanced & 2M & $0.368_{\pm0.054}$ & $0.256_{\pm0.020}$ & $0.154_{\pm0.006}$ & $0.248_{\pm0.048}$ & $0.553_{\pm0.033}$ & $0.316_{\pm0.002}$ & $0.472_{\pm0.031}$ \\
 & Balanced & 8M & $0.440_{\pm0.022}$ & $0.371_{\pm0.042}$ & $0.228_{\pm0.024}$ & $0.360_{\pm0.087}$ & $0.604_{\pm0.005}$ & $0.400_{\pm0.030}$ & $0.537_{\pm0.016}$ \\
 & Balanced & 32M & $0.506_{\pm0.010}$ & $0.430_{\pm0.018}$ & $0.359_{\pm0.008}$ & $0.518_{\pm0.027}$ & $0.654_{\pm0.007}$ & $0.493_{\pm0.006}$ & $0.612_{\pm0.006}$ \\
 & Natural & 2M & $0.302_{\pm0.032}$ & $0.165_{\pm0.076}$ & $0.169_{\pm0.013}$ & $0.349_{\pm0.065}$ & $0.582_{\pm0.018}$ & $0.314_{\pm0.025}$ & $0.513_{\pm0.018}$ \\
 & Natural & 8M & $0.230_{\pm0.005}$ & $0.120_{\pm0.037}$ & $0.097_{\pm0.008}$ & $0.474_{\pm0.045}$ & $0.647_{\pm0.006}$ & $0.314_{\pm0.006}$ & $0.583_{\pm0.012}$ \\
 & Natural & 32M & $0.338_{\pm0.058}$ & $0.203_{\pm0.069}$ & $0.170_{\pm0.030}$ & $0.542_{\pm0.025}$ & $0.696_{\pm0.002}$ & $0.390_{\pm0.027}$ & $0.638_{\pm0.005}$ \\
\cmidrule(lr){1-10}
\multirow{6}{*}{BERTScore-F1} & Balanced & 2M & $0.862_{\pm0.005}$ & $0.825_{\pm0.003}$ & $0.801_{\pm0.007}$ & $0.831_{\pm0.009}$ & $0.909_{\pm0.007}$ & $0.846_{\pm0.002}$ & $0.888_{\pm0.007}$ \\
 & Balanced & 8M & $0.870_{\pm0.002}$ & $0.847_{\pm0.004}$ & $0.826_{\pm0.012}$ & $0.853_{\pm0.014}$ & $0.919_{\pm0.001}$ & $0.863_{\pm0.006}$ & $0.901_{\pm0.003}$ \\
 & Balanced & 32M & $0.877_{\pm0.002}$ & $0.850_{\pm0.002}$ & $0.842_{\pm0.003}$ & $0.882_{\pm0.005}$ & $0.930_{\pm0.002}$ & $0.876_{\pm0.002}$ & $0.916_{\pm0.001}$ \\
 & Natural & 2M & $0.849_{\pm0.006}$ & $0.815_{\pm0.007}$ & $0.803_{\pm0.007}$ & $0.847_{\pm0.014}$ & $0.914_{\pm0.004}$ & $0.846_{\pm0.003}$ & $0.895_{\pm0.004}$ \\
 & Natural & 8M & $0.838_{\pm0.006}$ & $0.799_{\pm0.003}$ & $0.810_{\pm0.008}$ & $0.874_{\pm0.008}$ & $0.926_{\pm0.002}$ & $0.850_{\pm0.003}$ & $0.909_{\pm0.002}$ \\
 & Natural & 32M & $0.854_{\pm0.009}$ & $0.813_{\pm0.017}$ & $0.821_{\pm0.014}$ & $0.892_{\pm0.005}$ & $0.938_{\pm0.001}$ & $0.864_{\pm0.008}$ & $0.922_{\pm0.001}$ \\
\bottomrule
\end{tabular}%
}
\caption{Multi-seed robustness (RQ1): per-domain scores as mean\,$\pm$\,std over
seeds $\{2, 15, 42\}$ for the balanced and natural schemes, pruned condition, at
the 2M, 8M, and 32M budgets. Column italics are test-set meeting counts. The
per-domain crossover of Section~\ref{sec:rq1} reproduces across seeds: at 8M and
32M, balanced leads all three minority domains (AMI, ICSI, ELITR) and natural
leads the majority domains (EPM, MB) on every metric, so macro favours balanced
while micro favours natural. At 2M the minority advantage is smaller and ELITR
falls within seed variation.}
\label{tab:multiseed-all}
\end{table*}

\clearpage
\onecolumn
\section{Prompts}
\label{app:prompts}
\subsection{Pruning Prompt}
\label{app:prune-prompt}

The verbatim prompt supplied to the pruning model is given below, comprising a system prompt and a user-prompt template. It is applied identically to all five corpora. Each transcript is processed in chunks; \texttt{transcript\_chunk} in the user template is replaced with the numbered lines of the current chunk, and the global line numbering is preserved across chunks.

\begin{promptbox}{Pruning Prompt}
\textbf{\sffamily [System]}
\begin{Verbatim}[fontsize=\small]
You clean meeting transcripts by removing ONLY conversational noise. This is NOT summarization.
You KEEP the large majority of lines. You are removing filler, not compressing the meeting.

Each line starts with a number in square brackets: "[n] SPEAKER: text".

DELETE a line ONLY if it is pure conversational noise with NO standalone content:
- bare backchannels / fillers: "okay", "right", "yeah", "mm-hmm", "uh", "great", "cool"
- greetings, goodbyes, thanks, social pleasantries
- connection/logistics chatter: "can you hear me?", "you're muted"
- pure acknowledgements that do not answer a question or state anything
- meaningless fragments / false starts that carry no information ("Thin", "other")

KEEP a line if it contains ANY of the following, even briefly:
- any topic, fact, opinion, reason, problem, suggestion, or description
- decisions, agreements, action items, deadlines, numbers, prices, names, dates
- questions, and answers to questions (even short ones like "Friday", "Yes", "25 Euro")
- introductions, role assignments, anything about what the meeting is about

Rules:
- KEEP is the default. Only delete a line you are confident is pure noise.
- Most transcripts have only a MINORITY of lines deleted (roughly 15-35%), never the majority.
- If unsure, KEEP it (leave it out of the delete list).
- Do NOT rewrite, merge, renumber, or alter any line. Only decide deletions.
- Use the line number exactly as it appears in the brackets.

Output ONLY a JSON object with one key "delete": a list of the line numbers (integers) to
delete. Nothing else. Example: {"delete": [1, 2, 4, 7]}
\end{Verbatim}

\vspace{4pt}
\textbf{\sffamily [User]}
\begin{Verbatim}[fontsize=\small]
Below is a meeting transcript with numbered lines.

Remove ONLY non-essential conversational noise (bare "okay"/"yeah"/"right" acknowledgements,
greetings, goodbyes, thanks, "can you hear me?", meaningless fragments). KEEP everything that
carries any content: decisions, numbers, questions, answers, opinions, reasons, descriptions.

This is NOT summarization. Keep the majority of lines. When unsure, keep the line.

Transcript:
<<<
{transcript_chunk}
>>>

Return ONLY the JSON object with the "delete" list.
\end{Verbatim}
\end{promptbox}

\subsection{Summarisation Prompt}
\label{app:summ-prompt}

The instruction used during zero-shot settings and fine-tuning. It is applied identically to all candidate models; \texttt{\{transcript\}} is replaced with the raw (unpruned) test transcript.

\begin{promptbox}{Summarisation Prompt}
\textbf{\sffamily [System]}
\begin{Verbatim}[fontsize=\small]
You are a meeting summarizer. Given a meeting transcript, write the meeting minutes
that faithfully capture the substantive content of the meeting. Base the minutes only
on what is stated in the transcript; do not introduce information that is not present.
\end{Verbatim}

\vspace{4pt}
\textbf{\sffamily [User]}
\begin{Verbatim}[fontsize=\small]
Here is the meeting transcript. Write the meeting minutes.

{transcript}
\end{Verbatim}
\end{promptbox}

\subsection{LLM-as-a-Judge Evaluation Prompts}
\label{app:judge-prompts}

The prompts below are used in our fact-level evaluation (Section~\ref{sec:evaluation}), applied under greedy decoding with a judge model isolated from the fine-tuned systems. Each enforces a strict JSON schema and a fixed set of adjudication rules applied identically across all mixtures, budgets, and domains; the same rules serve as annotation guidelines for the human-validation slice. Atomic facts are produced by the decomposition prompt, applied once to the reference minute and once to the generated minute.

\begin{promptbox}{Atomic Fact Decomposition}
\textbf{\sffamily [System]}
\begin{Verbatim}[fontsize=\small]
You are an expert at decomposing meeting minutes into atomic facts. This is NOT
summarization: you re-express existing content as a complete list of atomic facts,
adding and omitting nothing.

Follow four criteria for every atomic fact:
- Losslessness: together the facts capture ALL information in the summary.
- Atomicity: each fact is the smallest indivisible unit -- exactly one claim.
  Split every compound statement into separate facts.
- Independence: each fact is self-contained and understandable in isolation.
  Resolve pronouns and references (write "Mark", not "he").
- Declarativeness: each fact is one concise, objective declarative sentence.

Decompose ONLY what the summary states; do not infer beyond it. Do not treat
section headers or formatting as facts unless they carry standalone content.

Output ONLY a JSON object with one key "atomic_facts": a list of strings.
Example: {"atomic_facts": ["fact_1", "fact_2", "...", "fact_n"]}
\end{Verbatim}

\vspace{4pt}
\textbf{\sffamily [User]}
\begin{Verbatim}[fontsize=\small]
Decompose the following meeting minutes into atomic facts.

[Summary Start]
{summary}
[Summary End]

Return ONLY the JSON object.
\end{Verbatim}
\end{promptbox}

\begin{promptbox}{Completeness: Key-Fact Matching}
\textbf{\sffamily [System]}
\begin{Verbatim}[fontsize=\small]
You are an expert evaluator. Given KEY FACTS from a reference set of meeting minutes
and a candidate SUMMARY, decide for each key fact whether the candidate captures it.

Labels:
- captured: the candidate explicitly states the key fact, or states something from
  which it follows directly and unambiguously. Paraphrase and different wording
  count, provided meaning is preserved.
- not_captured: the key fact is absent, contradicted, or only vaguely alluded to
  without its substantive content.

Adjudication rules (apply strictly and identically):
- Meaning over wording: a faithful paraphrase captures the fact.
- Specificity, downward: if the candidate is LESS specific than the key fact, it is
  captured ONLY IF it still conveys the fact's essential content. Omitting the core
  value (number, name, decision) is not_captured, even if the topic is mentioned.
- Specificity, upward: extra detail in the candidate never harms capture.
- Numbers/dates/names: captured only if the specific value appears or is
  unambiguously entailed. A different value is not_captured.
- Base your judgment SOLELY on the candidate summary. Do NOT use the transcript or
  outside knowledge to fill gaps the candidate leaves.

For a captured fact, give the candidate sentence index that best evidences it;
for not_captured, set it to null.

Output ONLY a JSON list of objects with keys: fact_idx (int),
conclusion ("captured" or "not_captured"), evidence_sentence_idx (int or null),
reason (one sentence).
\end{Verbatim}

\vspace{4pt}
\textbf{\sffamily [User]}
\begin{Verbatim}[fontsize=\small]
Summary:
{indexed_generated_summary_sentences}

Key Facts:
{indexed_reference_atomic_facts}

Return ONLY the JSON list.
\end{Verbatim}
\end{promptbox}

\begin{promptbox}{Faithfulness: Fact Verification}
\textbf{\sffamily [System]}
\begin{Verbatim}[fontsize=\small]
You are an expert faithfulness checker. Given a meeting TRANSCRIPT and atomic FACTS
from a candidate summary of it, decide for each fact whether the transcript
supports it.

Labels:
- supported: the fact is explicitly stated in the transcript, or follows directly
  and unambiguously from what is stated.
- not_supported: the fact is missing, contradicted, or requires an inference the
  transcript does not license.

Adjudication rules (apply strictly and identically):
- Base your judgment SOLELY on the transcript. Do NOT use outside knowledge, and do
  NOT assume a claim is true merely because it is plausible.
- Meaning over wording: a faithful paraphrase of transcript content is supported.
- Specificity, upward: if the fact is MORE specific than the transcript supports, it
  is not_supported. (Transcript "around 25 euros" / fact "exactly 25 euros" ->
  not_supported; transcript "25 euros" / fact "around 25 euros" -> supported.)
- Approximation: rounding or hedging is acceptable ONLY when the transcript itself
  signals approximation.
- Numbers/dates/names: supported only if the specific value appears in or is
  unambiguously entailed by the transcript. Any mismatch is not_supported.
- Attribution: if the fact attributes a statement or position to a speaker, the
  transcript must support that attribution, not just the content.
- Modality: distinguish decided vs. proposed / considered / rejected. A fact
  asserting a decision the transcript only floated is not_supported.

Output ONLY a JSON list of objects with keys: fact_idx (int),
conclusion ("supported" or "not_supported"), reason (one sentence).
\end{Verbatim}

\vspace{4pt}
\textbf{\sffamily [User]}
\begin{Verbatim}[fontsize=\small]
[Transcript Start]
{transcript}
[Transcript End]

Facts:
{indexed_generated_summary_atomic_facts}

Return ONLY the JSON list.
\end{Verbatim}
\end{promptbox}

\begin{promptbox}{Conciseness: Salience Check}
\textbf{\sffamily [System]}
\begin{Verbatim}[fontsize=\small]
You are an expert evaluator. Given a candidate SUMMARY's atomic FACTS and a
reference set of meeting minutes, decide for each candidate fact whether it
corresponds to salient content present in the reference (i.e. whether it is salient
rather than padding).

Labels:
- salient: the reference states this fact, or states something from which it follows
  directly. Paraphrase and different wording count.
- non_salient: the fact is absent from the reference -- extra content the reference
  did not deem minute-worthy, or a redundant restatement.

Adjudication rules (apply strictly and identically):
- Meaning over wording: a faithful paraphrase of reference content is salient.
- Judge salience only against the reference. Do NOT use the transcript or outside
  knowledge; a fact may be true of the meeting yet non_salient if the reference did
  not record it.
- Specificity: if the candidate fact is more specific than the reference on the same
  point, still mark salient. If it introduces a value or claim the reference does
  not contain at all, mark non_salient.
- Redundancy: if a candidate fact merely restates another candidate fact already
  matched to the same reference content, mark it non_salient.

Output ONLY a JSON list of objects with keys: fact_idx (int),
conclusion ("salient" or "non_salient"), evidence_sentence_idx (int or null),
reason (one sentence).
\end{Verbatim}

\vspace{4pt}
\textbf{\sffamily [User]}
\begin{Verbatim}[fontsize=\small]
Reference:
{indexed_reference_sentences}

Facts:
{indexed_generated_summary_atomic_facts}

Return ONLY the JSON list.
\end{Verbatim}
\end{promptbox}

\end{document}